\documentclass{article}

    \PassOptionsToPackage{numbers, compress}{natbib}

\usepackage[final]{neurips_2022}

\usepackage[utf8]{inputenc} 
\usepackage[T1]{fontenc}    
\usepackage{hyperref}       
\usepackage{url}            
\usepackage{booktabs}       
\usepackage{amsfonts}       
\usepackage{nicefrac}       
\usepackage{microtype}      
\usepackage[dvipsnames]{xcolor}       
\usepackage{amsmath}
\usepackage[ruled, vlined]{algorithm2e}
\usepackage{algpseudocode} 
\usepackage{graphicx}
\usepackage{bbm}
\usepackage{amsthm}   
\usepackage{amssymb}
\usepackage{caption}
\usepackage{wrapfig}
\usepackage{subcaption}
\usepackage{float}
\usepackage{multirow}
\usepackage{array}
\usepackage{makecell}
\usepackage{enumitem}

\definecolor{mydarkblue}{rgb}{0,0.08,0.45}
\hypersetup{ %
    colorlinks=true,
    linkcolor=mydarkblue,
    citecolor=mydarkblue,
    filecolor=mydarkblue,
    urlcolor=mydarkblue,
    }
\numberwithin{equation}{section}
\newtheorem{theorem}{Theorem}[section]
\newtheorem{corollary}[theorem]{Corollary}
\newtheorem{lemma}[theorem]{Lemma}
\newtheorem{assumption}[theorem]{Assumption}
\newtheorem{proposition}[theorem]{Proposition}
\newtheorem{definition}[theorem]{Definition}

\usepackage{amsmath,amsfonts,bm}

\def\eqref#1{equation~\ref{#1}}

\def\algref#1{algorithm~\ref{#1}}

\def\1{\bm{1}}

\DeclareMathAlphabet{\mathsfit}{\encodingdefault}{\sfdefault}{m}{sl}
\SetMathAlphabet{\mathsfit}{bold}{\encodingdefault}{\sfdefault}{bx}{n}

\def\gA{{\mathcal{A}}}

\def\gD{{\mathcal{D}}}

\def\gL{{\mathcal{L}}}
\def\gM{{\mathcal{M}}}

\def\gO{{\mathcal{O}}}

\def\gS{{\mathcal{S}}}

\def\gW{{\mathcal{W}}}
\def\gX{{\mathcal{X}}}

\newcommand{\E}{\mathbb{E}}
\newcommand{\Ls}{\mathcal{L}}

\newcommand{\TV}{\mathcal{D}_{\mathrm{TV}}}

\title{When to Update Your Model: Constrained Model-based Reinforcement Learning}

\author{%
  Tianying Ji$^1$,  Yu Luo$^1$, Fuchun Sun\thanks{Corresponding authors: Fuchun Sun.}$\hspace{0.35em}^{,1}$, 
  Mingxuan Jing$^2$, Fengxiang He$^3$, Wenbing Huang$^{4,5}$\\
  $^1$ Department of Computer Science and Technology, Tsinghua University\\
  $^2$ Science \& Technology on Integrated Information System Laboratory, \\
  Institute of Software Chinese Academy of Sciences \\
  $^3$ JD Explore Academy, JD.com Inc \\
  $^4$ Gaoling School of Artificial Intelligence, Renmin University of China\\
  $^5$ Beijing Key Laboratory of Big Data Management and Analysis Methods, Beijing, China\\
  \texttt{\{jity20, luoyu19\}@mails.tsinghua.edu.cn; fcsun@tsinghua.edu.cn;}\\
  \texttt{jingmingxuan@iscas.ac.cn; fengxiang.f.he@gmail.com; hwenbing@126.com}
}

\begin{document}

\maketitle

\begin{abstract}
Designing and analyzing model-based RL (MBRL) algorithms with guaranteed monotonic improvement has been challenging, mainly due to the interdependence between policy optimization and model learning. 
Existing discrepancy bounds generally ignore the impacts of model shifts, and their corresponding algorithms are prone to degrade performance by drastic model updating.
In this work, we first propose a novel and general theoretical scheme for a non-decreasing performance guarantee of MBRL.  
Our follow-up derived bounds reveal the relationship between model shifts and performance improvement. 
These discoveries encourage us to formulate a constrained lower-bound optimization problem to permit the monotonicity of MBRL.
A further example demonstrates that learning models from a dynamically-varying number of explorations benefit the eventual returns. 
Motivated by these analyses, we design a simple but effective algorithm CMLO\footnote[1]{Code is available in \href{https://github.com/jity16/When-to-Update-Your-Model-Constrained-Model-based-Reinforcement-Learning}{https://github.com/jity16/When-to-Update-Your-Model-Constrained-Model-based-Reinforcement-Learning}.}
(Constrained Model-shift Lower-bound Optimization), 
by introducing an event-triggered mechanism that flexibly determines when to update the model.  
Experiments show that CMLO surpasses other state-of-the-art methods and produces a boost when various policy optimization methods are employed.

\end{abstract}

\section{Introduction}\label{introduction}
Reinforcement learning (RL)  has driven impressive advances in many complex decision-making problems in recent years~\cite{mnih2015human, silver2016mastering}. Many of these advances are obtained by model-free (MFRL) methods, whose desirable asymptotic performance yet comes at the cost of sample efficiency. Hence their applications are mostly limited to simulation scenarios~\cite{schulman2015high,lillicrap2016continuous, atari2013}. In contrast, Model-Based RL (MBRL) methods, which learn a transition model directly from orders-of-magnitude fewer samples and then derive the optimal policy from the learned model, have become an appealing alternative in small-data and more practical cases~\cite{morimoto2002minimax, deisenroth2011learning, polydoros2017survey, hester2012rtmba}.

In general, MBRL methods alternate between the two stages: model learning and policy optimization (\emph{e.g.}  the general Dyna-style~\cite{sutton1990integrated,sutton1991dyna}). A more accurate model will lead to a better policy. Various attempts have been proposed to improve model accuracy by investigating high-capacity models (the model ensemble technique~\cite{kurutach2018model,chua2018deep} and better function approximators~\cite{gal2016improving,nagabandi2018neural}) or amending the policy optimization stage based on model bias~\cite{kalweit2017uncertainty,pan2020trust,lai2021effective,janner2019trust,buckman2018sample, yu2020mopo}.  
Even so, their resultant models are just accurate in a local and relative sense, since the learning is conditional on a fixed number of state-action tuples explored by the policy at the current step, other than the full transition dynamics of the environment. Indeed, it is tricky to determine how much we should explore.  Insufficient exploration would trap the model and the following policy optimization, whereas excessive newly-encountered state-action pairs would confuse the model and later cause policy chattering. 
To derive a ``truly'' accurate model, we need a smarter scheme to choose different numbers of explorations at different times, instead of the unchanged setting in current methods. 

From an optimization point of view, the thinking of how to derive an accurate model for MBRL in a global sense also motivates us to investigate the monotonicity of the optimization target (\emph{i.e.} the return of the learned model and policy in MBRL), which, unfortunately, is less explored and not well guaranteed in current works. 
However, discussing the monotonicity guarantee for MBRL is challenging by any means, arising from the coupling of the model learning and policy optimization processes. 
Although there has been recent interest in related subjects, most of the theoretical works seek to characterize the monotonicity in terms of a fixed model of interest~\cite{sun2018dual, luo2018algorithmic,janner2019trust,farahmand2017value,lai2021effective}, which does not naturally fit our case when the model is dynamically shifted.

In this paper, we study how to guarantee the optimization monotonicity theoretically, upon which we then develop an event-triggered strategy that learns the model from a dynamically-varying number of explorations. 
In particular, we interestingly find that the lower bound of the performance improvement between two adjacent alternation steps in MBRL is dependent on the one-step model accuracy plus the constraint of the model shifts under certain mild assumptions. This discovery encourages us to formulate a constrained optimization problem, in order to permit positive performance improvement and thus the optimization monotonicity for MBRL. We also give a feasible solution example to show that dynamical alternation between model learning and policy exploration does benefit performance monotonicity. To resolve the constrained optimization problem, we design a simple but effective algorithm CMLO (Constrained Model-shift Lower-bound) equipped with  an event-triggered mechanism. 
This mechanism first estimates whether the model shifts meet the constraint and then decides when to train the model.

We evaluate CMLO on several continuous control benchmark tasks. The results show that CMLO learns much faster than other state-of-the-art MBRL methods and yields promising asymptotic performance compared with the model-free counterparts. Note that our optimization framework is general and can be applied to different backbones of policy optimization algorithms, which is also ablated in our experiments.

\section{Related works}\label{background}
Model-based reinforcement learning methods have shown great potential for sequential decision-making both in simulation and in the real world due to their sample efficiency~\cite{deisenroth2011learning, kaelbling1996reinforcement}. Generally, these MBRL algorithms can be grouped into several categories to highlight the range of uses of predictive models~\cite{wang2019benchmarking}. And our work falls into the Dyna-style category. In Dyna-style algorithms, training alternates between model learning under policy iterations with the real environments, and policy optimization using the model rollouts~\cite{sutton1990integrated,sutton1991dyna, sutton1991planning, feinberg2018model}. 
Many attempts have been devoted to improving these two stages.

For model learning stage, previous main concerns are function approximators and training objectives.  
The dynamics approximator has advanced from  Gaussian processes~\cite{ko2009gp, deisenroth2011learning}, time-varying linear dynamics~\cite{levine2013guided, levine2016end} to neural network predictive models~\cite{gal2016improving,nagabandi2018neural}. 
And training objectives vary from Mean Square Error (MSE)~\cite{nagabandi2018neural,luo2018algorithmic}, Negative Log Likelihood (NLL)~\cite{chua2018deep, janner2019trust}, \emph{etc}. 
Moreover, the deep ensemble technique is appealing for improving the robustness to model error.  Our method adopts an ensemble of probabilistic networks similarly as in ~\cite{chua2018deep,janner2019trust}.

The policy optimization stage allows Dyna-style algorithms to leverage various off-the-shelf model-free methods, such as SAC~\cite{haarnoja2018soft}, TRPO~\cite{schulman2015trust}, and TD3~\cite{fujimoto2018addressing}. 
Much owing to the progress of model-free methods,
our method is to invoke any reasonable optimization oracle for the empirical models, 
rather than entangle a particular policy optimization algorithm.

A consensus of MBRL is that a smart policy requires an accurate model.
However, model bias cannot be eliminated because the state-action distribution of the samples in the model learning stage and policy optimization stage is quite different.
Many prior works attend to this distribution mismatch issue and then tailor the data used for policy optimization according to the model bias. 
~\citet{janner2019trust,buckman2018sample}  encourage truncated rollout lengths. 
Besides, the ratio of real to model-generated data can be dynamically tuned according to the model uncertainty~\cite{kalweit2017uncertainty, pan2020trust,lai2021effective}.  ~\citet{yu2020mopo} penalizes rewards by the model uncertainty. 
Our method incorporates the truncated model rollouts mechanism.  Moreover, we further explore how real interactions affect overall performance which is rarely studied before. We construct an event-triggered mechanism to cope with overfitting in a small-data regime and suffering generalization error when facing a drastic distribution shift.

Monotonic improvement guarantee has been a fundamental concern in both model-free and model-based avenues. 
In MFRL methods, both CPI~\cite{kakade2002approximately}  and TRPO~\cite{schulman2015trust}  can be understood as approximating and optimizing the performance gap by forcing the new policy to be not too far away from the current policy. However,  in Model-based settings, their trust-region constraints cannot directly be satisfied because the policies highly depend on the randomness of the models. While constructing such a bound for performance gap is straightforward, it has not been explored in previous MBRL theorectical analyses, instead they~\cite{sun2018dual, luo2018algorithmic,janner2019trust,farahmand2017value,lai2021effective} turn to bound the discrepancy between returns under a model and those in the real environment. Although they could guarantee that the lower bound of policy performance improves under a certain model, this guarantee may face several issues regarding model updating. 
In contrast, we construct the performance difference scheme for MBRL algorithms and perform monotonicity analysis under this scheme.

Another line of theoretical works focus on  regret bounds~\cite{efroni2019tight, kakade2020information} or sample complexity properties~\cite{agarwal2020model, azar2013minimax, sidford2018near}, focusing on the convergence performance and sample complexity for model-based approaches.

\section{Preliminaries}\label{preliminaries}
\paragraph{Markov Decision Process}
A discounted Markov Decision Process (MDP) is a quintuple $M = ({\cal S},{\cal A}, P_{M}, r_{M}, \gamma)$, where ${\cal S}$  is the state space, ${\cal A}$ represents the action space, $P_M$ denotes the transition function, $r:{\cal S\times A}\rightarrow [-R, R]$ stands for the reward function, and $\gamma\in (0,1)$ is the discount factor. For a fixed policy $\pi$ and model $M$, we define  $V_{M}^\pi(\mu)$ as the return of the model ${M}$ with the starting state distribution $\mu$, and $V^{\pi}(\mu)$ denotes the returns under the real environment, 
\begin{equation}
    V_M^{\pi}(\mu) = \mathop{\mathbb{E}}\limits_{a_t\sim \pi(\cdot\vert s_t)\atop s_{t+1}\sim P_{M}(\cdot\vert s_t, a_t)}\Big[\sum\limits_{t=0}^\infty \gamma^t r_M(s_t,a_t)\vert \pi, s_0\Big].
\end{equation}
We make a mild assumption that the model $M$ to be identical to the real MDP except the transition function. 
Let $d_{M_i}^{\pi_k}(s,a;\mu)$ denote the visitation probability $s, a$ when starting at $s_0 \sim \mu$ and following $\pi_k$ under the dynamics $P_{M_i}$.   We will omit it as  $d_{M_i}^{\pi_k}$ henceforth for brevity. 
Besides, we denote $\cal M$ as a (parameterized) family of models of interest, and let $\Pi$ be a family of policies.

\paragraph{Generative Model} Many previous works~\cite{li2020breaking,kearns2002sparse,agarwal2020model}  focus on  a stylized generative model. By assuming an access to the generative model, we collect $N$ samples for each state-action pair $(s,a)\in {\cal S\times \cal A}$: $s_{s,a}^i \stackrel{i.i.d}\sim P(\cdot\vert s,a)$ which allows us to construct an empirical model defined as follows:
\begin{equation}
    \forall s' \in {\cal S},\quad \hat{P}(s'\vert s,a) = \frac{1}{N}\sum\limits_{i=1}^{N} \mathbbm{1}\{s^i_{s,a}=s' \}.
\end{equation}
where $\mathbbm{1}\{\cdot\}$ is the indicator function. This leads to an empirical MDP $\hat{M} = ({\cal S},{\cal A}, \hat{P}, r,\gamma)$.

\section{Monotonic improvement under model shifts}\label{Methods}

This section provides theoretical analyses for monotonic improvement of MBRL, factoring in the interdependence between policies and models.  
We first construct a general scheme for a non-decreasing performance guarantee and follow it up by characterizing the lower bound when shifting the model.
Towards a non-negative lower bound, we restrict the model shifts and then obtain a refined bound.
These discoveries encourage us to translate the bound maximization to a constrained optimization problem to permit monotonicity.
By deriving an instance solution under the generative model setting, we demonstrate the merits of the dynamic model learning interval. 

\subsection{Monotonic improvement with policy optimization oracle}\label{para:general-recipe}

Our goal is to construct a general recipe for a monotonicity guarantee.  
Naturally, we seek to build a performance difference scheme for model-based algorithms. 
\begin{definition}[\textbf{Performance Difference Bound Scheme}]\label{definition}
$V^{\pi_i\vert M_i}(\mu)$ denotes the return of the policy $\pi_i \in \Pi$ in the real environment, whereas $\pi_i$ is derived from the dynamical model $M_i \in \gM$.  Then, the lower bound on the true return gap of $\pi_1$ and $\pi_2$ can be stated in the form, 
\begin{equation}
    V^{\pi_2\vert M_2}(\mu) - V^{\pi_1\vert M_1}(\mu) \geq C.
\end{equation}
Such a statement guarantees that the policy allows non-decreasing performance in the real environment  when $C$ is non-negative.
\end{definition}

Although there has been interest in non-decreasing performance guarantee, previous works~\cite{luo2018algorithmic,janner2019trust} commonly derive under a "discrepancy bound" scheme disregarding the model shifts (\emph{i.e.} $M_1= M_2$). Their results imply that once a policy update $\pi_1 \rightarrow \pi_2$ increases the returns under the same model ($V_{M_1}^{\pi_2} > V_{M_1}^{\pi_1}$), the lower bound on the policy performance evaluated in the real environment  improves accordingly, $\inf\{V^{\pi_2\vert M_1}\} > \inf \{ V^{\pi_1\vert M_1} $\}.

When the model shift is introduced, establishing the performance difference bound turns out to be rather difficult, mainly due to the coupling of the policy optimization and model learning: 
the estimated model is generated from the policy explorations, while the policy derives from the model rollouts. 
Hence, we need to consider the performance gap arising from the integration of the two stages, which, unfortunately, has never been explored before. Since the performance of the policy with a fixed model is already well guaranteed, it is natural to make the following assumptions.

\begin{assumption}[\textbf{Policy Optimization Oracle}]
The policy optimization oracle is defined as the one that takes as input a model $M$ and returns a ${\epsilon_{opt}}$-optimal policy $\pi$ satisfying: $V_M^*(\mu)-\epsilon_{opt}\leq V_M^\pi(\mu)\leq V_M^*(\mu)$. We assume that our policy optimization stage always meets the policy optimization oracle given its corresponding estimated model $M$. 
\end{assumption}

Such assumption usually holds in practice when we explore existing off-the-shelf model-free algorithms~\cite{puterman2014markov, agarwal2020model, haarnoja2018soft, schulman2015trust}. With this assumption, we can focus directly on the eventual performance difference under the real environment encountering the model updating.

The bound $C$ we seek can be expressed in terms of two kinds of gaps: the inconsistency gap between the model and the environment, and the optimal returns gap between the two models.
When a policy $\pi_k$ samples in a model $M_i$, it will encounter states not consistent with those generated by the real environment. 
We denote this inconsistency by $\epsilon_{M_i}^{\pi_k} = \mathbb{E}_{s,a\sim d^{\pi_k}} [\TV (P(\cdot\vert s,a)\Vert  P_{M_i}(\cdot\vert s,a))]$. 
Besides, for each model $M_i \in {\cal M}$, a remarkable property is that there always exists a policy that maximizes the value function $V^{\pi}_{M_i}(\mu)$ ~\cite{sutton2018reinforcement}. Hence, we define the ceiling performance of model $M$ as $V^*_{M_i}(\mu) = \sup_{\pi\in \Pi} V^\pi_{M_i}(\mu)$. We will omit $\mu$ henceforth for simplicity unless confusion exists. 
With these two terms well-defined, we now present our bound.

\begin{theorem}[\textbf{Performance Difference Bound for Model-based RL}]\label{thm: general-bound}
Let $M_i \in \gM$ be the estimated models and $\pi_i$ be the $\epsilon_{opt}$-optimal policy for $M_i$. Recalling $\kappa = \frac{2R\gamma}{(1-\gamma)^2}$ where $R$ is the bound of the reward function, we have the performance difference of $\pi_1$ and $\pi_2$ evaluated under the real environment be bounded as below,
\begin{equation} \label{general-bound}
    V^{\pi_2\vert M_2} - V^{\pi_1\vert M_1} \geq \kappa \cdot (\epsilon_{M_1}^{\pi_1} - \epsilon_{M_2}^{\pi_2}) + V_{M_2}^* - V_{M_1}^*  - \epsilon_{opt}.
\end{equation}
\end{theorem}

\begin{proof}
 See Appendix \ref{ap-omittedproofs}, Theorem \ref{proof-general-gap}.
\end{proof}

This bound implies that if the model update $M_1 \rightarrow M_2$ can (1) shorten the divergence between the estimated  dynamics and the true dynamics and (2) improve the ceiling performance on the model, it may guarantee overall performance improvement under the true dynamics. 

\subsection{Lower-bound optimization with model shift constraints} \label{para: refine}

Theorem~\ref{thm: general-bound} provides a general performance difference bound that suggests models with higher ceiling performance and lower bias would raise the overall performance. 
However, finding a non-negative lower bound may face several issues regarding a drastic model shift in practice. 
On the one hand, performing an abrupt model update could potentially lead to a tumble in ceiling performance and then fail to make incremental improvements. On the other hand, huge distribution divergence between model rollouts would confound the policy optimization stage and hamper its access to optimal policies. Thus, we shoot for refining the bound upon adding model shifts constraint.

First, we seek to further unfold the ceiling performance gap, which serves as the main building block toward the desired bound. During derivation, we additionally introduce the $L$-Lipschitz assumption.
\begin{assumption}[\textbf{$L$-Lipschitzness of Value Function}]
We call a value function $V^{\pi}_{M}$ on the estimated dynamical model $M$ is $L$-Lipschitz  w.r.t to some norm $\Vert \cdot \Vert$ in the sense that 
\begin{equation}
    \forall s,s'\in {\cal S}, \vert V^\pi_{M}(s) - V^\pi_{M}(s')\vert \leq L\cdot \vert s-s'\vert.  
\end{equation}
\end{assumption}
We assume that our estimated model $M \in {\cal M}$ satisfies the Lipschiz character, inspired from~\cite{asadi2018lipschitz, luo2018algorithmic}. 
Under the $L$-Lipschitzness assumption, we can derive a bound for the ceiling return gap.
\begin{theorem}[\textbf{Ceiling Return Gap under Model Shift}]
For an estimated model $M_i \in \gM$,  the ceiling return gap is  bounded as:
\begin{equation}
    V_{M_2}^* - V_{M_1}^*\geq - \frac{\gamma}{1-\gamma}L\cdot \sup_{\pi\in \Pi} \mathbb{E}_{s,a\sim d^\pi_{M_2}}
\Big[\vert  P_{M_2}(\cdot\vert s,a)- P_{M_1}(\cdot\vert s,a)\vert\Big].
\end{equation}
\end{theorem}
\begin{proof}
 See Appendix \ref{ap-omittedproofs}, Theorem \ref{ap:ceiling-bound}.
\end{proof}
This conclusion reveals the connection between the ceiling return gap and models' disagreement.
In a benign scenario, the term of ceiling performance gap in the RHS of Eq.~\ref{general-bound} should be dominated by the term $\kappa \cdot (\epsilon_{M_1}^{\pi_1} - \epsilon_{M_2}^{\pi_2}) $ when the model shift is sufﬁciently small. A sharp model shift, however, risks a massive reduction in ceiling performance that is hardly bridged by the other parts in Eq.~\ref{general-bound}, and therefore corrupts monotonicity. 
This inspires us to introduce the constraint of the model shift. We further refine our performance difference bound to better characterize the relationship between performance gap and model shift.

\begin{theorem}[\textbf{Refined Bound with Constraint}]
\label{refine-bound}
Let policy $\pi_i\in \Pi$ denotes the $\epsilon_{opt}$ optimal policy under the dynamical model $M_i\in \gM$, and $\sigma_{M_1, M_2}$ be the constraint threshold for $M_1$ and $M_2$. Note that $L \geq \frac{R}{1-\gamma}$.
Then we can refine performance difference lower bound under the model shifts constraint as, 
\begin{align}
\begin{split}
     V^{\pi_2\vert M_2} - V^{\pi_1\vert M_1}& \geq   
     \kappa\cdot\Big\{ \mathbb{E}_{s,a\sim d^{\pi_1}} \TV \big[ P(\cdot \vert s,a) \Vert  P_{M_1}(\cdot\vert s,a)\big]  
     \\- & \mathbb{E}_{s,a\sim d^{\pi_2}} \TV \big[ P(\cdot \vert s,a)\Vert P_{M_2}(\cdot\vert s,a)\big] \Big\}
     -  \frac{\gamma}{1-\gamma}L\cdot (2\sigma_{M_1,M_2}) - \epsilon_{opt},
     \label{eq: refined-lowerboud}
\end{split}\\
         \textit{s.t.} \quad& \TV(P_{M_2}(\cdot\vert s,a) \Vert P_{M_1}(\cdot\vert s,a)) \leq  \sigma_{M_1,M_2}, \quad \forall (s,a) \in {\cal S\times A}.
         \label{eq: model-shift-constraints}
\end{align}

\end{theorem}
\begin{proof}
 See Appendix\ref{ap-omittedproofs}, Theorem \ref{proof-constraint-bound}.
\end{proof}

Theorem~\ref{refine-bound} implies that policy $\pi_2$ is guaranteed to outperform policy $\pi_1$ once it makes the RHS  in Eq.~\ref{eq: refined-lowerboud} greater than zero under the constraint Eq.~\ref{eq: model-shift-constraints}. More specifically, to guarantee a non-decreasing performance, $M_2$ and $\pi_2$ should meet the following two requirements,
\begin{align*}
    \label{R1}
\mathop{\E}\limits_{s,a\sim d^{\pi_2}} \Big[ \sum_{s'\in \gS}\vert  P(s'\vert s,a) - P_{M_2}(s'\vert s,a) \vert \Big] \leq 2\epsilon_{M_1}^{\pi_1} - \frac{(1-\gamma)L}{R}(2\sigma_{M_1,M_2}) - \frac{2}{\kappa}\cdot \epsilon_{opt},\tag{R1}
\\
\label{R2}
\TV(P_{M_2}(\cdot\vert s,a) \Vert P_{M_1}(\cdot\vert s,a)) \leq  \sigma_{M_1,M_2}, \quad \forall (s,a) \in {\cal S\times A}. \tag{R2}
\end{align*}
\ref{R1} encourages us to alleviate model bias as much as possible.
Moreover, when the policy $\pi_1$ samples too many states not encountered by the model $M_1$, it may lead to excessive generalization errors during model learning, i.e. the estimated model will suffer extrapolation error in these unexperienced regions, which further causes instability or crashes in the following derived sub-optimal policies. Thus, the introduction of the \ref{R2} constraint can help solve the above problem. Finally, we abstract these two requirements to a constrained  optimization problem.
\begin{proposition}[\textbf{Constrained Lower-Bound Optimization Problem}]\label{simple-opt} 
We reduce the issue of finding a non-negative $C$ to the following constrained optimization problem. Here, $\pi_i$ is still the sub-optimal policy under model $M_i$. The minimal objective in E.q.~\ref{eq:constrained-optimization} leads to the maximum of $C$.  Then the overall optimization problem can be formalized as,
\begin{equation} \label{eq:constrained-optimization}
    \begin{aligned}
& \min_{M_2\in {\cal M}\atop \pi_2 \in \Pi}  \mathop{\E}\limits_{s,a\sim d^{\pi_2}} \Big[ \sum_{s'\in \gS}\vert  P(s'\vert s,a) - P_{M_2}(s'\vert s,a) \vert \Big],\\
& \textit{s.t.}\quad  \sup\limits_{s\in \gS, a\in \gA}\TV (P_{M_1}(\cdot \vert s,a)\Vert P_{M_2}(\cdot \vert s,a) )\leq \sigma_{M_1,M_2}.
    \end{aligned}
\end{equation}
\end{proposition}

\subsection{A feasible example for constrained optimization problem}\label{para: example}
We first remark that, Proposition~\ref{simple-opt} provides a useful guide for acquiring performance improvement through restricting the upcoming model shift into a safe zone.  
In this section, we  give an instance for a feasible solution of the constrained optimization problem under the generative model setting~\cite{agarwal2020model,li2020breaking}. 
Specifically, we construct an empirical model $M_1$ from the N samples per state-action pair that stems from the generative model.
Upon encountering another $k$ samples on each state-action pair, we segue into the model updating stage, which outputs $M_2$ given these newly collected samples as input.
Towards obtaining a non-decreasing performance, we yield the following feasible solution for model training interval through satisfying requirements \ref{R1} and \ref{R2}.

\begin{corollary} \label{corollary-k}
Here, $vol({\cal S})$ denotes the volume of the state coverage simplex.
For simplicity, we denote $\delta_{M_i}(\cdot\vert s,a) = \vert P(\cdot\vert s,a) - P_{M_i}(\cdot\vert s,a)\vert$ for each model $M_i \in \gM$.
Under the generative model setting, with a probability larger than $1- \xi$, we can provide a non-negative $C$ when given, 
\begin{equation}     \begin{aligned}
    k = \frac{2}{\epsilon^2} \log\frac{2^{vol(\gS)}-2}{\xi}-N.
\end{aligned}
\end{equation}
Here, $ \epsilon =\delta_{M_1}(\cdot\vert s,a) - \frac{(1-\gamma)L}{R}\cdot (2\sigma_{M_1,M_2})- \frac{(1-\gamma)^2}{R\gamma}\cdot \epsilon_{opt}  $  and $\xi \in(0,1) $ is a constant. 
\end{corollary}
\begin{proof}
See Appendix \ref{ap-omittedproofs}, Corollary \ref{sample-based-corollary}.
\end{proof}

One can deduce from Corollary~\ref{corollary-k} that dynamically adjusting the model training interval according to the model bias and the model shifts constraints threshold does benefit monotonicity.
Along with the model bias $\delta_{M_1}(\cdot\vert s,a)$ decaying, the model training interval $k$ requires to be scaled up to obtain adequate newly encountered samples for training $M_2$. 
Besides, once gathering excessive samples, we risk violating the model shifts constraint, thus impairing performance.
This instance supports the insight that determining ``when to update your model'' is vital for performance improvement and motivates a smarter scheme to choose different numbers of explorations at different times instead of the unchanged setting in current methods.

\section{CMLO framework}
It is nontrivial to tackle the proposed constrained optimization problem in Proposition~\ref{simple-opt}, since one cannot directly assign a value to the optimization variable $M_2$.
We decouple the optimization objective and the constraint as ``how to train the model'' and ``when to train the model'' through an event-triggered mechanism towards dynamic alternation.

\paragraph{Objective minimization.} 
Minimizing the objective function involves improving the model accuracy. Specifically, we adopt the model-ensemble technique to reduce model bias. For practical implementation, the probabilistic models  $\{ \hat {f_{\phi_1}},\hat {f_{\phi_2}}, \ldots, \hat {f_{\phi_K}}\}$ are fitted on shared but differently shuffled replay buffer ${\cal D}_e$,  and the target is to optimize the Negative Log Likelihood (NLL).
\paragraph{Constraint estimation.} 
The unobserved model $M_2$ makes the constraint function incalculable, and here we seek to  design an estimator for it. 
Recall that $\TV(P_{M_1}(\cdot \vert s,a)\Vert P_{M_2}(\cdot \vert s,a) )  = \frac{1}{2}\sum_{s'\in \gS} \vert P_{M_1}(s' \vert s,a) - P_{M_2}(s' \vert s,a)\vert$, then we can find that the distance arises from two parts, the state space coverage, and the models' disagreement. 
We estimate the policy coverage (state-space coverage) by computing the volume $vol({\cal S}_{\cal D})$ of the convex closure $\gS_{\gD}=\big\{ \sum_{s_i\in {\cal D}} \lambda_i S_i : \lambda _i \geq 0 , \sum_{i} \lambda_i = 1\big\}$ constructed on the replay buffer ${\cal D}$.  
We exploit the average prediction error on these new samples data to estimate the disagreement on newly encountered data  ($\Delta {\cal D}$) and get ${\cal L}(\Delta{\cal D}) = \mathop{\mathbb{E}}_{(s,a,s')\in \Delta_{\cal D}} \big[\frac{1}{K}\sum_{i=1}^K \Vert s' - \hat{f_{\phi_i}}(s,a) \Vert\big]$. Combining these two components, we can obtain an estimation for the model shift, \emph{i.e.}, $vol(\gS_{\gD})\cdot \Ls(\Delta {\gD})$.  This practical overestimation for the model shifts makes the constraint stay satisfied  during objective optimization.

\paragraph{Event-triggered mechanism.}
We design an event-triggered mechanism to determine the occasion to pause collecting and turn to solve the optimization objective.  Remark that although diverting to train models as long as not to violate the constraint is theoretically reasonable, we refrain from doing so in practice because performing an update on data with a minor shift in coverage and distribution is wasteful and may risk overfitting.
Therefore, we trigger when the constraint boundary is touched to reduce computational cost and escape overfitting. The event-triggering mechanism is based on the following condition:
\begin{equation}
\label{event-condition}
    \frac{vol(\gS_{\gD \cup \Delta D(\tau)})}{vol(\gS_{\gD})} \cdot \Ls(\Delta D(\tau)) \geq \alpha .
\end{equation}
Here, $\tau$ is the event-triggering time, and $\alpha$ is a given constant.

\paragraph{Policy optimization oracle.} Clearly, we can leverage many model-free RL methods (SAC~\cite{haarnoja2018soft}, TRPO~\cite{schulman2015trust}, PPO~\cite{schulman2017proximal} \emph{etc.}) as our policy optimization oracle.  Besides, we adopt a truncated short model rollouts technique to mitigate compounding error while encouraging model usage. Based on the fresh model rollouts, we perform the policy optimization oracle, employing SAC as an example.

\paragraph{Algorithm Overview.}
We briefly give an overview of our proposed CMLO in \algref{algorithm}. 
Notably, the event-triggered mechanism subtly determines the occasion to perform model updating, promoting performance monotonicity and reducing computation load.

\begin{algorithm}[H]
\caption{CMLO}
\SetKwInOut{Input}{initialize}
\SetKwInOut{Output}{output}
\Input{policy $\pi_\theta$, ensemble models $\hat{f_{\phi_1}},\hat{f_{\phi_2}},\ldots, \hat{f_{\phi_K}}$, environment buffer ${\cal D}_e$ and model buffer ${\cal D}_m$\
}

\Repeat{the policy performs well in the environment}{
    Sample $\Delta \mathcal{D}_e \sim \pi_\theta$ from real environment; add to ${\cal D}_e$\\
    Estimate model shifts by $vol({\cal S}_{D_e})\mathcal{L}(\Delta \mathcal{D}_e)$\\
    \If{Event-triggered condition (\ref{event-condition}) is reached}{
        Train all models $\hat{f_{\phi_1}},\hat{f_{\phi_2}},\ldots, \hat{f_{\phi_K}}$ on  ${\cal D}_e$ 
    }
    Perform $h$-step model rollouts using policy $\pi_\theta$; add to ${\cal D}_m$\\
    Update $\pi_{\theta}$ on ${\cal D}_m$ through SAC~\cite{haarnoja2018soft}
}
\label{algorithm}
\end{algorithm}

\section{Experiments}
Our experimental evaluation aims to investigate the following questions: (1) How well does our algorithm perform on standard reinforcement learning benchmarks compared to prior state-of-the-art model-based and model-free algorithms? (2) Does the performance with or without the constraint consistent with previous theoretical analyses? 

\begin{figure}[h]
    \centering
    \includegraphics[width=0.95\textwidth]{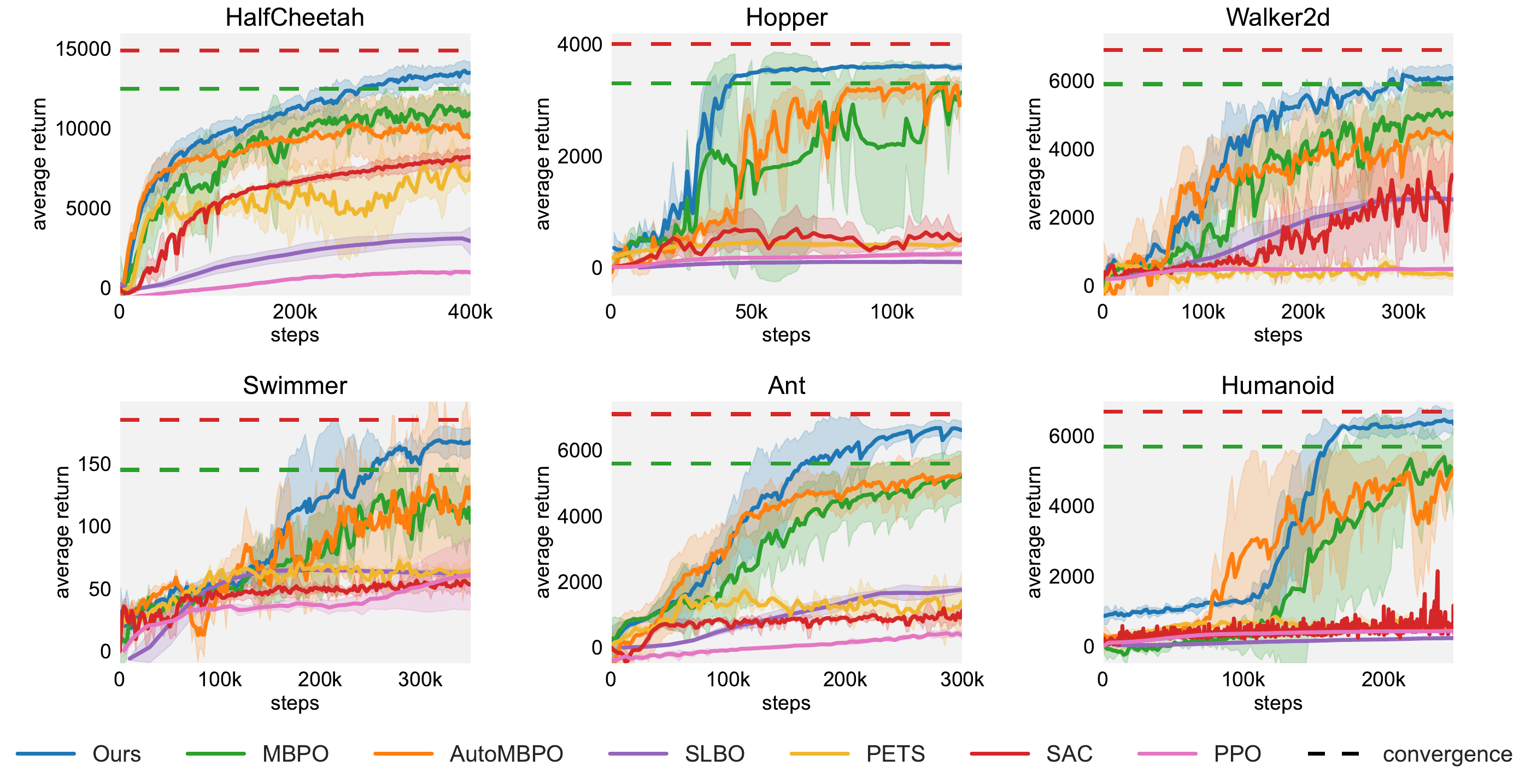} 
    \caption{Comparison of learning performance on continuous control benchmarks. We evaluate each algorithm on the standard 1000-step versions. Solid curves indicate the average performance among seven trials under different random seeds, while the shade corresponds to the standard deviation over these trails. The dashed lines are the asymptotic performance of  SAC (at 5M steps) and MBPO.}
    \label{fig:main-res}
    \vspace{-0.3cm}
\end{figure}

\subsection{Comparative evaluation}
To illustrate the effectiveness of our method, we contrast several popular model-based and model-free baselines. Model-free counterparts include: 
(1) SAC~\cite{haarnoja2018soft}, the state-of-the-art in terms of asymptotic performance. 
(2) PPO~\cite{schulman2017proximal} that explores monotonic improvement as well. 
Model-based baselines include:
(3) PETS~\cite{chua2018deep}, which employs models directly for planning, different from the Dyna-style.
(4) SLBO~\cite{luo2018algorithmic}, that explores monotonicity under the discrepancy bound scheme.
(5) MBPO~\cite{janner2019trust}, that employs a similar design of model ensemble technique (ensemble of probabilistic dynamics networks) and policy optimization oracle (SAC)  as we do.
(6) AutoMBPO~\cite{lai2021effective}, a variant of MBPO, that uses an automatic hyperparameter controller to tune the model-training frequency but suffers from high pre-training cost and lacks theoretical analysis on parameters rationality.

We evaluate CMLO and these baselines on six continuous control tasks in OpenAI Gym~\cite{brockman2016openai} with the  MuJoCo~\cite{todorov2012mujoco} physics simulator, including HalfCheetah, Hopper, Walker2d, Swimmer, Ant,  Humanoid. For fair comparison, we adopt the standard full-length version of these tasks and align the same environment settings.

Figure~\ref{fig:main-res} shows the learning curves of all compared methods, along with the asymptotic performance. These results show that our algorithm is far ahead of the model-free method in terms of sample efficiency, coupled with an asymptotic performance comparable to that of the state-of-the-art model-free counterparts SAC. 
Compared to model-based baselines, our method gains faster convergence speed and better eventual performance. Notably, credit to the event-triggered mechanism, our method enjoys a more stable training curve. The better monotonic property of the learning curve agrees with our previous analyses.

\subsection{Ablation studies}
Next, we make ablations and modifications to our method to validate the effectiveness and generalizability of the mechanism we devised.
\begin{figure}[h]
    \centering
    \begin{minipage}[t]{.49\textwidth}
        \raggedleft
         \begin{subfigure}[t]{1.0\textwidth}
            \centering
            \includegraphics[width = 0.48\textwidth]{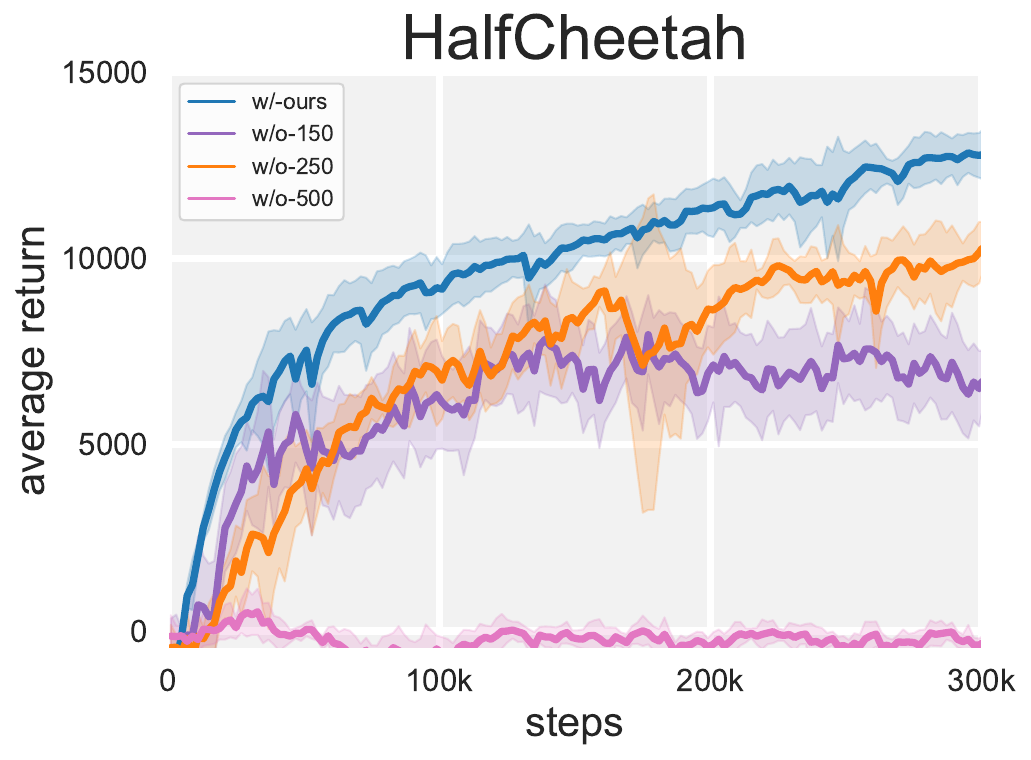}
            \includegraphics[width = 0.48\textwidth]{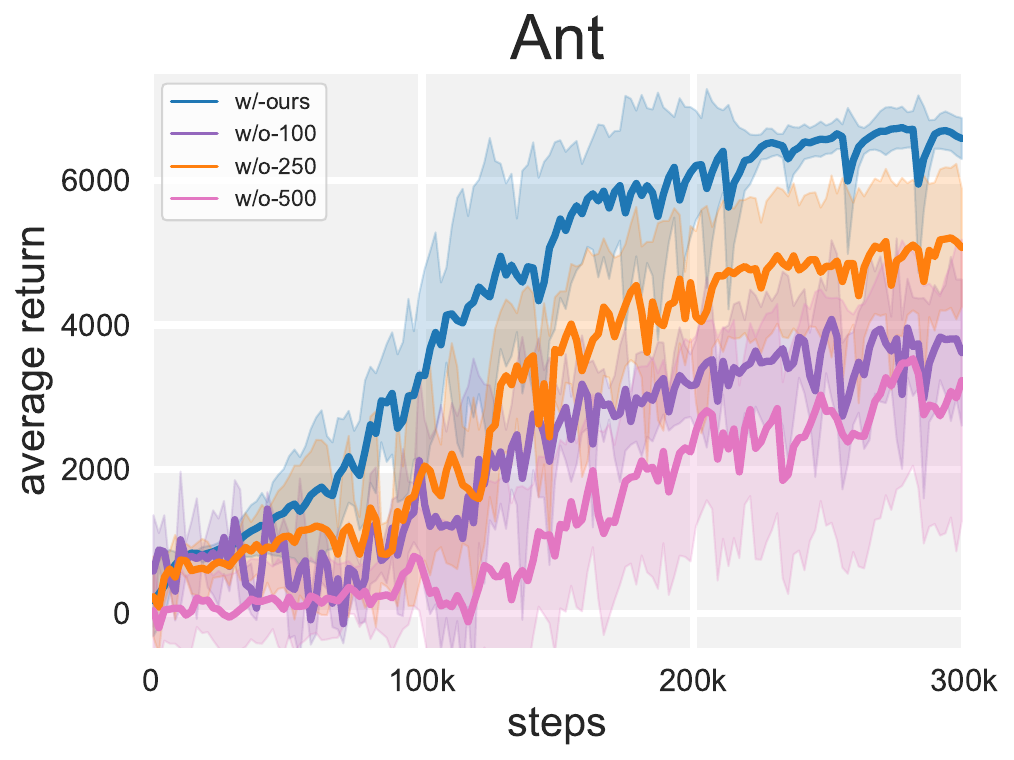} 
            \caption{average return}
            \label{fig:ab-return}
         \end{subfigure}
         \begin{subfigure}[t]{1.0\textwidth}
            \centering
            \includegraphics[width = 0.48\textwidth]{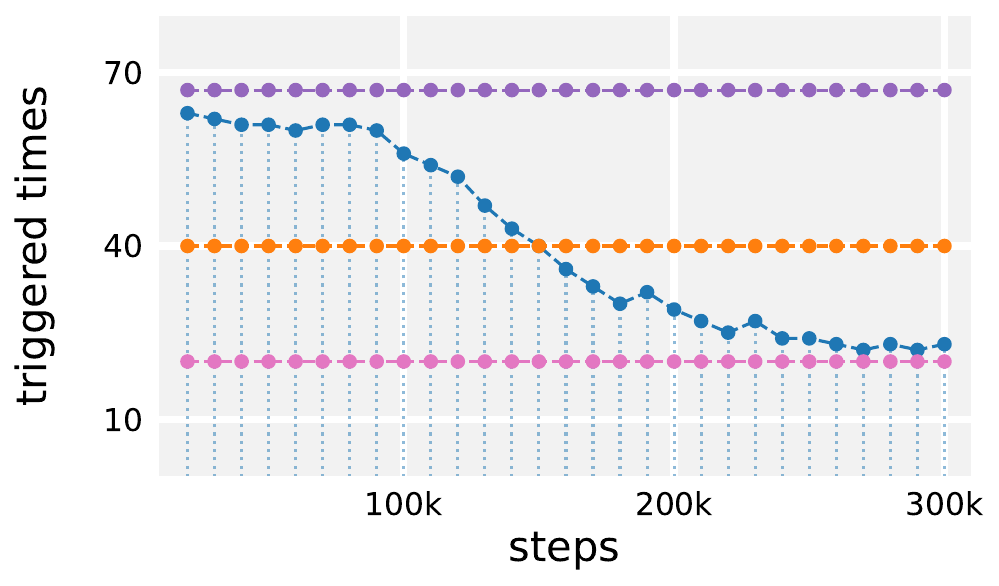}
            \includegraphics[width = 0.48\textwidth]{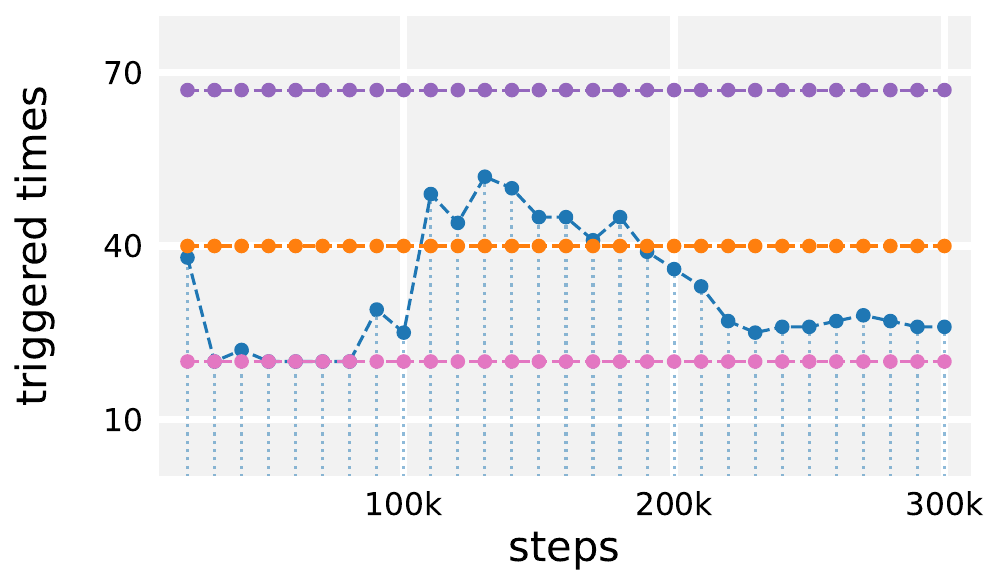}
            \caption{triggered times}
            \label{fig:ab-triggeredtimes}
         \end{subfigure}
         \caption{Ablation on event-triggered mechanism. These experiments are average over 5 random seeds. (a) shows the average return with or with-out event-triggered mechanism in HalfCheetah and Ant benchmarks. (b) shows the number of triggered times per 10k step. }
         \label{fig:main-abl}
    \end{minipage}%
    \hfill
    \hspace{5pt}
    \begin{minipage}[t]{.49\textwidth}
        \raggedright
        \begin{subfigure}[t]{1.0\textwidth}
            \centering
             \includegraphics[width=0.48\textwidth]{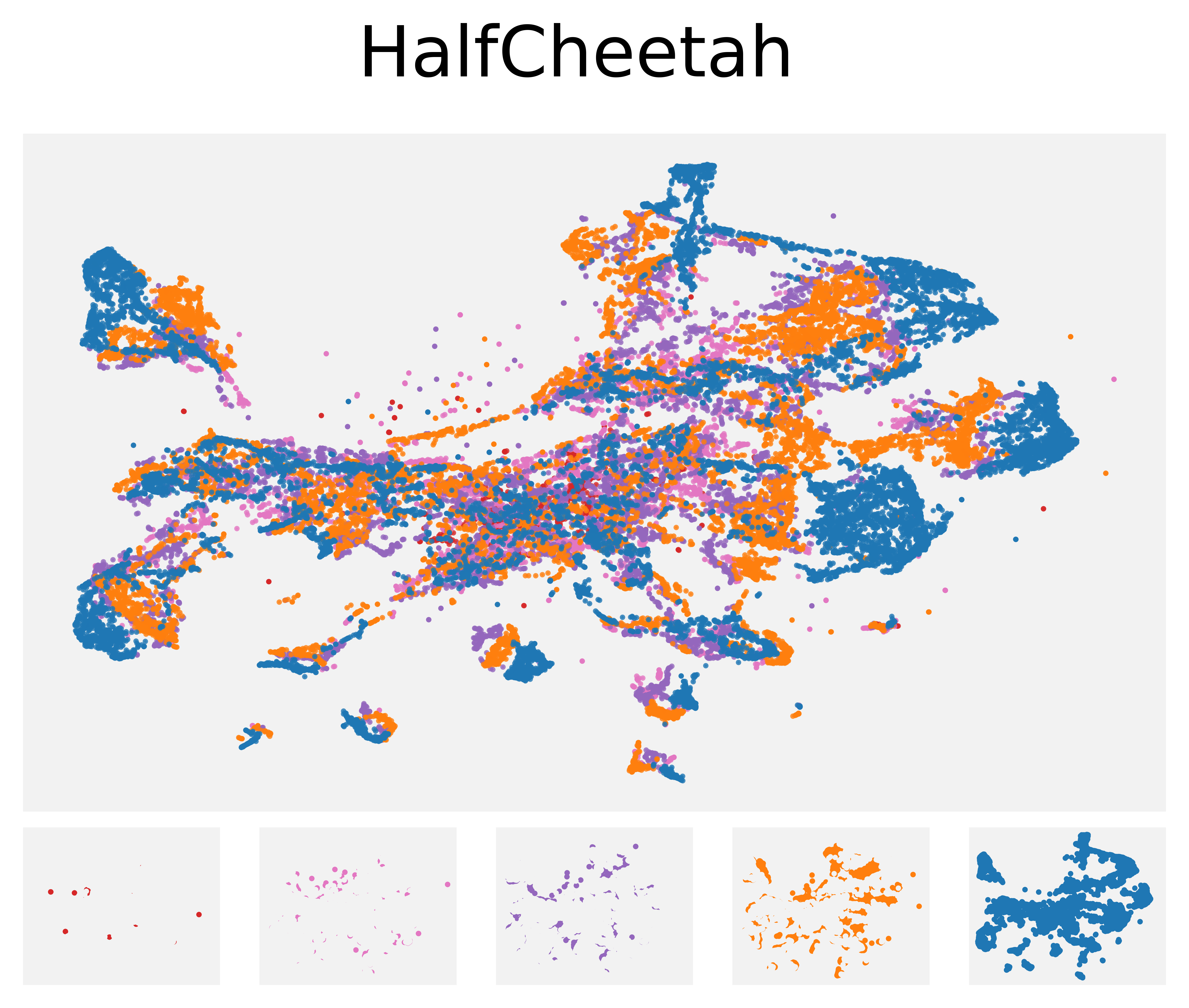}
             \includegraphics[width=0.48\textwidth]{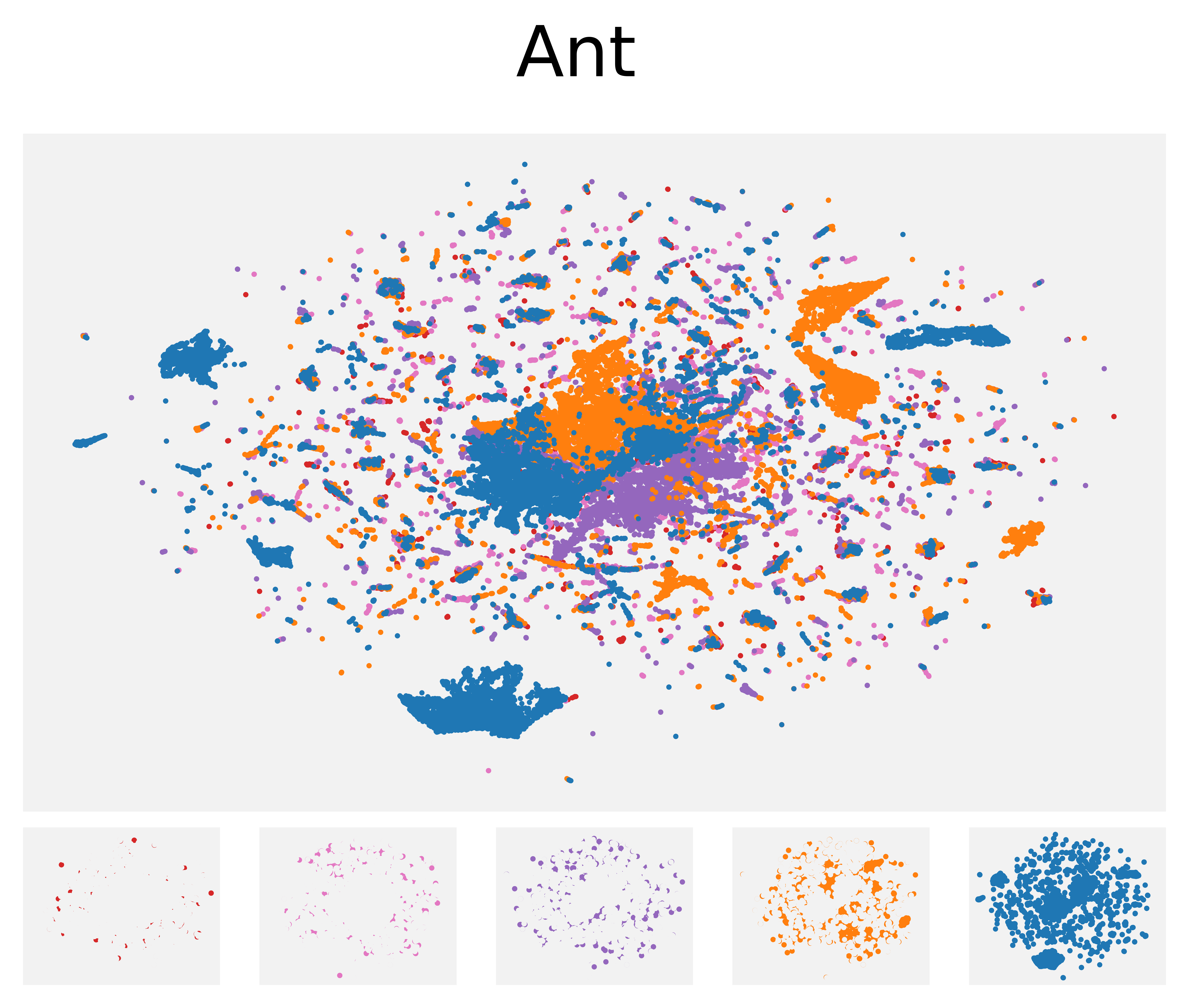}
            \label{fig:policy-coverage}
            \caption{policy coverage}
        \end{subfigure}
        \hfill
        \begin{subfigure}[t]{1.0\textwidth}
          \includegraphics[width=0.48\textwidth]{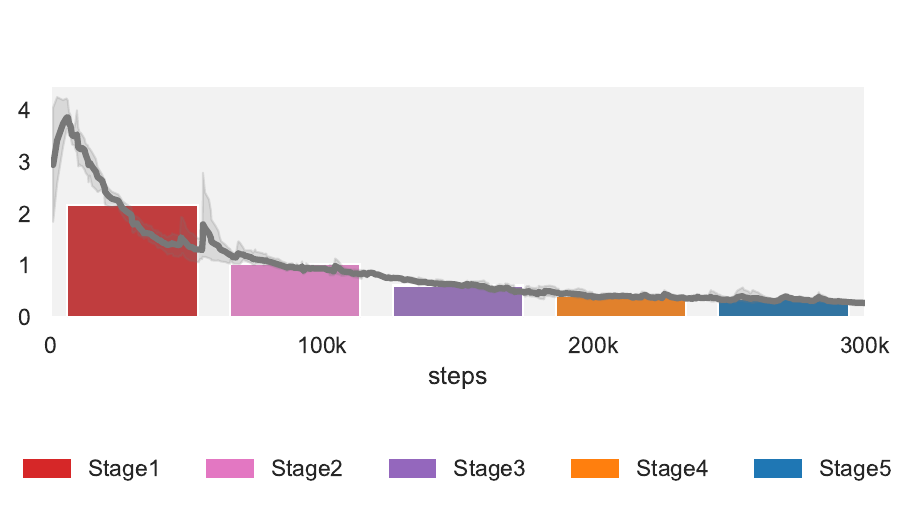}
          \includegraphics[width=0.48\textwidth]{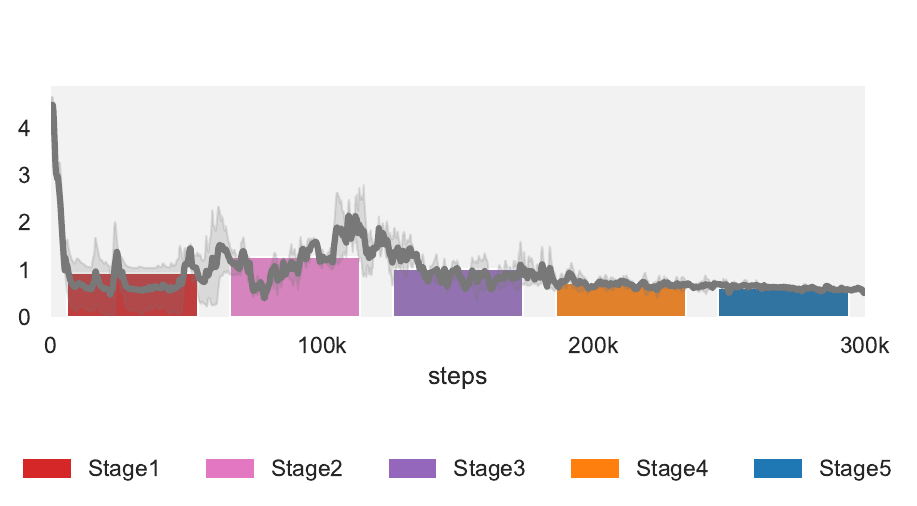}
            \caption{average prediction error}
            \label{fig:inconsistency}
        \end{subfigure}
        \caption{Visualization of the policy coverage and prediction error on HalfCheetah and Ant. Each stage contains 60k steps. (a) implies that policy coverage expands over stages.  (b) shows the prediction error over newly collected data.  The bars are average values of each stage.}
        \label{fig:event-ele}
    \end{minipage}
    \hfill
    \vspace{10pt}
    \begin{minipage}[t]{1.0\textwidth}
        \begin{subfigure}[t]{0.49\textwidth}
          \includegraphics[width=1.0\linewidth]{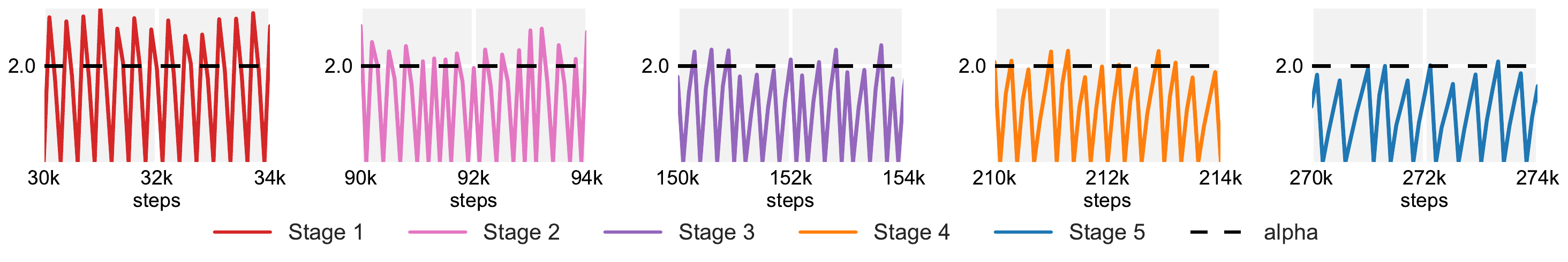}
            \caption{Halfcheetah}
        \end{subfigure}
        \begin{subfigure}[t]{0.49\textwidth}
          \includegraphics[width=1.0\linewidth]{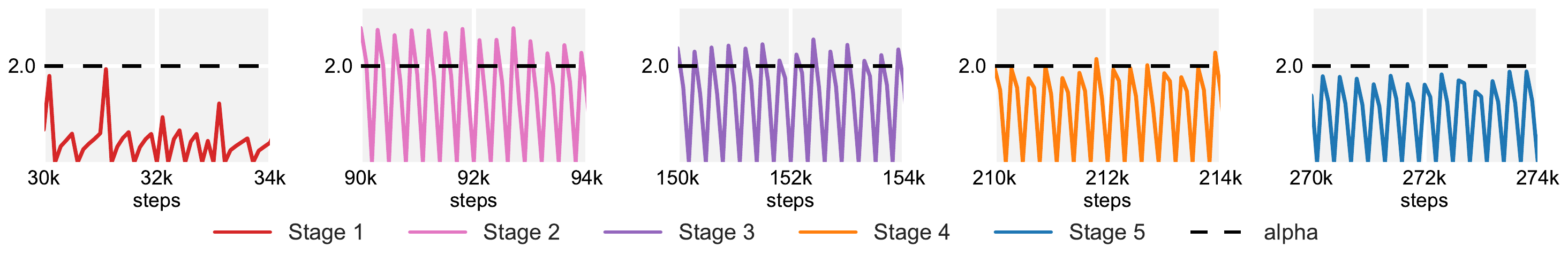}
            \caption{Ant}
        \end{subfigure}%
        \caption{Visualization of our event-triggered mechanism on HalfCheetah and Ant benchmarks. Solid lines show the model shifts estimation and dotted lines reveals the threshold of triggered condition. Note that here we apply log value.}
        \label{fig:event}
    \end{minipage}
    \vspace{-5pt}
\end{figure}

\paragraph{The necessity of event-triggered mechanism.}
To verify the necessity of event-triggered mechanism, we compare to three unconstrained cases (given fixed model training interval) under two environments. 
The training curve and triggered times are shown in Figure~\ref{fig:main-abl}. 
Clearly,  our mechanism improves the performance while reducing the total times of model training.
Besides, we notice that the performance is comparable to other MBRL baselines (MBPO \emph{etc.}) when fixing our model training interval at 250. Still it  performs worse than that equipped with a smart mechanism to decide whether to train the model at current exploration step. 

To better understand why the event-triggered mechanism brings up our outperformance, we asses its main bricks. Model shift, which reflects the current ability to digest new data, is the basis of triggered condition. And we estimate model shift from two parts, policy coverage and prediction error. In Figure~\ref{fig:event-ele} we observe that, gradually, as the training progresses, the policy coverage increases, which reflects our policy has new explorations at every stage without falling into a local optimum prematurely. 
Also, the prediction error gradually decreases, which implies that our estimated dynamics come closer to the true dynamics in the explored region. Figure~\ref{fig:event} implies that the number of samples required to hit the constraint tends to grows and the model training frequency then goes down. 
This is also consistent with our intuition that, as the model refines and the exploration novelty fades, then the sample size required to reach a certain level of model shifts grows up. The result agrees with our theoretical analyses which describe that a dynamic alternation subject to the model shifts constraint does benefit to monotonicity rather than an assigned one.

\begin{wrapfigure}[20]{r}{0.48\textwidth}
    \vspace{-15pt}
     \begin{minipage}{\linewidth}
        \begin{subfigure}[t]{1.0\textwidth}
            \centering
            \includegraphics[width=0.49\linewidth]{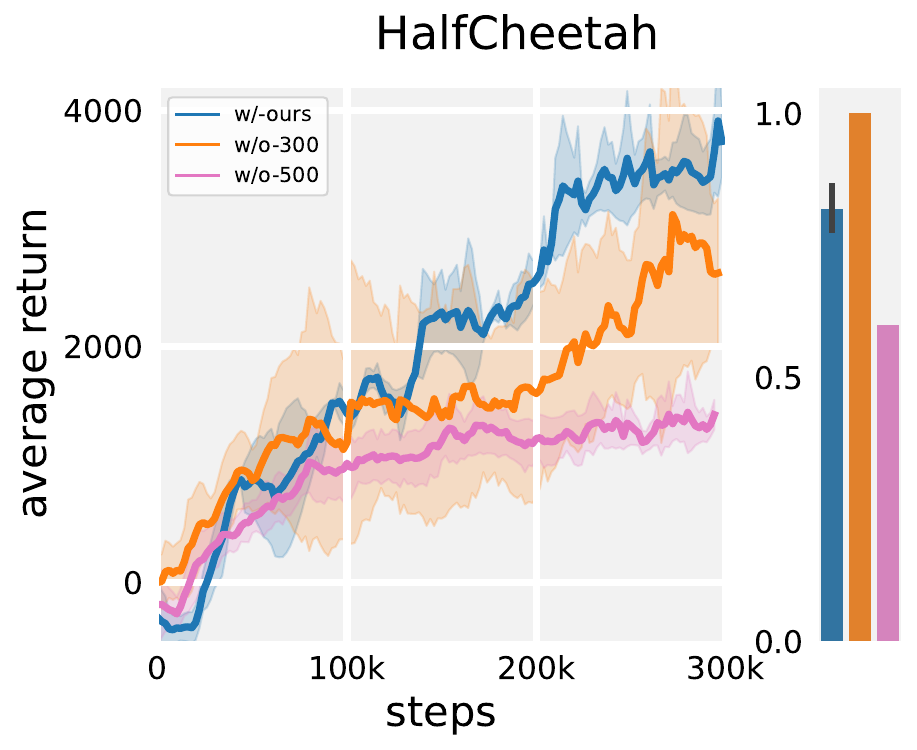}
            \includegraphics[width=0.49\linewidth]{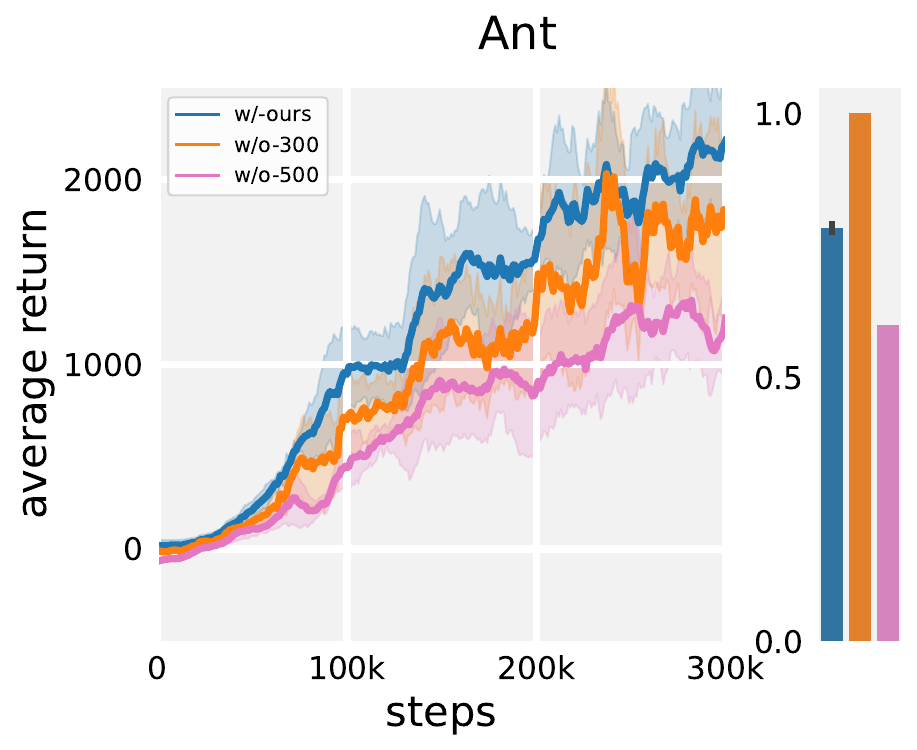} 
            \caption{policy optimization oracle: TRPO}
            \label{fig:trpo}
         \end{subfigure} 
         \hfill
     \begin{subfigure}[t]{1.0\textwidth}
        \centering
        \includegraphics[width=0.49\linewidth]{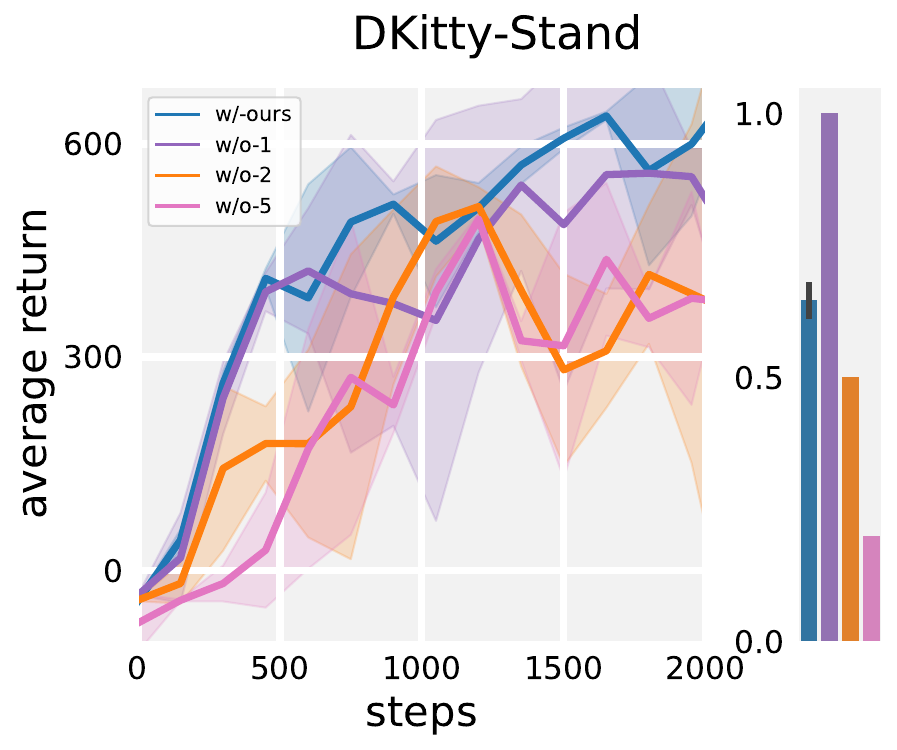} 
        \includegraphics[width=0.49\linewidth]{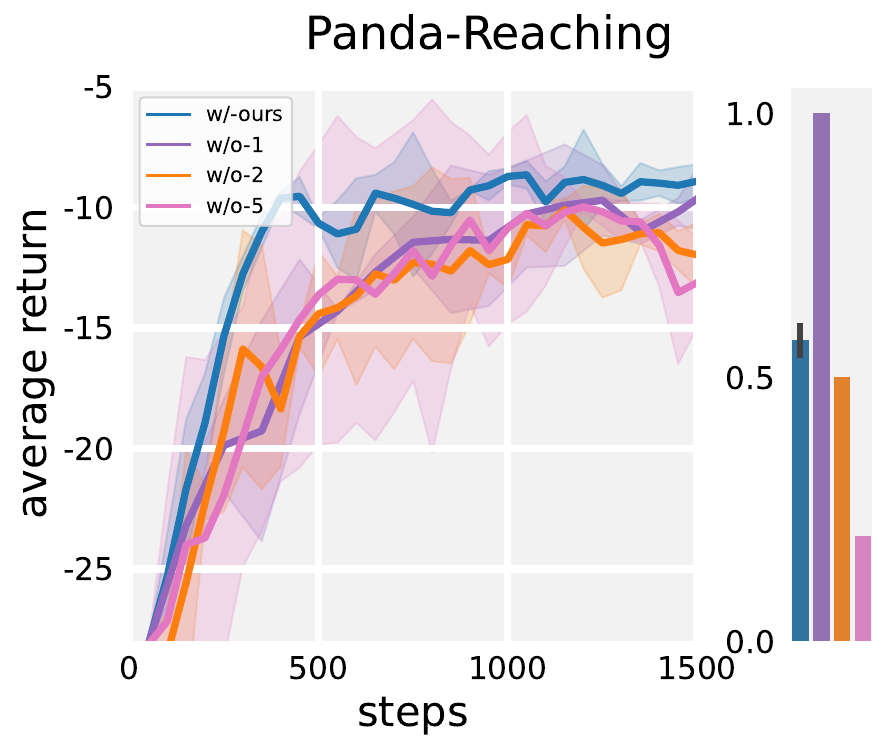} 
        \caption{policy optimization oracle: iLQR}
        \label{fig:trpo}
     \end{subfigure}
    \end{minipage}%
    \caption{Ablation on generalizability via various policy optimization oracles. Line plots reflect the average return and the standard deviation over 5 trails. Bar plots indicate the total triggered times and the y-axis is scaled to [0,1].}
     \vspace{-5pt}
    \label{fig:policyoracle}
\end{wrapfigure}

\paragraph{The generalizability of event-triggered mechanism.}
We further investigate the generalizability of our proposed mechanism through ablation on policy optimization oracle, and results are shown in Figure~\ref{fig:policyoracle}. 
\paragraph{1) under Dyna-style.} We adopt TRPO~\cite{schulman2015trust} as the policy optimization oracle and test the performance with or without event-triggered mechanism in Halfcheetah and Ant benchmarks. Observably, our mechanism can effectively guarantee the overall performance improvement and alleviate the local optimization issue.
\paragraph{2) jumping off Dyna-style.} We utilize iLQR~\cite{amos2018differentiable} and then conduct ablation experiments in the DKitty-Stand~\cite{ahn2020robel} and Panda-Reaching~\cite{james2019pyrep} scenarios. The result shows that adding event-triggered mechanism has a comparable asymptotic performance as the most-frequently triggered case but enjoys a lower computation cost.

\section{Conclusion}\label{conclusion}
We have investigated the role of the decision on "when to update the model"  in joint optimization procedures through theoretical and empirical lens.  
We devise a general novel scheme for exploring the monotonicity of MBRL methods, distinguished from the existing discrepancy bound scheme.
We then derive lower bounds under the scheme, suggesting that models with higher ceiling performance and lower bias guarantee a non-decreasing performance evaluated in the real environment.  
An effective constrained optimization problem comes from 
the follow-up refined bound to seek a non-negative lower bound.
Further, the instance under the generative model setting further verifies the effectiveness of learning models from a dynamically varying number of explorations. 
The algorithm CMLO, stemming from these analyses, has asymptotic performance rivaling the best model-free algorithms and boasts better monotonicity. 
Further ablation studies reveal that the proposed mechanism scales to various policy optimization oracles and benefits computation cost reduction. 
Currently, we observed empirically that the event-triggered condition is usually related to specific environments, which will cost a little time for tuning. 
Thus, one direction that merits further investigation is to construct the dual problem of our constrained optimization problem for better exploring optimality and monotonicity.

\begin{ack}
  This work was jointly supported by the Sino-German Collaborative Research Project "Crossmodal Learning" (NSFC  62061136001/ DFG TRR169), the CAS Project for Young Scientists in Basic Research (Grant No.YSBR-040), the National Natural Science Foundation of China (No.62006137) and Beijing Outstanding Young Scientist Program (No.BJJWZYJH012019100020098). The authors would also like to thank the anonymous reviewers for their careful reading and their many insightful comments.
\end{ack}

\bibliographystyle{plainnat}
\bibliography{IEEEabrv,ref}

\begin{thebibliography}{57}
\providecommand{\natexlab}[1]{#1}
\providecommand{\url}[1]{\texttt{#1}}
\expandafter\ifx\csname urlstyle\endcsname\relax
  \providecommand{\doi}[1]{doi: #1}\else
  \providecommand{\doi}{doi: \begingroup \urlstyle{rm}\Url}\fi

\bibitem[Agarwal et~al.(2020)Agarwal, Kakade, and Yang]{agarwal2020model}
Alekh Agarwal, Sham Kakade, and Lin~F Yang.
\newblock Model-based reinforcement learning with a generative model is minimax
  optimal.
\newblock In \emph{Conference on Learning Theory}, 2020.

\bibitem[Ahn et~al.(2020)Ahn, Zhu, Hartikainen, Ponte, Gupta, Levine, and
  Kumar]{ahn2020robel}
Michael Ahn, Henry Zhu, Kristian Hartikainen, Hugo Ponte, Abhishek Gupta,
  Sergey Levine, and Vikash Kumar.
\newblock Robel: Robotics benchmarks for learning with low-cost robots.
\newblock In \emph{Conference on robot learning}, 2020.

\bibitem[Amos et~al.(2018)Amos, Jimenez, Sacks, Boots, and
  Kolter]{amos2018differentiable}
Brandon Amos, Ivan Jimenez, Jacob Sacks, Byron Boots, and J~Zico Kolter.
\newblock Differentiable mpc for end-to-end planning and control.
\newblock In \emph{Advances in Neural Information Processing Systems}, 2018.

\bibitem[Asadi et~al.(2018)Asadi, Misra, and Littman]{asadi2018lipschitz}
Kavosh Asadi, Dipendra Misra, and Michael Littman.
\newblock Lipschitz continuity in model-based reinforcement learning.
\newblock In \emph{International Conference on Machine Learning}, 2018.

\bibitem[Azar et~al.(2013)Azar, Munos, and Kappen]{azar2013minimax}
Mohammad~Gheshlaghi Azar, R{\'e}mi Munos, and Hilbert Kappen.
\newblock Minimax pac bounds on the sample complexity of reinforcement learning
  with a generative model.
\newblock \emph{Machine Learning}, 91\penalty0 (3):\penalty0 325--349, 2013.

\bibitem[Brockman et~al.(2016)Brockman, Cheung, Pettersson, Schneider,
  Schulman, Tang, and Zaremba]{brockman2016openai}
Greg Brockman, Vicki Cheung, Ludwig Pettersson, Jonas Schneider, John Schulman,
  Jie Tang, and Wojciech Zaremba.
\newblock Openai gym.
\newblock \emph{arXiv preprint arXiv:1606.01540}, 2016.

\bibitem[Buckman et~al.()Buckman, Hafner, Tucker, Brevdo, and
  Lee]{buckman2018sample}
Jacob Buckman, Danijar Hafner, George Tucker, Eugene Brevdo, and Honglak Lee.
\newblock Sample-efficient reinforcement learning with stochastic ensemble
  value expansion.
\newblock In \emph{Advances in Neural Information Processing Systems}.

\bibitem[Chua et~al.()Chua, Calandra, McAllister, and Levine]{chua2018deep}
Kurtland Chua, Roberto Calandra, Rowan McAllister, and Sergey Levine.
\newblock Deep reinforcement learning in a handful of trials using
  probabilistic dynamics models.
\newblock In \emph{Advances in Neural Information Processing Systems}.

\bibitem[Deisenroth et~al.(2011)Deisenroth, Rasmussen, and
  Fox]{deisenroth2011learning}
Marc~Peter Deisenroth, Carl~Edward Rasmussen, and Dieter Fox.
\newblock Learning to control a low-cost manipulator using data-efficient
  reinforcement learning.
\newblock \emph{Robotics: Science and Systems VII}, 7:\penalty0 57--64, 2011.

\bibitem[Efroni et~al.()Efroni, Merlis, Ghavamzadeh, and
  Mannor]{efroni2019tight}
Yonathan Efroni, Nadav Merlis, Mohammad Ghavamzadeh, and Shie Mannor.
\newblock Tight regret bounds for model-based reinforcement learning with
  greedy policies.
\newblock In \emph{Advances in Neural Information Processing Systems}.

\bibitem[Fan and Ming(2021)]{fan2021model}
Ying Fan and Yifei Ming.
\newblock Model-based reinforcement learning for continuous control with
  posterior sampling.
\newblock In \emph{International Conference on Machine Learning}, 2021.

\bibitem[Farahmand et~al.(2017)Farahmand, Barreto, and
  Nikovski]{farahmand2017value}
Amir-massoud Farahmand, Andre Barreto, and Daniel Nikovski.
\newblock Value-aware loss function for model-based reinforcement learning.
\newblock In \emph{Artificial Intelligence and Statistics}, 2017.

\bibitem[Feinberg et~al.(2018)Feinberg, Wan, Stoica, Jordan, Gonzalez, and
  Levine]{feinberg2018model}
Vladimir Feinberg, Alvin Wan, Ion Stoica, Michael~I Jordan, Joseph~E Gonzalez,
  and Sergey Levine.
\newblock Model-based value estimation for efficient model-free reinforcement
  learning.
\newblock \emph{arXiv preprint arXiv:1803.00101}, 2018.

\bibitem[Fujimoto et~al.(2018)Fujimoto, Hoof, and
  Meger]{fujimoto2018addressing}
Scott Fujimoto, Herke Hoof, and David Meger.
\newblock Addressing function approximation error in actor-critic methods.
\newblock In \emph{International Conference on Machine Learning}, 2018.

\bibitem[Gal et~al.(2016)Gal, McAllister, and Rasmussen]{gal2016improving}
Yarin Gal, Rowan McAllister, and Carl~Edward Rasmussen.
\newblock Improving pilco with bayesian neural network dynamics models.
\newblock In \emph{ICML Workshop on Data-Efﬁcient Machine Learning Workshop},
  2016.

\bibitem[Haarnoja et~al.(2018)Haarnoja, Zhou, Abbeel, and
  Levine]{haarnoja2018soft}
Tuomas Haarnoja, Aurick Zhou, Pieter Abbeel, and Sergey Levine.
\newblock Soft actor-critic: Off-policy maximum entropy deep reinforcement
  learning with a stochastic actor.
\newblock In \emph{International Conference on Machine Learning}, 2018.

\bibitem[Heemels et~al.(2012)Heemels, Johansson, and
  Tabuada]{heemels2012introduction}
Wilhelmus~PMH Heemels, Karl~Henrik Johansson, and Paulo Tabuada.
\newblock An introduction to event-triggered and self-triggered control.
\newblock In \emph{2012 ieee 51st ieee conference on decision and control
  (cdc)}, pages 3270--3285. IEEE, 2012.

\bibitem[Hester et~al.(2012)Hester, Quinlan, and Stone]{hester2012rtmba}
Todd Hester, Michael Quinlan, and Peter Stone.
\newblock Rtmba: A real-time model-based reinforcement learning architecture
  for robot control.
\newblock In \emph{International Conference on Robotics and Automation}, 2012.

\bibitem[James et~al.(2019)James, Freese, and Davison]{james2019pyrep}
Stephen James, Marc Freese, and Andrew~J. Davison.
\newblock Pyrep: Bringing v-rep to deep robot learning.
\newblock \emph{arXiv preprint arXiv:1906.11176}, 2019.

\bibitem[Janner et~al.(2019)Janner, Fu, Zhang, and Levine]{janner2019trust}
Michael Janner, Justin Fu, Marvin Zhang, and Sergey Levine.
\newblock When to trust your model: Model-based policy optimization.
\newblock In \emph{Advances in Neural Information Processing Systems}, 2019.

\bibitem[Kaelbling et~al.(1996)Kaelbling, Littman, and
  Moore]{kaelbling1996reinforcement}
Leslie~Pack Kaelbling, Michael~L Littman, and Andrew~W Moore.
\newblock Reinforcement learning: A survey.
\newblock \emph{Journal of artificial intelligence research}, 4:\penalty0
  237--285, 1996.

\bibitem[Kakade and Langford(2002)]{kakade2002approximately}
Sham Kakade and John Langford.
\newblock Approximately optimal approximate reinforcement learning.
\newblock In \emph{International Conference on Machine Learning}, 2002.

\bibitem[Kakade et~al.(2020)Kakade, Krishnamurthy, Lowrey, Ohnishi, and
  Sun]{kakade2020information}
Sham Kakade, Akshay Krishnamurthy, Kendall Lowrey, Motoya Ohnishi, and Wen Sun.
\newblock Information theoretic regret bounds for online nonlinear control.
\newblock In \emph{Advances in Neural Information Processing Systems}, 2020.

\bibitem[Kalweit and Boedecker(2017)]{kalweit2017uncertainty}
Gabriel Kalweit and Joschka Boedecker.
\newblock Uncertainty-driven imagination for continuous deep reinforcement
  learning.
\newblock In \emph{Conference on Robot Learning}, 2017.

\bibitem[Kearns et~al.(2002)Kearns, Mansour, and Ng]{kearns2002sparse}
Michael Kearns, Yishay Mansour, and Andrew~Y Ng.
\newblock A sparse sampling algorithm for near-optimal planning in large markov
  decision processes.
\newblock \emph{Machine learning}, 49\penalty0 (2):\penalty0 193--208, 2002.

\bibitem[Ko and Fox(2009)]{ko2009gp}
Jonathan Ko and Dieter Fox.
\newblock Gp-bayesfilters: Bayesian filtering using gaussian process prediction
  and observation models.
\newblock \emph{Autonomous Robots}, 27\penalty0 (1):\penalty0 75--90, 2009.

\bibitem[Kurutach et~al.(2018)Kurutach, Clavera, Duan, Tamar, and
  Abbeel]{kurutach2018model}
Thanard Kurutach, Ignasi Clavera, Yan Duan, Aviv Tamar, and Pieter Abbeel.
\newblock Model-ensemble trust-region policy optimization.
\newblock In \emph{International Conference on Learning Representations}, 2018.

\bibitem[Lai et~al.(2021)Lai, Shen, Zhang, Huang, Zhang, Tang, Yu, and
  Li]{lai2021effective}
Hang Lai, Jian Shen, Weinan Zhang, Yimin Huang, Xing Zhang, Ruiming Tang, Yong
  Yu, and Zhenguo Li.
\newblock On effective scheduling of model-based reinforcement learning.
\newblock In \emph{Advances in Neural Information Processing Systems}, 2021.

\bibitem[Levine and Koltun(2013)]{levine2013guided}
Sergey Levine and Vladlen Koltun.
\newblock Guided policy search.
\newblock In \emph{International Conference on Machine Learning}, 2013.

\bibitem[Levine et~al.(2016)Levine, Finn, Darrell, and Abbeel]{levine2016end}
Sergey Levine, Chelsea Finn, Trevor Darrell, and Pieter Abbeel.
\newblock End-to-end training of deep visuomotor policies.
\newblock \emph{The Journal of Machine Learning Research}, 17\penalty0
  (1):\penalty0 1334--1373, 2016.

\bibitem[Li et~al.(2020)Li, Wei, Chi, Gu, and Chen]{li2020breaking}
Gen Li, Yuting Wei, Yuejie Chi, Yuantao Gu, and Yuxin Chen.
\newblock Breaking the sample size barrier in model-based reinforcement
  learning with a generative model.
\newblock In \emph{Advances in Neural Information Processing Systems}, 2020.

\bibitem[Li and Shi(2014)]{li2014event}
Huiping Li and Yang Shi.
\newblock Event-triggered robust model predictive control of continuous-time
  nonlinear systems.
\newblock \emph{Automatica}, 50\penalty0 (5):\penalty0 1507--1513, 2014.

\bibitem[Lillicrap et~al.(2016)Lillicrap, Hunt, Pritzel, Heess, Erez, Tassa,
  Silver, and Wierstra]{lillicrap2016continuous}
Timothy~P Lillicrap, Jonathan~J Hunt, Alexander Pritzel, Nicolas Heess, Tom
  Erez, Yuval Tassa, David Silver, and Daan Wierstra.
\newblock Continuous control with deep reinforcement learning.
\newblock In \emph{International Conference on Learning Representations}, 2016.

\bibitem[Luo et~al.(2018)Luo, Xu, Li, Tian, Darrell, and
  Ma]{luo2018algorithmic}
Yuping Luo, Huazhe Xu, Yuanzhi Li, Yuandong Tian, Trevor Darrell, and Tengyu
  Ma.
\newblock Algorithmic framework for model-based deep reinforcement learning
  with theoretical guarantees.
\newblock In \emph{International Conference on Learning Representations}, 2018.

\bibitem[Mnih et~al.(2013)Mnih, Kavukcuoglu, Silver, Graves, Antonoglou,
  Wierstra, and Riedmiller]{atari2013}
Volodymyr Mnih, Koray Kavukcuoglu, David Silver, Alex Graves, Ioannis
  Antonoglou, Daan Wierstra, and Martin Riedmiller.
\newblock Playing atari with deep reinforcement learning.
\newblock In \emph{Advances in Neural Information Processing Systems}, 2013.

\bibitem[Mnih et~al.(2015)Mnih, Kavukcuoglu, Silver, Rusu, Veness, Bellemare,
  Graves, Riedmiller, Fidjeland, Ostrovski, et~al.]{mnih2015human}
Volodymyr Mnih, Koray Kavukcuoglu, David Silver, Andrei~A Rusu, Joel Veness,
  Marc~G Bellemare, Alex Graves, Martin Riedmiller, Andreas~K Fidjeland, Georg
  Ostrovski, et~al.
\newblock Human-level control through deep reinforcement learning.
\newblock \emph{nature}, 518\penalty0 (7540):\penalty0 529--533, 2015.

\bibitem[Morimoto and Atkeson(2002)]{morimoto2002minimax}
Jun Morimoto and Christopher Atkeson.
\newblock Minimax differential dynamic programming: An application to robust
  biped walking.
\newblock In \emph{Advances in Neural Information Processing Systems}, 2002.

\bibitem[Nagabandi et~al.(2018)Nagabandi, Kahn, Fearing, and
  Levine]{nagabandi2018neural}
Anusha Nagabandi, Gregory Kahn, Ronald~S Fearing, and Sergey Levine.
\newblock Neural network dynamics for model-based deep reinforcement learning
  with model-free fine-tuning.
\newblock In \emph{International Conference on Robotics and Automation}, 2018.

\bibitem[Pan et~al.()Pan, He, Tu, and He]{pan2020trust}
Feiyang Pan, Jia He, Dandan Tu, and Qing He.
\newblock Trust the model when it is confident: Masked model-based
  actor-critic.
\newblock In \emph{Advances in Neural Information Processing Systems}.

\bibitem[Pineda et~al.(2021)Pineda, Amos, Zhang, Lambert, and
  Calandra]{Pineda2021MBRL}
Luis Pineda, Brandon Amos, Amy Zhang, Nathan~O. Lambert, and Roberto Calandra.
\newblock Mbrl-lib: A modular library for model-based reinforcement learning.
\newblock \emph{Arxiv}, 2021.
\newblock URL \url{https://arxiv.org/abs/2104.10159}.

\bibitem[Polydoros and Nalpantidis(2017)]{polydoros2017survey}
Athanasios~S Polydoros and Lazaros Nalpantidis.
\newblock Survey of model-based reinforcement learning: Applications on
  robotics.
\newblock \emph{Journal of Intelligent \& Robotic Systems}, 86\penalty0
  (2):\penalty0 153--173, 2017.

\bibitem[pranz24(2018)]{saclib}
pranz24.
\newblock pytorch-soft-actor-critic.
\newblock \url{https://github.com/pranz24/pytorch-soft-actor-critic}, 2018.

\bibitem[Puterman(2014)]{puterman2014markov}
Martin~L Puterman.
\newblock \emph{Markov decision processes: discrete stochastic dynamic
  programming}.
\newblock John Wiley \& Sons, 2014.

\bibitem[Schulman et~al.(2015{\natexlab{a}})Schulman, Levine, Abbeel, Jordan,
  and Moritz]{schulman2015trust}
John Schulman, Sergey Levine, Pieter Abbeel, Michael Jordan, and Philipp
  Moritz.
\newblock Trust region policy optimization.
\newblock In \emph{International Conference on Machine Learning},
  2015{\natexlab{a}}.

\bibitem[Schulman et~al.(2015{\natexlab{b}})Schulman, Moritz, Levine, Jordan,
  and Abbeel]{schulman2015high}
John Schulman, Philipp Moritz, Sergey Levine, Michael Jordan, and Pieter
  Abbeel.
\newblock High-dimensional continuous control using generalized advantage
  estimation.
\newblock \emph{arXiv preprint arXiv:1506.02438}, 2015{\natexlab{b}}.

\bibitem[Schulman et~al.(2017)Schulman, Wolski, Dhariwal, Radford, and
  Klimov]{schulman2017proximal}
John Schulman, Filip Wolski, Prafulla Dhariwal, Alec Radford, and Oleg Klimov.
\newblock Proximal policy optimization algorithms.
\newblock \emph{arXiv preprint arXiv:1707.06347}, 2017.

\bibitem[Sidford et~al.(2018)Sidford, Wang, Wu, Yang, and Ye]{sidford2018near}
Aaron Sidford, Mengdi Wang, Xian Wu, Lin~F Yang, and Yinyu Ye.
\newblock Near-optimal time and sample complexities for solving discounted
  markov decision process with a generative model.
\newblock \emph{arXiv preprint arXiv:1806.01492}, 2018.

\bibitem[Silver et~al.(2016)Silver, Huang, Maddison, Guez, Sifre, Van
  Den~Driessche, Schrittwieser, Antonoglou, Panneershelvam, Lanctot,
  et~al.]{silver2016mastering}
David Silver, Aja Huang, Chris~J Maddison, Arthur Guez, Laurent Sifre, George
  Van Den~Driessche, Julian Schrittwieser, Ioannis Antonoglou, Veda
  Panneershelvam, Marc Lanctot, et~al.
\newblock Mastering the game of go with deep neural networks and tree search.
\newblock \emph{nature}, 529\penalty0 (7587):\penalty0 484--489, 2016.

\bibitem[Sun et~al.()Sun, Gordon, Boots, and Bagnell]{sun2018dual}
Wen Sun, Geoffrey~J Gordon, Byron Boots, and J~Bagnell.
\newblock Dual policy iteration.
\newblock In \emph{Advances in Neural Information Processing Systems}.

\bibitem[Sutton(1990)]{sutton1990integrated}
Richard~S Sutton.
\newblock Integrated architecture for learning, planning, and reacting based on
  approximating dynamic programming.
\newblock In \emph{International Conference on Machine Learning}, 1990.

\bibitem[Sutton(1991{\natexlab{a}})]{sutton1991dyna}
Richard~S Sutton.
\newblock Dyna, an integrated architecture for learning, planning, and
  reacting.
\newblock \emph{ACM Sigart Bulletin}, 2\penalty0 (4):\penalty0 160--163,
  1991{\natexlab{a}}.

\bibitem[Sutton(1991{\natexlab{b}})]{sutton1991planning}
Richard~S Sutton.
\newblock Planning by incremental dynamic programming.
\newblock In \emph{International Conference on Machine Learning},
  1991{\natexlab{b}}.

\bibitem[Sutton and Barto(2018)]{sutton2018reinforcement}
Richard~S Sutton and Andrew~G Barto.
\newblock \emph{Reinforcement learning: An introduction}.
\newblock MIT press, Cambridge, MA, 2018.

\bibitem[Todorov et~al.(2012)Todorov, Erez, and Tassa]{todorov2012mujoco}
Emanuel Todorov, Tom Erez, and Yuval Tassa.
\newblock Mujoco: A physics engine for model-based control.
\newblock In \emph{International Conference on Intelligent Robots and Systems},
  2012.

\bibitem[Wang et~al.(2019)Wang, Bao, Clavera, Hoang, Wen, Langlois, Zhang,
  Zhang, Abbeel, and Ba]{wang2019benchmarking}
Tingwu Wang, Xuchan Bao, Ignasi Clavera, Jerrick Hoang, Yeming Wen, Eric
  Langlois, Shunshi Zhang, Guodong Zhang, Pieter Abbeel, and Jimmy Ba.
\newblock Benchmarking model-based reinforcement learning.
\newblock \emph{arXiv preprint arXiv:1907.02057}, 2019.

\bibitem[Weissman et~al.(2003)Weissman, Ordentlich, Seroussi, Verdu, and
  Weinberger]{weissman2003inequalities}
Tsachy Weissman, Erik Ordentlich, Gadiel Seroussi, Sergio Verdu, and Marcelo~J
  Weinberger.
\newblock Inequalities for the l1 deviation of the empirical distribution.
\newblock \emph{Hewlett-Packard Labs, Tech. Rep}, 2003.

\bibitem[Yu et~al.(2020)Yu, Thomas, Yu, Ermon, Zou, Levine, Finn, and
  Ma]{yu2020mopo}
Tianhe Yu, Garrett Thomas, Lantao Yu, Stefano Ermon, James~Y Zou, Sergey
  Levine, Chelsea Finn, and Tengyu Ma.
\newblock Mopo: Model-based offline policy optimization.
\newblock In \emph{Advances in Neural Information Processing Systems}, 2020.

\end{thebibliography}

\newpage
\section*{Checklist}

\begin{enumerate}

\item For all authors...
\begin{enumerate}
  \item Do the main claims made in the abstract and introduction accurately reflect the paper's contributions and scope?
    \answerYes{}
  \item Did you describe the limitations of your work?
    \answerYes{See Section \ref{conclusion}.}
  \item Did you discuss any potential negative societal impacts of your work?
    \answerNo{No potential negative societal impacts have been found yet.}
  \item Have you read the ethics review guidelines and ensured that your paper conforms to them?
    \answerYes{}
\end{enumerate}

\item If you are including theoretical results...
\begin{enumerate}
  \item Did you state the full set of assumptions of all theoretical results?
    \answerYes{See Section \ref{Methods} and Appendix \ref{ap-omittedproofs}.}
        \item Did you include complete proofs of all theoretical results?
    \answerYes{See Appendix \ref{ap-omittedproofs} and Appendix \ref{toolbox}.}
\end{enumerate}

\item If you ran experiments...
\begin{enumerate}
  \item Did you include the code, data, and instructions needed to reproduce the main experimental results (either in the supplemental material or as a URL)?
    \answerYes{In the supplemental material.}
  \item Did you specify all the training details (e.g., data splits, hyperparameters, how they were chosen)?
    \answerYes{See Appendix \ref{implementation} and our provided code.}
        \item Did you report error bars (e.g., with respect to the random seed after running experiments multiple times)?
    \answerYes{See in corresponding figure captions.}
        \item Did you include the total amount of compute and the type of resources used (e.g., type of GPUs, internal cluster, or cloud provider)?
    \answerYes{See Appendix \ref{implementation}.}
\end{enumerate}

\item If you are using existing assets (e.g., code, data, models) or curating/releasing new assets...
\begin{enumerate}
  \item If your work uses existing assets, did you cite the creators?
    \answerYes{See Appendix \ref{implementation}}
  \item Did you mention the license of the assets?
    \answerYes{See Appendix \ref{implementation}}
  \item Did you include any new assets either in the supplemental material or as a URL?
    \answerNA{}
  \item Did you discuss whether and how consent was obtained from people whose data you're using/curating?
    \answerNA{}
  \item Did you discuss whether the data you are using/curating contains personally identifiable information or offensive content?
    \answerNA{Our work does not use existing data.}
\end{enumerate}

\item If you used crowdsourcing or conducted research with human subjects...
\begin{enumerate}
  \item Did you include the full text of instructions given to participants and screenshots, if applicable?
    \answerNA{No crowdsourcing/human subjects are used.}
  \item Did you describe any potential participant risks, with links to Institutional Review Board (IRB) approvals, if applicable?
    \answerNA{No crowdsourcing/human subjects are used.}
  \item Did you include the estimated hourly wage paid to participants and the total amount spent on participant compensation?
    \answerNA{No crowdsourcing/human subjects are used.}
\end{enumerate}

\end{enumerate}
\newpage
\appendix
{\Large \textbf{Appendices}}

\section{Sketch of Theoretical Analyses}\label{sketch}
Here, we present our sketch of theoretical analyses. 
We first construct a general scheme (Definition \ref{definition}) for non-decreasing performance guarantee and follow it up by characterizing the lower bound under model shifts (Theorem \ref{thm: general-bound}). Towards seeking a non-negative lower bound, we  restrict model shift and refine the bound (Theorem \ref{refine-bound}), then further reduce this issue to a constrained optimization problem (Proposition \ref{simple-opt}). With an instance under the generative model setting (Corollary \ref{corollary-k}), we demonstrate the merits of the dynamic model learning interval. 
\begin{figure}[h]
    \centering
    \includegraphics[width=1.0\textwidth]{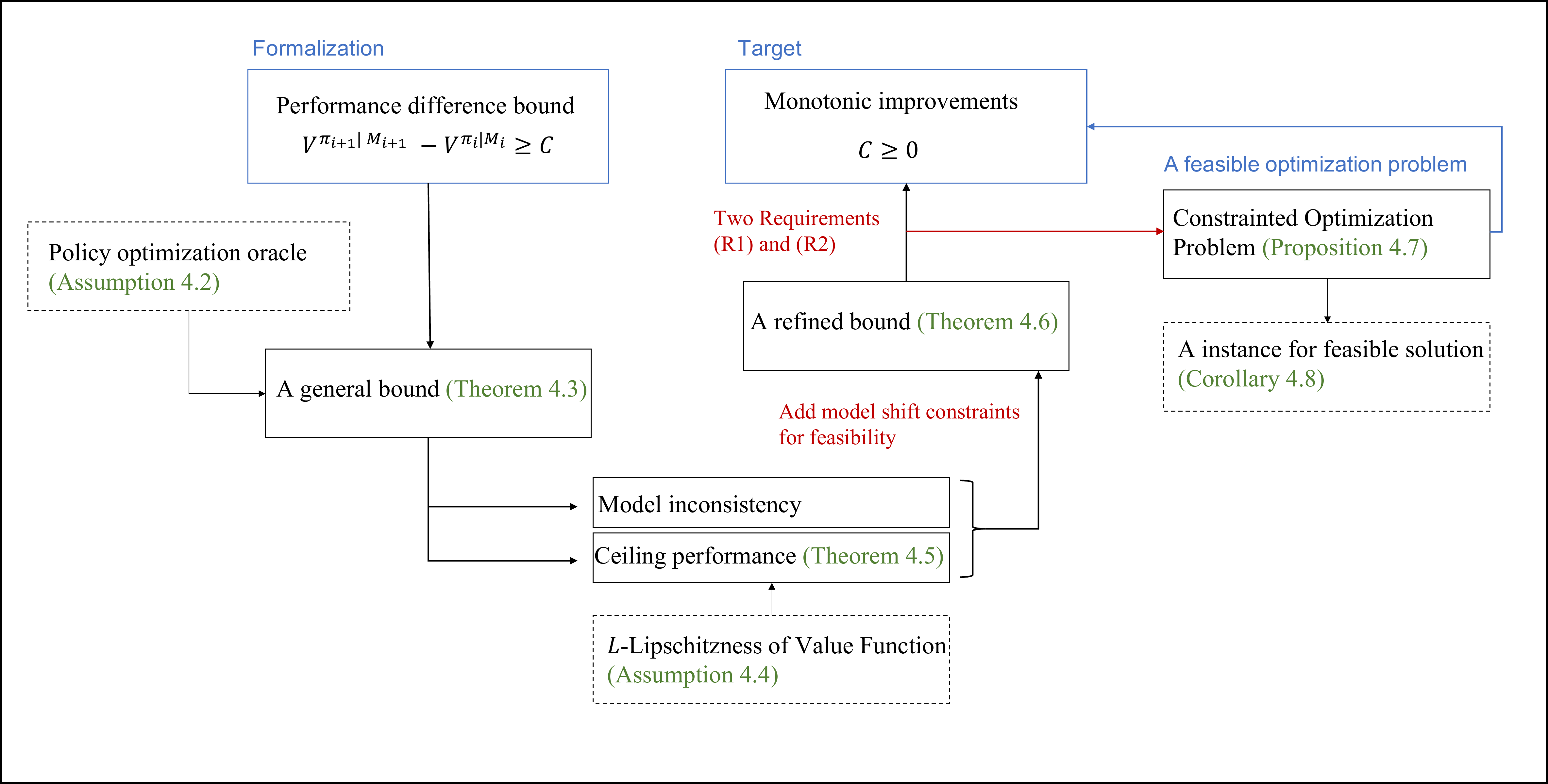}
    \caption{Theoretical sketch of CMLO }
    \label{fig:theoryrecipe}
\end{figure}

\section{Omitted Proofs} \label{ap-omittedproofs}

\begin{theorem}[Performance difference bound for Model-based RL]\label{proof-general-gap}
Let $\epsilon_{M_i}^{\pi_j}$ denote the inconsistency between the learned dynamics $P_{M_i}$ and the true dynamics, i.e. $\epsilon_{M_i}^{\pi_j} = \mathbb{E}_{s,a\sim d^{\pi_j}(s,a;\mu)} [\TV(P(\cdot\vert s,a)\Vert  P_{M_i}(\cdot\vert s,a))]$, where $d^\pi(s,a;\mu)$  is the probability of visiting $s, a$ after starting at state $s_0 \sim \mu$ and following $\pi_j \in \Pi$ thereafter under the true dynamics.  
Let policy $\pi_i$ be the $\epsilon_{opt}$-optimal policy under model $M_i$.  Assume the performance discrepancy of policy $\pi$ between the estimated model ${M_1}$and the true dynamics be approximated as $V_{M_1}^{\pi_1}(\mu) - V^{\pi_1}(\mu)  = \kappa \cdot \epsilon_{M_1}^{\pi_1} $. Recall $\kappa=\frac{2R\gamma}{(1-\gamma)^2}$, then the performance gap between ${\pi_1}$ and ${\pi_2}$ evaluated in the true MDP can be bounded by:
\begin{equation*}
    V^{\pi_2\vert M_2} - V^{\pi_1\vert M_1} \geq - \kappa \cdot (\epsilon_{M_2}^{\pi_2} - \epsilon_{M_1}^{\pi_1}) + V_{M_2}^* - V_{M_1}^*  - \epsilon_{opt}
\end{equation*}
\end{theorem}

\begin{proof}
We overload notation $V^{\pi_i\vert M_i}$ and write $V^{\pi_i}$ for simplicity.
 \begin{equation*}
    \begin{aligned}
V^{\pi_2} - V^{\pi_1} &= \underbrace{V^{\pi_2}-V^{\pi_2}_{M_2}}_{L_1} +\underbrace{V^{\pi_2}_{M_2}-V^{\pi_1}_{M_1}}_{L_2} - \underbrace{(V^{\pi_1}-V^{\pi_1}_{M_1})}_{L_3}\\
\end{aligned}
\end{equation*}

We can bound $L_1-L_3$ using Lemma~\ref*{lemma:dis-bound}, and bound $L_2$ using the property of $\epsilon_{opt}$-optimal.

For $L_1-L_3$, with the performance gap approximation of $M_1$ and $\pi_1$, we apply Lemma~\ref*{lemma:dis-bound}, and obtain: $L_1 -L_3\geq -\kappa \cdot (\epsilon_{M_2}^{\pi_2} - \epsilon_{M_1}^{\pi_1})$.

We call a policy $\pi$ $\epsilon$-optimal under the dynamical model $M$, if $V_M^*(s)\geq  V^\pi_M(s)\geq V_M^*(s)-\epsilon_{opt}$ for all $s\in {\cal S}$.  Under the assumptions of black-box optimization oracle, we obtain:
\begin{equation*}
    L_2\geq V_{M_2}^* - V_{M_1}^* - \epsilon_{opt}
\end{equation*}
Adding these two bounds together yields the desired result.

Remark that, when $V_{M_1}^{\pi_1} - V^{\pi_1} $ gets a value far away from $\kappa\cdot \epsilon_{M_1}^{\pi_1}$,  it indicates the performance discrepancy that evaluated between the model ${M_1}$ and the environment is near to zero, which further indicates that the inconsistency between $P_{M_1}$ and the true dynamics is quite small and thus the optimization process has reached a stopping point.
\end{proof}

\begin{theorem}[Ceiling return gap Under model shifts]\label{ap:ceiling-bound}
For a dynamical model $M_i \in \gM$,  $V_{M_i}^*(\mu)$  denotes the maximal returns on dynamics $P_{M_i}$. Then the gap of optimal returns under these two models $M_1$, $M_2$ can be bounded as:
\begin{equation*}
    V_{M_2}^* - V_{M_1}^*\geq - \frac{\gamma}{1-\gamma}L\cdot \sup_{\pi\in \Pi} \mathbb{E}_{s,a\sim d^\pi_{M_2}} \Big[\vert  P_{M_2}(\cdot\vert s,a)- P_{M_1}(\cdot\vert s,a)\vert\Big]
\end{equation*}
\end{theorem}

\begin{proof}
Let $G^{\pi}_{M_i,M_j}(s,a)$ be the discrepancy between $M_i$ and $M_j$ on a single state-action pair $(s,a)$, i.e. $G^\pi_{M_i,M_j}(s,a)= \mathbb{E}_{{\tilde{s}'}\sim P_{M_j}(\cdot\vert s,a)}[V^\pi_{M_j}({\tilde{s}'})] - \mathbb{E}_{s'\sim P_{M_i}(\cdot\vert s,a)}[V^{\pi}_{M_j}(s')]$. 
We construct $Z_k$ be the discounted return when using $\pi$ to sample in model ${M_i}$ for $k$ steps and then in $M_j$ for the rest with a starting point $s_0 = s$, that is, 
\begin{equation*}
    Z_k = \mathop{\E}\limits_{\substack{\forall t, a_t \sim \pi(\cdot\vert s_t) \\ \forall t < k , s_{t+1}\sim P_{M_i}(\cdot \vert s_t, a_t) \\ \forall t\geq k, s_{t+1}\sim P_{M_j}(\cdot\vert s_t, a_t)} } 
\Bigg[ \sum_{t=0}^{\infty} \gamma^t R(s_t, a_t) s_0 = s  \Bigg]
\end{equation*}
Base on this definition, we have that $V^{\pi}_{M_i}(s) = Z_{\infty}$ and $V_{M_j}^{\pi}(s) = Z_0$. Then, we can  decompose $V^{\pi}_{M_i}(s) - V^{\pi}_{M_j}(s)$ into a sum of $Z_k$: 
\[
V_{M_i}^{\pi}(s) - V_{M_j}^{\pi}(s) = \sum_{k = 0}^{\infty} (Z_{k+1} - Z_k)
\]
We can find that, $Z_{k +1}$ and $Z_k$ only differ in their dynamical model used in the $k$-th step rollout. And we can rewrite them to be : 
\begin{align*}
    & Z_k = r_{k} + \E_{s_k, a_k \sim \pi,P_{M_i} } \Big[\E_{\tilde{s}_{k+1}\sim P_{M_j}(\cdot\vert s_k, a_k)}\Big[ \gamma^{k+1} V^{\pi}_{M_j}(\tilde{s}_{k+1} )\Big] \Big]\\
    & Z_{k+1} = r_{k} + \E_{s_k, a_k \sim \pi, P_{M_i}}\Big[\E_{s_{k+1}\sim P_{M_i}(\cdot\vert s_k, a_k)}\Big[ \gamma^{k+1} V_{M_j}^{\pi}(s_{k+1})\Big]\Big]
\end{align*}
Here, $r_k$ denotes the reward from the first $j$ steps from the real environment. Combine the two equations above together and we get:
\[
Z_{k+1} - Z_{k} = \gamma^{k+1} \E_{s_k, a_k \sim \pi, P_{M_i}} \Bigg[ 
\mathop{\E}\limits_{\substack{s_{k+1}\sim P_{M_i}(\cdot \vert s_k, a_k) \\ \tilde{s}_{k+1}\sim P_{M_j}(\cdot \vert s_k, a_k)}}  \Big[ V^{\pi}_{M_j}(s_{k+1})  - V^{\pi}_{M_j} (\tilde{s}_{k+1}) \Big]
\Bigg]
\]
Then, we can obtain the following conclusion by adding up all $Z_{k+1} - Z_{k}$:
\begin{equation*}
    V_{M_i}^{\pi} - V_{M_j}^{\pi} = \frac{\gamma}{1-\gamma} 
    \mathop{\E}\limits_{s\sim d_{M_i}^{\pi}\atop a\sim \pi(\cdot \vert s)}
    \Bigg[
\E_{s' \sim P_{M_i}(\cdot\vert s,a)} \big[V^{\pi}_{{M_j}} (s')\big]
- \E_{\tilde{s}' \sim P_{M_j}(\cdot \vert s,a)} \big[V^{\pi}_{{M_j}}(\tilde{s}') \big]
\Bigg]
\end{equation*}
Here, $d_{M_i}^\pi$ denotes the distribution of state-action pair induced by policy $\pi$ under the dynamical model $M_i$. 

For a starting state $s_0 = s$, and two dynamical models $M_2$, $M_1$, with the definition of $G_{M_1, M_2}^\pi$ we have: 
\[
V^{ \pi}_{M_2}(s) - V^\pi_{M_1}(s) = \frac{\gamma}{1-\gamma} \mathbb{E}_{s,a\sim d_{M_1}^\pi} \Big[G^\pi_{M_1,M_2}(s,a) \Big]
\]

From the definition: $G^\pi_{M_1,M_2}(s,a)= \mathbb{E}_{{\tilde{s}'}\sim P_{M_2}(\cdot\vert s,a)}[V^\pi_{M_2}({\tilde{s}'})] - \mathbb{E}_{s'\sim P_{M_1}(\cdot\vert s,a)}[V^{\pi}_{M_2}(s')]$.

In the case of deterministic dynamics: for clarity, we write $s' = M_i(s,a)$ instead of $s'\sim P_{M_i}(s'\vert s,a)$, then we rewrite $G_{M_1,M_2}(s,a)$ as: $G^\pi_{M_1,M_2}(s,a) = V^\pi_{M_2}(M_2(s,a)) - V^{\pi}_{M_2}(M_1(s,a))$, with the Lispchitzness, we have that $\vert G^\pi_{{M_1}, M_2}(s,a)\vert  \leq L\cdot \vert M_2(s,a) - M_1(s,a)\vert $.

In the case of stochastic dynamics:  when $L \geq \frac{R}{1-\gamma}$, we have 
\[
\begin{aligned}
\vert G^\pi_{M_1,M_2}(s,a)\vert &= \vert \mathbb{E}_{{\tilde{s}'}\sim P_{M_2}(\cdot\vert s,a)}[V^\pi_{M_2}({\tilde{s}'})] - \mathbb{E}_{s'\sim P_{M_1}(\cdot\vert s,a)}[V^{\pi}_{M_2}(s')]\vert \\
&= \vert \sum_{\tilde{s}'\in {\cal S}}(P_{M_2}(s'\vert s,a)-P_{M_1}(s'\vert s,a))V_{M_2}^\pi(s')\vert \\
&\leq \vert \max_{s'}V_{M_2}^\pi(s') \vert \cdot \vert P_{M_2}(\cdot\vert s,a ) - P_{M_1}(\cdot\vert s,a)\vert \\
&\leq L\cdot \vert P_{M_2}(\cdot\vert s,a ) - P_{M_1}(\cdot\vert s,a)\vert.
\end{aligned}
\]
Note that we require $L$ to be larger than $\frac{R}{1-\gamma}$ in the infinite horizon settings. Previous work~\cite{fan2021model} has found that $L$ has a dependence on $R\cdot H$ when equipped with a maximum horizon of $H$.

Thus, with the Lipschitzness of $V_{M_1}^\pi$ and $V_{M_2}^\pi$, we have that $\vert G^\pi_{{M_1}, M_2}(s,a)\vert  \leq L\cdot \vert P_{M_2}(\cdot\vert s,a) - P_{M_1}(\cdot\vert s,a)\vert $.

\[
\begin{aligned}
|V_{M_2}^\pi(s) - V_{M_1}^\pi(s)\vert &= \frac{\gamma}{1-\gamma}\cdot \vert\mathbb{E}_{s,a\sim d_{M_2}^\pi} [G^\pi_{M_1,M_2}(s,a)] \vert \\
&\leq \frac{\gamma}{1-\gamma}\cdot \mathbb{E}_{s,a\sim d_{M_2}^\pi} [\vert G^\pi_{M_1,M_2}(s,a)\vert] \\
&\leq  \frac{\gamma}{1-\gamma}L\cdot \mathbb{E}_{s,a\sim d_{M_2}^\pi}[\vert P_{M_2}(\cdot\vert s,a) - P_{M_1}(\cdot\vert s,a)\vert]
\end{aligned}
\]
 Observe that $\vert \sup_x f(x)- \sup_x g(x)\vert \leq \sup_x\vert f(x) - g(x)\vert$, where $f$ And $g$ are real valued functions. This implies:~~
\[
\vert V_{M_2}^* (s) - V_{M_1}^*(s)\vert  = \vert \sup_{\pi\in \Pi} V_{M_2}^{\pi}(s) - \sup_{\pi\in \Pi} V_{M_1}^{\pi}(s)\vert \leq \sup_{\pi\in \Pi} \vert  V_{M_2}^\pi(s) - V_{M_1}^\pi(s)\vert
\]
Thus, we have that for starting state $s_0 \sim \mu$:
\[
V_{M_2}^*(\mu) - V_{M_1}^*(\mu) \geq - \frac{\gamma}{1-\gamma}L\cdot \sup_{\pi\in \Pi} \mathbb{E}_{s,a\sim d_{M_2}^\pi}\Big[\vert P_{M_2}(\cdot\vert s,a)- P_{M_1}(\cdot\vert s,a)\vert\Big]
\]
\end{proof}

\begin{theorem}[Refined bound with constraints]\label{proof-constraint-bound}
Here, $\gS$ denotes the state space simplex.  Here, policy $\pi_i\in \Pi$ is the $\epsilon_{opt}$ optimal policy under the dynamical model $M_i\in \gM$. 
For a dynamical model $M_i \in \gM$,  $V_{M_i}^*(\mu)$  denotes the maximal returns on dynamics $P_{M_i}$.  We give the model shift constraints under the TV-distance. 
Then the performance difference bound can be refined under the model shift constraints as:
\begin{equation*}
     \begin{split}
        V^{\pi_2\vert M_2} - V^{\pi_1\vert M_1}\geq &  \kappa\cdot (\mathbb{E}_{s,a\sim d^{\pi_1}} \TV [P(\cdot \vert s,a) \Vert  P_{M_1}(\cdot\vert s,a)] 
        \\
        -& \mathbb{E}_{s,a\sim d^{\pi_2}} \TV [P(\cdot \vert s,a)\Vert P_{M_2}(\cdot\vert s,a)]) -  \frac{\gamma}{1-\gamma}L\cdot  2\sigma_{M_1,M_2} - \epsilon_{opt}\\
         \textit{s.t.} \quad& \TV(P_{M_2}(\cdot\vert s,a) \Vert P_{M_1}(\cdot\vert s,a)) \leq  \sigma_{M_1,M_2}, \quad \forall (s,a) \in {\cal S\times A}
     \end{split}
\end{equation*}
\end{theorem}

\begin{proof}
Let $\mu$ and $v$ be two probability distributions on the configuration space $\cal X$, according to Lemma~\ref*{tv-distance}, then we have ${\TV}(\mu\Vert v) = \frac{1}{2}\sum\limits_{x\in {\cal X}}\vert \mu(x)- v(x)\vert$. 

Recall $\epsilon_{M_i}^\pi$ denote the discrepancy between the learned dynamics $P_{M_i}$ and the true dynamics, i.e. $\epsilon_{M_i}^\pi = \mathbb{E}_{s,a\sim d^\pi} [{\TV}(P(\cdot\vert s,a)\Vert  P_{M_i}(\cdot\vert s,a))]$. 

Under these definitions, we can yield the following intermediate outcome by applying the results from ~\ref*{ap:ceiling-bound} and ~\ref*{proof-general-gap}
\begin{equation*}.
\begin{split}
       V^{\pi_2\vert M_2} - V^{\pi_1\vert M_1} \geq& -\kappa\cdot(\epsilon_{M_2}^{\pi_2}-\epsilon_{M_1}^{\pi_1}) + V_{M_2}^* - V_{M_1}^* - \epsilon_{opt}\\
\geq & -\kappa\cdot \Big\{ \mathbb{E}_{s,a\sim d^{\pi_2}} {\TV}[P(\cdot \vert s,a) \Vert  P_{M_2}(\cdot\vert s,a)]  \\&- \mathbb{E}_{s,a\sim d^{\pi_1}} {\TV}[P(\cdot \vert s,a)\Vert P_{M_1}(\cdot\vert s,a)]\Big\}
\\ & - \frac{\gamma}{1-\gamma}L\cdot \sup_{\pi\in \Pi} \mathbb{E}_{s,a\sim d^\pi_{M_2}} \Big[\vert  P_{M_2}(\cdot\vert s,a)- P_{M_1}(\cdot\vert s,a)\vert\Big] - \epsilon_{opt}\\
\end{split}
\end{equation*}
Recall the following constraint on model shift that we subject to: 
\[
\TV(P_{M_2}(\cdot\vert s,a) \Vert P_{M_1}(\cdot\vert s,a)) \leq  \sigma_{M_1,M_2}, \quad \forall (s,a) \in {\cal S\times A}
\]
Then, the ceiling performance difference can be further bounded as:
\begin{equation*}
\begin{split}
    V_{M_2}^* - V_{M_1}^* &\geq - \frac{\gamma}{1-\gamma}L\cdot \sup_{\pi\in \Pi} \E_{s,a\sim d^\pi_{M_2}} \Big[\vert  P_{M_2}(\cdot\vert s,a)- P_{M_1}(\cdot\vert s,a)\vert\Big]\\
    & \geq -\frac{\gamma}{1-\gamma}L\cdot \sup_{\pi\in\Pi}\E_{s,a\sim d_{M_2}^\pi}\Big[2\sum_{s'\in{\cal S}_2} \frac{1}{2}\vert  P_{M_2}(s'\vert s,a)- P_{M_1}(s'\vert s,a)\vert \Big]\\
    & \geq -\frac{\gamma}{1-\gamma}L\cdot \sup_{\pi\in \Pi}\E_{s,a\sim d_{M_2}^\pi}\Big[ 2\TV(P_{M_2}(\cdot\vert s,a) \Vert P_{M_1}(\cdot\vert s,a)) \Big] \\
    &\geq -\frac{\gamma}{1-\gamma}L\cdot (2\sigma_{M_1,M_2})
\end{split}
\end{equation*}
And finally, we have the following refined bound:
\[
\begin{aligned}
 V^{\pi_2\vert M_2} - V^{\pi_1\vert M_1}\geq & \kappa\cdot (\mathbb{E}_{s,a\sim d^{\pi_1}} {\TV}[P(\cdot \vert s,a) \Vert  P_{M_1}(\cdot\vert s,a)]  - \mathbb{E}_{s,a\sim d^{\pi_2}} {\TV}[P(\cdot \vert s,a)\Vert P_{M_2}(\cdot\vert s,a)])  \\
 &-  \frac{\gamma}{1-\gamma}L\cdot (2\sigma_{M_1,M_2}) - \epsilon_{opt}\\
\end{aligned}
\]
This refined bound is subject to the model shift constraint we set. 

\end{proof}

\begin{corollary}\label{sample-based-corollary}
Under the generative model setting,  let model $M_1$  has already trained on $N$ samples of each $(s, a)$ pair and model $M_2$ on $N+k$ samples per $(s,a)$ pair. Policy $\pi_1$ is the $\epsilon_{opt}$-optimal policy on $M_1$ and so is $\pi_2$  on $M_2$. Recall  $ \epsilon =\delta_{M_1}(\cdot\vert s,a) - \frac{(1-\gamma)L}{R}\cdot (2\sigma_{M_1,M_2})- \frac{(1-\gamma)^2}{R\gamma}\cdot \epsilon_{opt} $ and $\xi \in(0,1) $ is a constant.  As the model is trained on true interaction samples, we can work out the amount of samples we need to satisfy the monotonic improvement requirements:
\[
k = \frac{2}{\epsilon^2} \log\frac{2^{vol(\gS)}-2}{\xi}-N
\]
\end{corollary}

\begin{proof}
For simplicity, we denote $\delta_{M_i}(s'\vert s,a) = \vert P(s'\vert s,a) - P_{M_i}(s'\vert s,a)\vert $.

Recall the lower bound for performance difference:
\begin{align*}
C_{M_1,\pi_1}(M_2,\pi_2) &=\frac{2R\gamma}{(1-\gamma)^2} \Big\{\mathop{\mathbb{E}}_{s,a\sim d^{\pi_1}} \big[\frac{1}{2}\sum_{s'\in \gS_2}(\delta_{M_1}(s'\vert s,a)\big] \\
    -&\mathop{\mathbb{E}}_{s,a\sim d^{\pi_2}} \big[\frac{1}{2}\sum_{s'\in \gS_2}(\delta_{M_2}(s'\vert s,a)\big] \Big\}- \frac{\gamma}{1-\gamma}L \cdot (2\sigma_{M_1,M_2}) - \epsilon_{opt}
\end{align*}

Towards this lower bound, we give the assumption that similar models derives similar sub-obptimal policies. 
To be specific, when $M_1$ and $M_2$ are close to each other in terms of $L_1$-norm distance at any transition pair $(s,a)$ : $\vert P_{M_2}(\cdot\vert s,a) - P_{M_1}(\cdot\vert s,a)\vert_1 \leq  \delta$, 
then the sub-optimal policies derived from them are close as well, \emph{i.e.}, there exists $\alpha$, subject to $\vert \pi_2(\cdot\vert s) - \pi_1(\cdot\vert s)\vert_1 \leq  \alpha \cdot \delta$ for all transition pairs.
Here, we show the feasibility of given the $\alpha$ in the control theory perspective. Let a trajectory $s_{1:N}, a_{1:N-1}$ be generated from the dynamical model $M_1$ that satisfies:
\[
s_{t+1} = f_{M_1} (x_t, a_t, 0)
\]
We call it a nominal trajectory. Here, $f_{M_1}$ is a nonlinear function that represents the dynamics of model $M_1$. Then, we can formalize the inconsistency between $M_2$ and $M_1$ by disturbance, $w_i\in \gW$. By entering $w_i$ into $f_{M_1}$ in a general nonlinear way we can get the dynamic of model $M_2$ as:
\[
s_{t+1} = f_{M_1}(s_t, a_t, w_t)
\]
Further, the deviations from the nominal trajectory can be calculated as when giving a disturbance sequence, $w_{1:N-1}$. Let $\delta y$ denote the deviations of variable $y$.
\[
\delta s_t = f_{M_1} (s_t + \delta x_t, u_t, + \delta u_t, w_t) - s_{t+1}
\]
Assume that the deviations are computed with a linear feedback controller, that is,
\[
\delta a_t = - K_t \delta s_t
\]
Actually, we can utilize any reasonable linear controller. Here, we take the time-varying linear quadratic regulator as an instance for illustrating the rationality of our assumption on $\alpha$.
Based on the dynamic Riccati equation, we have the solution as:
\begin{align*}
    & K_t = (R + B_t^T P_{t+1} B_t ) ^{-1} (B_t^T P_{t+1} A_t)\\
    & P_{t-1} = Q_t + A_t^T P_t A_t - A_t^T P_t B_t (R + B_t^T P_t B_t)^{-1}(B_t^T P_t A_t)
\end{align*}
where $A_t = \partial{f_{M_1}}/\partial{s}\vert _{s_t, a_t, 0}$ and $B_t = \partial f_{M_1} / \partial a \vert _{s_t, a_t, 0}$. And, $Q_i \succeq 0$ and $R\succeq 0$ are state and input cost matrices. That is, we have a feasible solution for $\alpha$. 
We then seek to explain the state-action distribution is similar thus we can use $d^{\pi_1}$ as an approximation of $d^{\pi_2}$. 
First of all, we have the distance between two policies, 
\begin{align*}
    \vert \pi_{2} (\cdot \vert s) - \pi_{1} (\cdot \vert s) \vert_{1}\leq \alpha \cdot \delta \triangleq \beta
\end{align*}

Denote $\mathbb{P}_\pi^h$ as the state distribution resulting from $\pi$ at time step $h$ with $\mu$ as the initial state distribution. We consider bounding $\vert \mathbb{P}_h^{\pi_2} - \mathbb{P}_h^{\pi_1}\vert_1$ with $h>1$.
\begin{align*}
    &\mathbb{P}_{h}^{\pi_{2}} (s')
    - \mathbb{P}_{h}^{\pi{1}} (s')\\
    =&
    \sum \limits_{s,a}
    \left(
        \mathbb{P}_{h-1}^{\pi_{2}} (s) \pi_{2} (a \vert s)
        - \mathbb{P}_{h-1}^{\pi_{1}} (s) \pi_{1} (a \vert s)
    \right)
    P(s' \vert s,a) \\
    =&
    \sum \limits_{s,a}
    \left(
        \mathbb{P}_{h-1}^{\pi_{2}} (s) \pi_{2} (a \vert s)
        - \mathbb{P}_{h-1}^{\pi_{2}} (s) \pi_{1} (a \vert s)
        + \mathbb{P}_{h-1}^{\pi_{2}} (s) \pi_{1} (a \vert s)
        - \mathbb{P}_{h-1}^{\pi_{1}} (s) \pi_{1} (a \vert s)
    \right)
    P(s' \vert s,a) \\
    =&
    \sum \limits_{s} \mathbb{P}_{h-1}^{\pi_{2}} (s)
    \sum \limits_{a}
    \left(
        \pi_{2} (a \vert s) - \pi_{1} (a \vert s)
    \right)
    P(s' \vert s,a) +
    \sum \limits_{s}
    \left(
        \mathbb{P}_{h-1}^{\pi_{2}} (s)
        - \mathbb{P}_{h-1}^{\pi_{1}} (s)
    \right)
    \sum \limits_{a} \pi_{1} (a \vert s) P(s' \vert s,a)
    .
\end{align*}
Apply absolute value on both sides, we then get:
\begin{align*}
    \sum \limits_{s'}
    \vert  
        \mathbb{P}_{h}^{\pi_{2}} (s')
        - \mathbb{P}_{h}^{\pi_{1}} (s')
    \vert
    \le &
    \sum \limits_{s} \mathbb{P}_{h-1}^{\pi_{2}} (s)
    \sum \limits_{a}
    \vert
        \pi_{2} (a \vert s)
        - \pi_{1} (a \vert s)
    \vert
    \sum \limits_{s'} P(s' \vert s,a) \\
    & + \sum \limits_{s}
    \vert
        \mathbb{P}_{h-1}^{\pi_{2}} (s)
        - \mathbb{P}_{h-1}^{\pi_{1}} (s)
    \vert
    \sum \limits_{s'}
    \sum \limits_{a} \pi_{1} (a \vert s) P(s' \vert s,a) \\
    \le &
     \beta +
    \lVert
        \mathbb{P}_{h-1}^{\pi_{2}}
        - \mathbb{P}_{h-1}^{\pi_{1}}
    \rVert_{1}
    \le
     2\beta +
    \vert
        \mathbb{P}_{h-2}^{\pi_{2}}
        - \mathbb{P}_{h-2}^{\pi_{1}}
    \vert_{1}
    =
     h \beta
    .
\end{align*}
Under the definition of $d_{\mu}^{\pi}$, we have:
\begin{align*}
    \vert d_{\mu}^{\pi_{2}} - d_{\mu}^{\pi_{1}}\vert_1
    =
    \vert(1 - \gamma)
    \sum \limits_{h=0}^{\infty}
    \gamma^{h}
    \left(
        \mathbb{P}_{h}^{\pi_{2}}
        - \mathbb{P}_{h}^{\pi_{1}}
    \right)\vert
    \leq \beta\gamma/(1-\gamma)
    .
\end{align*}
Upon these analyses, we find that similar model derives similar policy, which invokes similar state-action distribution. Thus we can approximate state-action visitation density $d^{\pi_2}$ by previous visitation density $d^{\pi_1} $. 
Also, we assume the same state space $\gS$. Then, we get a approximation of $C_{M_1,\pi}(M_2,\hat \pi)$ as following: 
\begin{equation*}
\begin{split}
    & \tilde{C}_{M_1,\pi_1}(M_2,\pi_2) \\
    =&\frac{2R\gamma}{(1-\gamma)^2}\mathop{\mathbb{E}}_{s,a} \Big[\frac{1}{2}\sum_{s'\in \gS}(\delta_{M_1}(s'\vert s,a)- \delta_{M_2}(s'\vert s,a))\Big] - \frac{\gamma}{1-\gamma}L\cdot (2\sigma_{M_1,M_2}) - \epsilon_{opt}\\
    = & \frac{\gamma}{(1-\gamma)} \Big\{ \frac{R}{1-\gamma} \mathop{\mathbb{E}}_{s,a}
    \big[\sum_{s'\in \gS}(\delta_{M_1}(s'\vert s,a)- \delta_{M_2}(s'\vert s,a))\big]
   - L\cdot \mathop{\mathbb{E}}_{s,a}\big[\sum_{s'\in \gS}(\frac{1}{vol(\gS)}2\sigma_{M_1,M_2})\big] \\
   &\quad -  \frac{(1-\gamma)}{\gamma} \mathop{\E}\limits_{s,a}\big[ \frac{1}{vol(\gS)}\sum_{s'\in \gS}(\epsilon_{opt}) \big]\Big\}
\end{split}
\end{equation*}
Thus, when meeting the following requirements for each $(s,a)$ pair, we can guarantee the monotonic improvement for that $V^{\pi_2\vert M_2} - V^{\pi_1\vert M_1} \geq \tilde{C}_{M_1,\pi_1}(M_2,\pi_2) \geq 0$. 
\begin{equation*}
\delta_{M_2}(\cdot \vert s,a) \leq \delta_{M_1}(\cdot\vert s,a) - \frac{(1-\gamma)L}{R}\cdot (2\sigma_{M_1,M_2}) - \frac{(1-\gamma)^2}{R\gamma}\cdot \epsilon_{opt}
\end{equation*}
By applying Lemma~\ref*{center}, the $L_1$ deviation of the empirical distribution ${P}_{M_2}(\cdot\vert s,a)$and true $P(\cdot\vert s,a)$ over $vol(\gS)$ distinct events from $n$ samples is bounded by:
\[
\Pr (\vert  P(\cdot\vert s,a) - {P}_{M_2}(\cdot\vert s,a)\vert_1< \epsilon) \geq 1 - (2^{vol(\gS)}-2)\exp(-\frac{(N+k)\epsilon^2}{2})
\]
Then for a fixed $(s,a)$, with probability greater than $1-\xi$, we have:
\[
\vert P(\cdot\vert s,a) - P_{M_2}(\cdot\vert s,a)\vert_1  \leq \sqrt{\frac{2}{N+k}\cdot \log \frac{2^{vol(\gS)}-2}{\xi}}
\]
Let $ \epsilon = \delta_{M_1}(\cdot\vert s,a) - \frac{(1-\gamma)L}{R}\cdot (2\sigma_{M_1,M_2}) - \frac{(1-\gamma)^2}{R\gamma}\cdot \epsilon_{opt} $. 
Our requirements can be further shown in this form:
\[
\sqrt{\frac{2}{N+k}\cdot \log \frac{2^{vol(\gS)}-2}{\xi}}  = \delta_{M_1}(\cdot\vert s,a) - \frac{(1-\gamma)L}{R}\cdot (2\sigma_{M_1,M_2})- \frac{(1-\gamma)^2}{R\gamma}\cdot \epsilon_{opt}
\]
Finally, with probability greater than $1-\xi$,  we can guarantee the monotonic improvement when having:
\[
k = \frac{2}{\epsilon^2} \log\frac{2^{vol(\gS)}-2}{\xi}-N
\]
\end{proof}

\section{Toolbox}\label{toolbox}

\begin{lemma}[Total variation distance]\label{tv-distance}
Let $\mu$ and $v$ be two probability distributions on the configuration space $\gX$. Then
\[
\TV (\mu \Vert v) = \frac{1}{2} \sum_{x\in \gX} \vert \mu(x) - v(x)\vert 
\]
\end{lemma}
\begin{proof}
Let $B = \{x \in \gX: \mu(x) \geq v(x)\}$, and $A \subseteq \gX$ be any event. 
Since $\mu(x)-v(x) <0 $ for any $x\in A \cap B^c$, we have
\[
\mu(A) - v(A) \leq \mu(A\cap B) - v(A\cap B) \leq \mu(B) - v(B)
\]
For all events $A$, $\vert \mu(A) - v(A) \vert \mu(B) - v(B)$, and the equality is achieved for $A = B$ or $A = B^c$. Thus, we get that
\[
\TV(\mu \Vert v) = \frac{1}{2} [\mu(B) - v(B) + v(B^c) - \mu(B^c)] = \frac{1}{2}\sum_{x\in \gX}\vert \mu(x) - v(x) \vert
\]
\end{proof}

\begin{lemma}[Relationship between true returns and model returns]\label{lemma:dis-bound}
Let $\epsilon_{M}^\pi$ denote the inconsistency between the learned dynamics $P_M$ and the true dynamics, $\mathbb{E}_{s,a\sim d^\pi} [\TV(P(\cdot\vert s,a)\Vert  P_{M}(\cdot\vert s,a))]$, where $d^\pi \doteq d^\pi(s,a;\mu)$  is the probability of visiting state-action pair $(s, a)$ after starting at state $s_0 \sim \mu$ and following $\pi$ thereafter under the true dynamics.  Then the true returns can be represented as below:
\[
V^\pi(\mu) \geq V_{M}^\pi(\mu) - \frac{2R\gamma}{(1-\gamma)^2}\epsilon_M^\pi
\]
\end{lemma}

\begin{proof}
Given policy $\pi$ and dynamics $P(\cdot\vert s,a)$, we denote the density of state-action visitation after $h$ steps from starting state $s_0\sim \mu$ as $\rho_h^\pi( \mu; P) = \mathbb{E}_{s_0 \sim \mu}[\rho_h^\pi(s_h,a_h \vert s_0; P)]$.

Then the discounted returns are bounded as:
\begin{equation*}
    \begin{split}
         V^\pi(\mu) - V_M^\pi(\mu) =&
         \sum\limits_{h=0}^{\infty} \mathbb{E}_{s,a\sim \rho_h^\pi(\mu;P)}[\gamma^h r(s,a)]  - \sum\limits_{h=0}^{\infty} \mathbb{E}_{s,a\sim \rho_h^\pi(\mu;P_M)}[\gamma^h r(s,a)] \\
        \geq & -\sum_{h=0}^{\infty} \gamma^h \vert \mathbb{E}_{s,a\sim \rho_h^\pi(\mu;P)}[r(s,a)] - \mathbb{E}_{s,a\sim \rho_h^\pi(\mu;P_M)}[\gamma^h r(s,a)] \vert\\
        \geq & -\sum_{h=0}^{\infty} \gamma^h  \sum_{s,a\in {\cal S\times A}} R\vert \rho_h^\pi(\mu; P) - \rho_h^\pi(\mu; P_M)\vert\\
= & -2R\cdot \sum_{h=0}^{\infty}\gamma^h \frac{1}{2}\sum_{s,a\in {\cal S,A}} \vert \rho_h^\pi(\mu; P) - \rho_h^\pi(\mu; P_M)\vert\\
= & -2R \cdot \sum_{h=0}^{\infty} \gamma^h \TV (\rho_h^\pi(\mu; P)\Vert \rho_h^\pi(\mu; P_M)) \\
    \end{split}
\end{equation*}

Using the property of Markov chain TV distance bound, then, 
\begin{equation*}
\begin{split}
         V^\pi(\mu) - V_M^\pi(\mu) \geq& -2R \cdot \sum_{h=0}^{\infty}\gamma^h\TV ( \rho_h^\pi(\mu; P) \Vert  \rho_h^\pi(\mu; P_M)) \\
\geq & 
\sum_{h=1}^{\infty}-2R\cdot \gamma^h \Big\{\TV (\rho_{h-1}^\pi(\mu; P) \Vert  \rho_{h-1}^\pi(\mu; P_M)) \\
& + \E_{s,a\sim d^\pi}[\TV[(P(\cdot\vert s,a)\Vert  P_{M}(\cdot\vert s,a))]] + \TV(\pi\Vert \pi)\Big\}
\end{split}
\end{equation*}
By plugging the results back, we then get, 
\[
V^\pi(\mu) - V_M^\pi(\mu)  \geq -2R\sum_{h=0}^{\infty} \gamma^h h\cdot \epsilon_M^\pi = -\frac{2R\gamma}{(1-\gamma)^2}\cdot \epsilon_M^\pi
\]
\end{proof}

\begin{lemma}[Inequalities for the $L_1$ deviation of the empirical distribution]\label{center}
Let $P$ be a probability distribution on the set $\gA=\{1, \ldots, a\}$. For a sequence of samples $x_1, \ldots, x_m \sim P$, let $\hat{P}$ be the empirical probability distribution on $\gA$ defined by $\hat{P}(j)= \frac{1}{m}\sum_{i =1}^{m} \mathbbm{1}(x_i = j)$.
The $L_1$-deviation of the true distribution $P$ and the empirical distribution $\hat{P}$ over $\gA$ from $m$ independent identically samples is bounded by,
\begin{equation*}
    Pr(\vert P-\hat{P})\vert_1 \geq \epsilon) \leq (2^{\vert \gA\vert}-2) e^{-m\epsilon^2/2}.
\end{equation*}
\begin{proof}
For a probability distribution $P$ on $\gA$, we define
\begin{equation*}
    \pi_p = \max\limits_{A\subseteq \gA}\min(P(A),1-P(A)).
\end{equation*}
And for $p \in [0, 1/2)$, we define
\begin{equation*}
    \varphi(p) = \frac{1}{1-2p}\log \frac{1-p}{p}.
\end{equation*}
and, by continuity, set $\varphi(1/2)=2$.

According to~\citet{weissman2003inequalities}, the  $L_1$-deviation of the true distribution $P$ and the empirical distribution $\hat{P}$ is bound by, 
\begin{equation*}
    Pr(\vert P-\hat{P})\vert_1 \geq \epsilon) \leq (2^{\vert \gA\vert}-2) e^{-m\varphi(\pi_P)\epsilon^2/4}.
\end{equation*}

Firstly, for any $P$, we have
\begin{equation*}
    \pi_p = \max\limits_{A\subseteq \gA}\min(P(A), 1-P(A))
    \leq \max\limits_{A\subseteq \gA}(\frac{P(A) + 1-P(A)}{2}) = 1/2.
\end{equation*}
and note that $\pi_P =  1/2$ when $P(A)=1/2$.

Then, we claim that the function $\varphi(p)$ is strictly decreasing for $p\in [0, 1/2]$.
Differentiating $\varphi(p)$ with respect to $p$ yields
\begin{equation*}
    \varphi'(p) = \frac{1}{(1-2p)^2}\Big[ -\frac{1-2p}{1-p} - \frac{1-2p}{p} + 2\log  \frac{1-p}{p}\Big].
\end{equation*}

For $p\in (0, 1/2)$, there always exists  $\frac{1}{(1-2p)^2}> 0$. Thus, to show that $\varphi'(p) < 0$ for $p \in (0, 1/2)$, it suffices to show that
\begin{equation*}
    g(p) = -\frac{1-2p}{1-p} - \frac{1-2p}{p} + 2\log  \frac{1-p}{p} < 0.
\end{equation*}
The derivative of $g(p)$ is
\begin{equation*}
    g'(p) = (\frac{1}{1-p} - \frac{1}{p})^2 > 0, \quad p\in (0, 1/2).
\end{equation*}
Note that $g(1/2) = 0$, thus we have $\varphi'(p) < 0$ for $p \in (0, 1/2)$. And continuity arguments complete the claim for $p=1/2$ and $p=0$.

It is then no difficult to see that for any probability distribution $P$, 
\begin{equation*}
    \varphi(\pi_P) \geq \varphi(1/2) = 2.
\end{equation*}
Therefore
\begin{equation*}
     Pr(\vert P-\hat{P})\vert_1 \geq \epsilon) \leq (2^{\vert \gA\vert}-2) e^{-m\epsilon^2/2}.
\end{equation*}

\end{proof}

\end{lemma}

\section{Comparison with Prior Works}

To begin with, an important fact is that the effect of model shifts on trajectories is drastic. For example, even when the system dynamics satisfy $L$- Lipschitz continuity, along with the policy and the initial state are the same,  the difference in trajectories sampled in $M_1, M_2$ grows at $e^{LH}$ with the length $H$ of the trajectory~\cite{li2014event}. As the model shifts decay, the trajectory discrepancy will also decrease sharply. It implies that model shift stays a substantial influence during the MBRL training process. 

There are two main trends of local view analysis:

\paragraph{API~\cite{kakade2002approximately} class.} Their recipe for monotonicity analysis is $V^{\pi_{n+1}} (\mu)-  V^{\pi_n}(\mu) \geq C(\pi_n, \pi_{n+1}, \epsilon_m)$. If policies update $\pi_n\rightarrow \pi_{n+1}$  could provide a non-negative $C(\pi_n, \pi_{n+1}, \epsilon_m)$ , then the performance is guaranteed to increase.  Here, $\epsilon_m =\max\limits_{\pi\in \Pi, M\in {\cal M}}\mathbb{E}_{s,a\sim d^{\pi}}[{\cal D}_{TV}(P(\cdot\vert s,a)\Vert P_M(\cdot\vert s,a))] $ . Most previous works~\cite{schulman2015trust,kakade2002approximately} were derived under model-free settings ($\epsilon_m=0$). they use conservative policy iteration, for example, by forcing ${\cal D}_{TV}(\pi_n\Vert \pi_{n+1})\leq \alpha$), then the state-action distribution are close as well ${\cal D}_{TV}(d^{\pi_n}\Vert d^{\pi_{n+1}})\leq \frac{\alpha\gamma}{1-\gamma} $, so that they can optimize over their performance difference lemma, e.g., $C(\pi_n, \pi_{n+1}, 0) \approx \frac{1}{1-\gamma}\mathbb{E}_{s,a\sim d^{\pi_n}}[A^{\pi_n}(s,a)] $.
When $\epsilon_m>0$, this approximation $C(\pi_n, \pi_{n+1}, 0) \approx \frac{1}{1-\gamma}\mathbb{E}_{s,a\sim d^{\pi_n}}[A^{\pi_n}(s,a)] $ fails. It is non-trivial to apply the results to model-based settings.

DPI~\cite{sun2018dual} tries to force $\pi_{n+1}$ and $\pi_n$ to be close, which will result in a high similarity of the data they sample. Thus, a risk arises from it, this approach would limit the growth of the policy exploration in the real environment, thereby leading the inferred models to stay optimized in a restrictive local area. For example, in the Humanoid environment, the agent struggles to achieve balance at the beginning of training. By then, an updated restricted policy will cause the exploration space to be limited in such an unbalanced distribution for a long time, and the learned model in such highly repetitive data will converge quickly with a validation loss be zero. However, the success trajectory has not been explored yet, implying that both the policy and the learned model will fall into a poor local optimum.

  Besides, the definition of model accuracy (Eq.3) is a local view in DPI, i.e., $\hat{P}$ is $\delta$-opt under $d^{\pi_n}$. If we replace model accuracy with a more general, global definition (for example, $\hat{P}$ is $\delta$-opt under $d^{\pi_{n+1}}$ , or $\hat{P}$ is $\delta$-opt under all $(s,a,s')$ tuples), we find that the $\delta$ in (Eq. 3) will be large at the initial steps, making it difficult to obtain a local optimal solution in Theorem 3.1.

  Finally, theoretical analysis in DPI can only guide the policy iteration process, while the update of the model is passive, which is different from our global view theory.

\paragraph{Discrepancy bound class~\cite{luo2018algorithmic, janner2019trust}.}  They mostly derive upon $V^{\pi_n}(\mu)\geq V_M^{\pi_n}(\mu) - C(\epsilon_m, \epsilon_\pi)$.  As guaranteed in them,  once a policy update $\pi_n \rightarrow \pi_{n+1}$ has improved returns under the same model $M$, i.e., $V_{M}^{\pi_{n+1}}(\mu) > V_{M}^{\pi_n}(\mu) + C(\epsilon_m, \epsilon_\pi)$ , it would improve the lower bound on the performance evaluated in the real environment, i.e., $\inf\{V^{\pi_2\vert M}(\mu)\} > \inf \{ V^{\pi_1\vert M(\mu)}\} $\}.  

Their theory is based on a fixed model $M$, or an upper bound on the distribution shift of all models $\epsilon_m$.  It does not concern the change in model dynamics during updating, nor the performance varying due to the model shift. Moreover, The solution would be very coarse if only the upper bound of the model shift is given. Even worse, the given upper bound is likely to be too large, then it will fail to find a feasible solution for $V_M^{\pi_{n+1}}(\mu) - V_M^{\pi_n}(\mu) \geq C(\epsilon_m, \epsilon_\pi)$ in practice, thus making the monotonicity guarantee fails.

\section{Experimental Details}\label{implementation}
\subsection{Environment Setup}
We evaluate all algorithms on a set of MuJoCo~\cite{todorov2012mujoco} continuous control benchmark tasks. We adopt the standard full-length version of all these tasks.  Among then, we truncate some redundant observations for Hopper, Ant and Humanoid as our model-based baselines (MBPO\cite{janner2019trust}, AutoMBPO\cite{lai2021effective}) do.  The details of the experimental environments are provided in Table~\ref{env-setting}.

\begin{table}[h]
    \caption{Overview on Environment settings. Here, $\theta_t$ denotes the joint angle at time $t$. and $z_t$ denotes the height.}
    \begin{center}
    \resizebox{0.85\textwidth}{!}{
        \begin{tabular}{
            >{\centering}m{0.15\textwidth}
            | c
            | c
            | c
            | c
        }
            \toprule
            & \makecell[c]{State Space \\Dimension} & \makecell[c]{Action Space \\Dimension} & Horizon & Terminal Function \\
            \midrule
            Hopper-v2
            & 11
            & 3
            & 1000
            & $z_t \leq 0.7$ or $\mathbb{\theta}_t \geq 0.2$
            \\
            \midrule
            Swimmer-v2
            & 8
            & 2
            & 1000
            & None
            \\
            \midrule
            Walker2d-v2
            & 17
            & 6
            & 1000
            & \makecell[c]{$z_t \geq 2.0$ or $z_t \leq 0.8$ or \\ $\mathbb{\theta}_t \leq -1.0$  or $\mathbb{\theta}_t\geq 1.0$}\\
            \midrule
            HalfCheetah-v2
            & 17
            & 6
            & 1000
            & None
            \\
            \midrule
            Ant-v2
            & 27
            & 8
            & 1000
            & $z_t < 0.2$ or $z_t > 1.0$
            \\
            \midrule
            Humanoid-v2
            & 45
            & 17
            & 1000
            & $z_t < 1.0$ or $z_t > 2.0$
            \\
            \bottomrule
        \end{tabular}}
    \end{center}
    \label{env-setting}
\end{table}

The environment settings for the ablation study on the 
generalizability of event-triggered mechanism are presented in Table~\ref{env-setting-ablation}
\begin{table}[h]
    \caption{Overview on Environment settings in Ablation.}
    \begin{center}
    \resizebox{0.85\textwidth}{!}{
    
        \begin{tabular}{
            >{\centering}m{0.15\textwidth}
            | c
            | c
            | c
            | c
        }
            \toprule
            & \makecell[c]{State Space \\Dimension} & \makecell[c]{Action Space \\Dimension} & Horizon & Terminal Function \\
            \midrule
            Kitty Stand
            & 61
            & 12
            & 50
            & $u_{t,kitty} \leq 0$
            \\
            \midrule
            Panda Reach
            & 20
            & 7
            & 50
            & None
            \\
            \bottomrule
        \end{tabular}}
    \end{center}
    \label{env-setting-ablation}
\end{table}

\subsection{Baselines and implementation}
\paragraph{MFRL algorithms.} We compare to two state-of-the-art model-free baselines, SAC~\cite{haarnoja2018soft} and PPO~\cite{schulman2017proximal}. The hyperparameters are kept the same as the authors’. Regarding the low sample efficiency of MFRL methods, we ran 5M steps for them, which is an order of magnitude more than in  MBRL, to fairly evaluate the asymptotic performance of these MFRL algorithms. The implementation of SAC is based on the opensource repo (\citet{saclib}, MIT License).
\paragraph{MBRL algorithms.} As for model-based methods, we compare with several algorithms including PETS~\cite{chua2018deep}, SLBO~\cite{luo2018algorithmic}, MBPO~\cite{janner2019trust} and AutoMBPO~\cite{lai2021effective}.  Our algorithm CMLO is implemented based on the opensource toolbox for MBRL algorithms, MBRL-LIB~\cite{Pineda2021MBRL} (MIT License). The implementation of SLBO mainly follows~\citet{wang2019benchmarking}. To ensure a fair comparison, we run CMLO and MBPO with the same network architectures and training configurations based on MBRL-LIB.

We report the asymptotic performance on six benchmark tasks in Table~\ref{maximum-average-returns}. Results show that our method has comparable asymptotic performance in each benchmarks to both MBRL and MFRL baselines. Each result is averaged over seven trials using different random seeds. For MBRL baselines, the performances on different tasks are capped at different timesteps when the learning curves come to converge, we choose 125k for Hopper, 350k for Walker2d and Swimmer, 400k for HalfCheetah, 300k for Ant and 250k for Humanoid.

\begin{table}[htbp]
    \scriptsize
    \caption{Comparative results. The results show the average and standard deviation on the \textbf{maximum average returns} among different trails.}
    
    \begin{center}
    \resizebox{\textwidth}{!}{
        \begin{tabular}{
            c|c|ccc
        }
            \toprule
            \multicolumn{2}{c}{} & Hopper & Walker2d & Swimmer \\
            \midrule
            
            \multirow{2}{*}{\makecell[c]{MFRL\\ (@5M steps)}}
                & SAC
                & 4257.92 $\pm$ 100.23
                & 7898.01 $\pm$ 563.45
                & 195.60 $\pm$ 5.97
            \\
            
            {}
                & PPO
                & 3114.76 $\pm$ 1039.06
                & 5740.75 $\pm$ 500.89
                & 129.66 $\pm$ 9.78
            \\
            
            \midrule
             & PETS
                & 571.25 $\pm$ 71.14
                & 1174.79 $\pm$ 471.39
                & 92.61 $\pm$ 4.29
            \\
            
            {}
            \multirow{5}{*}{MBRL}
                & SLBO
                & 278.82 $\pm$ 65.83
                & 3129.70 $\pm$ 154.16
                & 71.02 $\pm$ 1.98
            \\
            
            {}
                & AutoMBPO
                & 3534.46 $\pm$ 77.53
                & 6276.99 $\pm$ 1878.56
                & 184.89 $\pm$ 58.84
            \\
            
            {}
                & MBPO
                & 2831.23 $\pm$ 1109.63
                & 6285.64 $\pm$ 538.32
                & 145.70 $\pm$ 18.14
            \\
            
            {}
               
                & Ours
                & \textbf{3666.90 $\pm$ 22.71}
                & \textbf{7749.90 $\pm$ 523.27}
                & \textbf{185.14 $\pm$ 1.73}
            \\
            \midrule
             \multicolumn{2}{c}{} & HalfCheetah & Ant & Humanoid \\
            \midrule
            
            \multirow{2}{*}{\makecell[c]{MFRL\\ (@5M steps)}}
                & SAC
                & 16015.64 $\pm$ 351.21
                & 7105.49 $\pm$ 169.61
                & 8036.12 $\pm$ 480.60
            \\
            
            {}
                & PPO
                & 6733.45 $\pm$ 1528.87
                & 4427.39 $\pm$ 836.02
                & 3068.95 $\pm$ 1600.89
            \\
            
            \midrule
            
            \multirow{5}{*}{MBRL}
             & PETS
                & 12023.84 $\pm$ 3340.02
                & 3558.99 $\pm$ 140.76
                & 1335.84 $\pm$ 292.27
            \\
            
            {}
                & SLBO
                & 3993.43 $\pm$ 127.17
                & 2492.19 $\pm$ 92.02
                & 644.16 $\pm$ 237.00
            \\
            
            {}
                & AutoMBPO
                & 12044.35 $\pm$ 1550.21
                & 5792.35 $\pm$ 415.46
                & 5780.14 $\pm$ 245.01
            \\
            
            {}
                & MBPO
                & 13171.53 $\pm$ 937.65
                & 5894.45 $\pm$ 702.39
                & 5905.68 $\pm$ 420.64
            \\
            
            {}
               
                & Ours
                & \textbf{14623.45 $\pm$ 612.10}
                & \textbf{6798.39 $\pm$ 196.84}
                & \textbf{6967.54 $\pm$ 317.07}
            \\
            \bottomrule
        \end{tabular}}
    \end{center}
    \label{maximum-average-returns}
\end{table}

\subsection{Implementation details of CMLO}

\paragraph{Modeling and learning the dynamical models.} 
As inferred from the optimization objective, the minimization of the objective function can be achieved when we try to minimize the difference between $M_2$ and the real environment. To reduce model bias, we chose to use NLL as a loss function in our implementation, which has been shown an effective way to learn model dynamics.
More specifically, CMLO adopts a bootstrap ensemble of dynamical models $\{ \hat {f_{\phi_1}},\hat {f_{\phi_2}}, \ldots, \hat {f_{\phi_K}}\}$ .
Specifically, each forward dynamical model $f_{\phi_i}$  approximates the transition function of the real environment, that is $ \hat{s}_{t+1} \sim f_{\phi_i}(s_t, a_t) $.
The probabilistic models  are fitted on shared but differently shuffled replay buffer ${\cal D}_e$,  and the target is to optimize the Negative Log Likelihood (NLL).
\begin{equation*} \label{nll}
    {\cal L}^H(\phi) = \sum\limits_{t}^{H}[\mu_{\phi}(s_t,a_t)-s_{t+1}]^T\Sigma_{\phi}^{-1}(s_t,a_t)[\mu_{\phi}(s_t,a_t)-s_{t+1}] + \log \det \Sigma_{\phi}(s_t,a_t)
\end{equation*}
And the prediction for these ensemble models is, $\hat{s}_{t+1} = \frac{1}{K}\sum_{i = 1}^{K} f_{\phi_i}(s_t, a_t)$.  More details on network settings are presented in Table~\ref{hyperparameters}.

\paragraph{Model shifts estimation.}
Recall that we partition the incalculable model shifts into two components for estimation, one for state-space coverage and the other for model divergence.
\begin{itemize}[leftmargin=16pt]
\vspace{-5pt}
  \item \textit{state-space coverage}.
  State coverage (policy coverage) is the range of state spaces that our algorithm can explore in the real environment under the current policy $\pi_i$ (derived from the learned model $M_i$). In the existing works, [1] defined the return set for two state sub-space as $\bar{R}_{ret}=\lim_{n\rightarrow\infty}R_{ret}^n(X,\bar X)$, where $R_{ret}^n(X,\bar X) $ means an n-step returnability from $X$ to $\bar X$. Referring to this definition, the state coverage of $\pi_i$ can be defined as $\mathcal{S}_{pc}^{\pi_i}: \forall s\in \mathcal{S}_{pc}^{\pi_i}, a\sim\pi_i(\cdot|s), s'\sim P(\cdot|s,a)\in\mathcal{S}_{pc}^{\pi_i}$. Besides, in the description of La Salle's Invariance Principle [2], we verify the equivalence of Invariant Set and state coverage. Intuitively, the Humanoid example in our response to your major concerns also shows that the variation of state coverage in the different training stages.
  
  We estimate the policy coverage (state-space coverage) by computing the volume $vol({\cal S}_{\cal D})$ of the convex closure $\gS_{\gD}$ constructed on the replay buffer ${\cal D}$.  Since estimation on the full historical experiences involves a huge computational burden, we instead sample $N$ tuples (\emph{e.g.} 1000 tuples) from the replay buffer upon each estimation.  As for the convex hull, we first perform Principal Component Analysis on the states to reduce the dimension and then leverage the Graham-Scan algorithm to construct a convex hull of these $N$ points, which only takes $\gO(N\log N)$ for time complexity.
  
  \item \textit{model divergence}. We estimate the model divergence by computing the average prediction error on newly encountered data. Upon it, we get the estimation for the model divergence from the $K$ ensemble models, ${\cal L}(\Delta{\cal D}) = \mathop{\mathbb{E}}_{(s,a,s')\in \Delta_{\cal D}} \big[\frac{1}{K}\sum_{i=1}^K \Vert s' - \hat{f_{\phi_i}}(s,a) \Vert\big]$. 
\end{itemize}

\paragraph{Event-triggered mechanism.}
Recall our proposed optimization problem:
\begin{equation*} 
    \begin{aligned}
& \min_{M_2\in {\cal M}\atop \pi_2 \in \Pi}  \mathop{\E}\limits_{s,a\sim d^{\pi_2}} \Big[ \sum_{s'\in \gS}\vert  P(s'\vert s,a) - P_{M_2}(s'\vert s,a) \vert \Big],\\
& \textit{s.t.}\quad  \sup\limits_{s\in \gS, a\in \gA}\TV (P_{M_1}(\cdot \vert s,a)\Vert P_{M_2}(\cdot \vert s,a) )\leq \sigma_{M_1,M_2}.
    \end{aligned}
\end{equation*}

We design an event-triggered mechanism to determine the interval instant $\tau$ on the condition that the optimization problem is solved at step $t + \tau$. The mechanism is developed based on the difference between the current model $M_1$ (trained at step $t$) and the upcoming model $M_2$, which is estimated on the newly encountered data. If their model shifts  reaches a certain value, it stands to reason that a new dynamic model is required to be trained. 
Thus, the event-triggered mechanism is based on the condition:
\begin{equation*}
\label{event-condition}
    \frac{vol(\gS_{\gD_t \cup \Delta D(\tau)})}{vol(\gS_{\gD_t})} \cdot \Ls(\Delta D(\tau)) \geq \alpha .
\end{equation*}
Here, we adopt the fraction form for the triggered condition. Denominator $vol({\cal S}_{{\cal D}\cup \Delta {\cal D}(\tau)})\cdot {\cal L}(\Delta {\cal D}(\tau))$  is used to obtain an estimation for the model shift, as detailed in Line 239-248. The numerator $vol (\cal S_D)$, on the one hand, is to reduce numerical errors; on the other hand, this fraction reflects the relative change of the policy coverage and model shift if we turn to train $M_2$ under different $\tau$  starting from $M_1$. This fraction reflects the current ability to digest new data. It can facilitate the setting of threshold, for we do not need to tune $\alpha$ once the policy coverage updates.

In CMLO, the condition estimation execute per $F$ steps because excessively frequently estimation doesn’t make huge difference but bring up the computation load. In order to reuse the result of intermediate computation, we apply a log value to the result of each estimation so that the log value condition function can be approximated by the sum of each estimation value within the interval. We additionally append a constant $\beta = 1.0$ for the penalty of accumulated interval steps.
\begin{equation*}
    \sum_{i=0}^{[\tau/F]} \log \Big(\frac{vol(\gS_{\gD_t \cup \Delta\gD(Fi)})}{vol(\gS_{\gD_t})} \cdot \gL(\Delta \gD(Fi)) + \beta \Big) \geq \alpha
\end{equation*}
\begin{wrapfigure}[21]{r}{0.48\textwidth}
    \vspace{-5pt}
        \centering
        \includegraphics[width=0.9\linewidth]{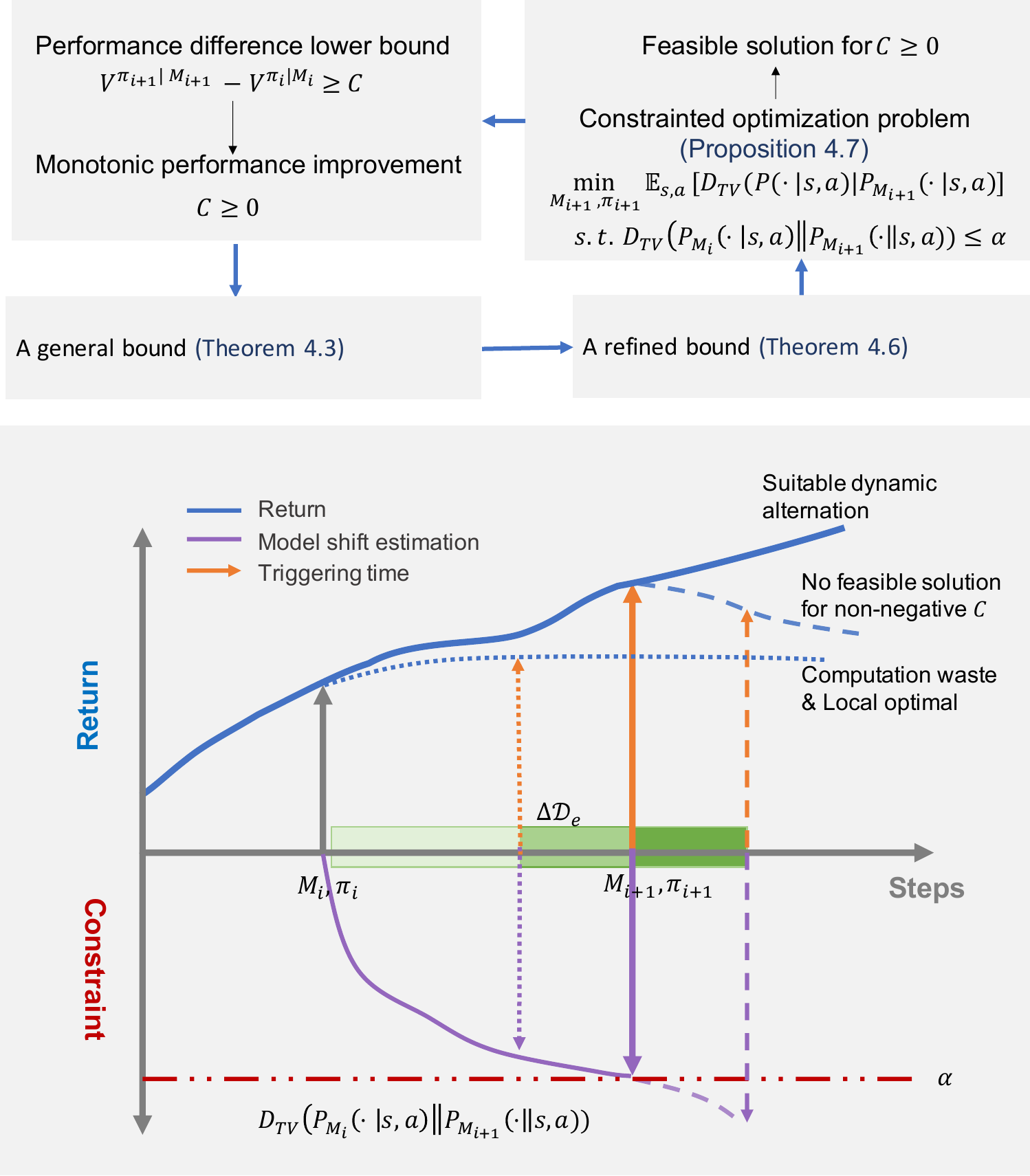}
    \caption{Illustration of the proposed event-triggered mechanism. }
     \vspace{-5pt}
    \label{fig:mechanism-structure}
\end{wrapfigure}
\textit{Remark 1:} We observe from the event-triggered condition that the interevent time can be enlarged by increasing the threshold $\alpha$, which implies more exploration samples will be  collected by current policy and the optimization objective will be solved less frequently. In other words, the triggered threshold is environment-specific. Notably, the tuning load required for our event-triggered mechanism is not heavier than in those fixed settings, due to that only a hyperparameter $\alpha$ is introduced for model training frequency in CMLO, while those algorithms with fixed settings need to tune the fixed model training interval. We claim that it is crucial to dynamically adapt the numbers of explorations to update the model  according to the current training and exploration status.

\textit{Remark 2:} Zeno behavior~\cite{heemels2012introduction} is common in the event-triggered mechanism, which leads to a most frequently triggering. The zeno behavior is naturallt alleviated by the introdution  of $\beta$, which acts as a penalty for interevnet time. We additionaly add the lower and upper bounds of the interevent time to further keep the interval in a safe zone to aviod zeno behavior caused by some extreme situations. The minimal and maximal interevent time are given by 
$\inf\{\tau\} = \underline{T}$ and $\sup \{ \tau\}= \overline{T}$. 

\paragraph{Policy optimization and model rollouts.} We can adopt a standard off-policy model-free RL method  SAC~\cite{haarnoja2018soft} as the policy optimization oracle of CMLO. 
Another key concern is the way of training data generation. 
We adopt the truncated short model rollouts strategy inspired by some current MBRL works  ~\cite{janner2019trust,pan2020trust,lai2021effective}, which helps to escape from compounding error and encourage model usage. 
The main difference from the general rollouts mechanism is that we restrict our rollouts to be generated from fresh models, rather than using outdated models to generate rollout data as MBPO~\cite{janner2019trust} and AutoMBPO~\cite{lai2021effective} do in their implementations.
Based on the dataset ${\cal D}_m$ of the fresh model rollouts, we perform  SAC. In the policy evaluation step, SAC repeatedly apply a Bellman backup operator ${\cal T}^\pi$ to the soft Q-value, ${\cal T}^{\pi} Q^{\pi}(s,a) {\triangleq} r(s,a) + \gamma \mathop{\mathbb{E}}\limits_{s'} [V^\pi(s')]$, and in the policy improvement step, SAC updates the policy according to $\pi = arg\min\limits_{\pi \in \Pi}\mathbb{E}_{s_t \in {\cal D}_m}{\cal D}_{KL}(\pi(\cdot\vert s_t)\Vert \exp(Q^\pi -V^\pi))$.

\textit{Remark 3:} The data distribution introduced by the outdated model has a drift from the data distribution introduced by the fresh model. This data shift will somehow mislead policy training and, in addition, the policy trained on the outdated model suffers from the limited sampling coverage during interacting with the real environment, which might in return cause the following models to fall into a local trap. In other words, the less the  model differs from the real dynamics, the data it rolls out is more valuable.

\subsection{Hyperparameters}
\paragraph{Hyperparameters for Main Experiments.}
Table~\ref{hyperparameters} lists the hyperparameters used in training CMLO. Here, $x\rightarrow y$ over epochs $a\rightarrow b$ denotes a threshold linear function, \emph{i.e.}, at epoch $t$, $f(t) = \min(\max(x + \frac{t-a}{b-a}\cdot (y-x), x),y)$.
\begin{table}[h]
    \caption{Hyperparameter Settings for CMLO.}
    \begin{center}
    \resizebox{\textwidth}{!}{
        \begin{tabular}{
            >{\centering}m{0.15\textwidth}
            | c
            | c
            | c
            | c
            | c
            | c
        }
            \toprule
            & Hopper & Walker &  Swimmer & HalfCheetah &Ant &  Humanoid  \\
            \midrule
            epochs
            & 300
            & 125
            & 300
            & 300
            & 400
            & 250
            \\
            \midrule
            environment steps per epoch & \multicolumn{6}{c}{
                1000
            } \\
            \midrule
            dynamical models network & \multicolumn{6}{c}{
                Gaussian MLP with 4 hidden layers of size 200
            }\\
            \midrule
            ensemble size & \multicolumn{6}{c}{
                5
            }\\
            \midrule
            model rollouts per policy update & \multicolumn{6}{c}{
                400
            } \\
            \midrule
            rollout schedule 
            & \makecell[c]{{1 $\rightarrow$ 15} \\over epochs\\ {20 $\rightarrow$ 100}}
            &\multicolumn{3}{c|}{
                1
            }
            & \makecell[c]{{1 $\rightarrow$ 25} \\over epochs\\ {20 $\rightarrow$ 100}}
            & \makecell[c]{{1 $\rightarrow$ 25} \\over epochs\\ {20 $\rightarrow$ 300}}
            \\
            \midrule
            SAC policy network
            &\multicolumn{4}{c|}{
                Gaussian  with hidden size 512
            }
            &\multicolumn{2}{c}{
                Gaussian  with hidden size 1024
            }
            \\
            \midrule
            policy updates per step
            & 40
            & 20
            & 20
            & 10
            & 20
            & 20
            \\
            \midrule
            event-triggered threshold $\alpha$
            & 1.2
            & 3.0
            & \multicolumn{3}{c|}{2.0}
            & 2.5
            \\
            \midrule
            computing frequency $F$
            &20
            &\multicolumn{5}{c}{50}
            \\
            \midrule
            minimal interevent time $\underline{T}$
            &\multicolumn{6}{c}{150}
            \\
            \midrule
            maximal interevent time $\overline{T}$
            &\multicolumn{6}{c}{500}
            \\
            \bottomrule
        \end{tabular}
        }
    \end{center}
    \label{hyperparameters}
\end{table}

\paragraph{Hyperparameters for Ablation Studies.} Note that other hyperparameters we do not mention below are the same as hyperparameter settings in Table~\ref{hyperparameters}. 

\textbf{(1) Policy optimization oracle: TRPO.} For the TRPO part, the key parameters are listed below:
\begin{itemize}[leftmargin=16pt]
\vspace{-5pt}
\item Ant: $horizon = 1000, \gamma=0.99, gae=0.97, step\_size=0.01, iterations=40$
\item HalfCheetah:  $horizon = 1000, \gamma=0.99, gae=0.95, step\_size=0.01, iterations=40$
\end{itemize}
About Legend w/o-n, we use a data sampler with batchsize=20, thus we get $20\times n$ real interactions during the model training interval.
We compute the total triggered times and scale them to [0,1], which is shown in the bar plots.

\textbf{(2) Policy optimization oracle: iLQR.} 
Dynamical models network: Gaussian MLP with 3 hidden layers of size 200, batch size is 64, and the learning rate is 0.0001. 
For the iLQR part: $LQR\_ITER=10, R=0.001, Q=1, horizon=5$. 
About Legend w/o-n, we get n real interactions during the model training interval. And $\alpha=0.5$ in w/-ours. 
We compute the total triggered times and scale them to $[0,1]$, as shown in the bar plots.

\subsection{Additional Ablation Study}

\begin{figure}[h]
    \centering
        \includegraphics[width = 0.25\textwidth]{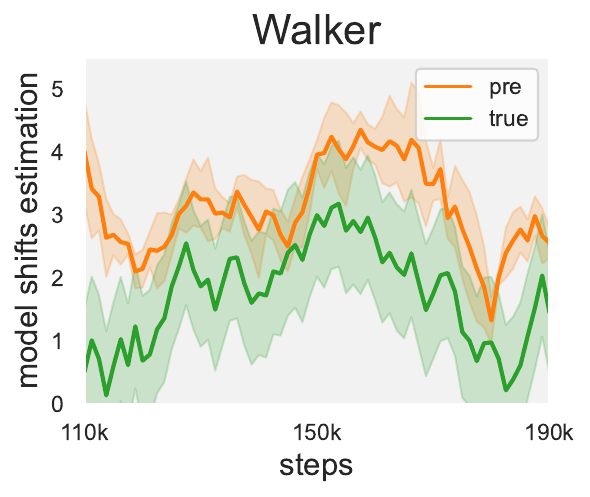}%
        \includegraphics[width = 0.25\textwidth]{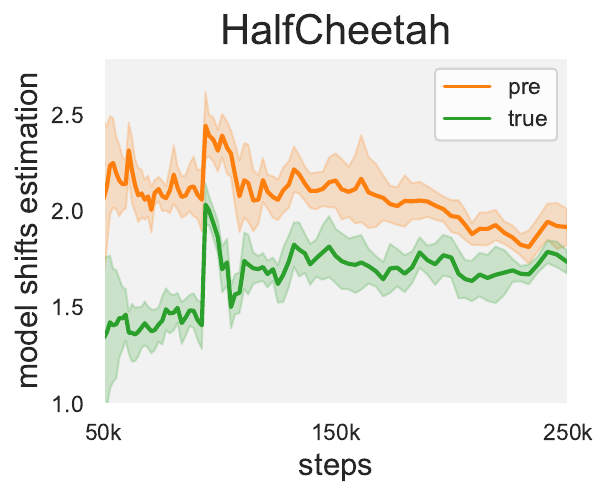}%
        \includegraphics[width = 0.25\textwidth]{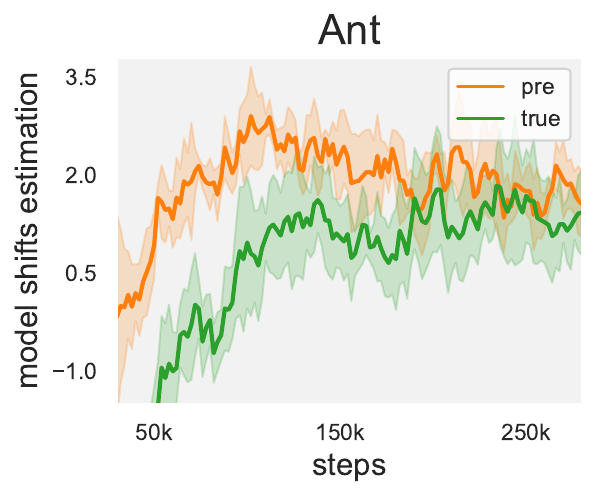}%
        \includegraphics[width = 0.25\textwidth]{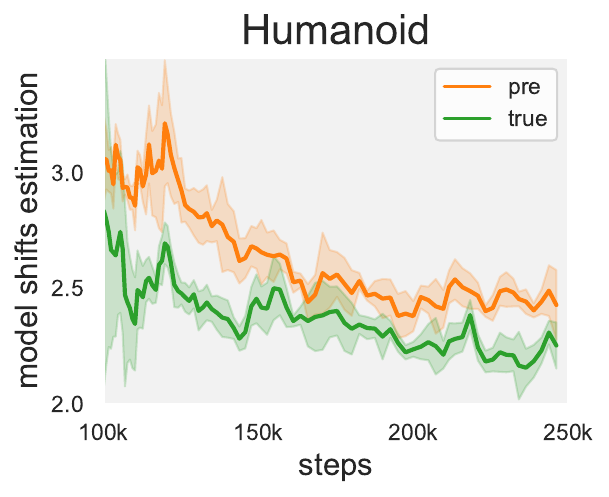}
     \hfill
    \caption{Estimation of model shifts in four environments. Green lines imply the estimation on model shifts after updating, which could be considered as a real value of our constraint function. And orange lines show our pre-estimation on model shifts before model updating.}
    \label{fig:model-shifts}
\end{figure}

\paragraph{Estimation on model shifts.}
The constraint function based on  model shifts is incalculable due to the unobserved model $M_2$. We design a practical predictor for the model shifts by computing the state-space coverage and the model prediction error. Also, the decoupling of the constraint and the objective is enabled partly owing to the slightly overestimation over the true value.
Recall our constraint function: 
\begin{equation*}
\begin{split}
    &\TV(P_{M_1}(\cdot\vert s,a) \Vert P_{M_2}(\cdot \vert s,a)) = \frac{1}{2}\sum_{s'\in \gS}\Big[ \vert P_{M_1}(s'\vert s,a) - P_{M_2}(s'\vert s,a)\vert \Big] \\
    \leq& \sum_{s'\in \gS}\frac{1}{2}\Big[\vert P_{M_1}(s'\vert s,a) - P_{M}(s'\vert s,a)\vert  + \vert P_{M_2}(s'\vert s,a) - P_{M}(s'\vert s,a)\vert  \Big]
\end{split}
\end{equation*}
The updated dynamics $P_{M_2}$ usually comes closer to the true dynamics than the previous one $P_{M_1}$, thus we turn to estimate $\sum_{s'\in \gS}[\vert P_{M_1}(s'\vert s,a) - P_{M}(s'\vert s,a)\vert]$. 
Once the model $M_2$ is trained, we can actually conduct a more realistic calculation for the constraint function. 
To show the connection between our predicted value in the absence of $M_2$ and the estimated value after obtaining $M_2$, we perform experiments on four environments and results are shown in Figure~\ref{fig:model-shifts}.
The results demonstrate that our prediction is higher than the true estimation and their trends stand consistent, which indicates that our predictor is well designed. This gap helps to decouple the constrained optimization problem,  and this overestimation part can be bridged by adjusting the $\alpha$.

\begin{figure}[h]
    \centering
        \includegraphics[width = 0.25\textwidth]{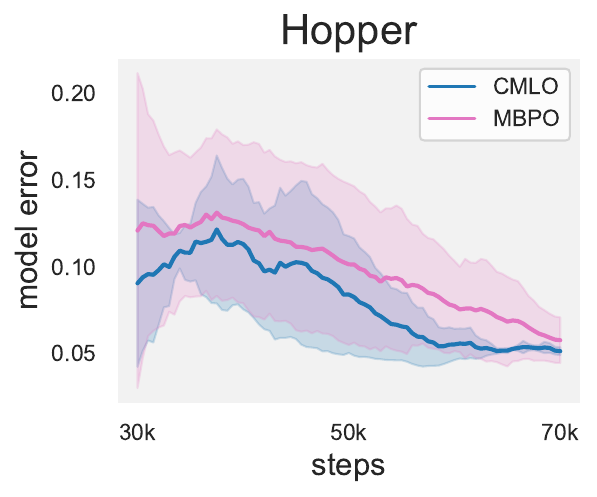}%
        \includegraphics[width = 0.25\textwidth]{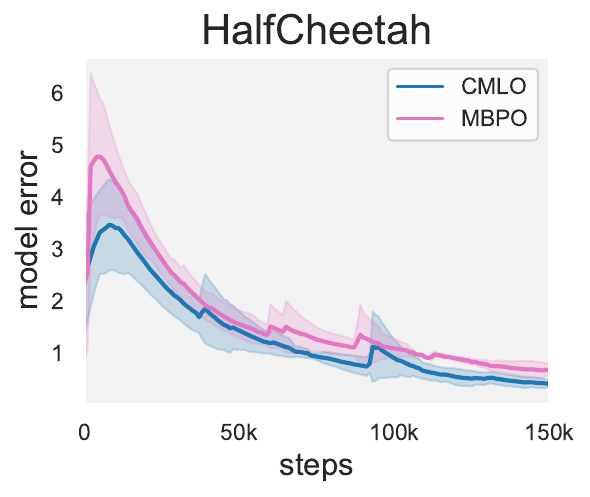}%
        \includegraphics[width = 0.25\textwidth]{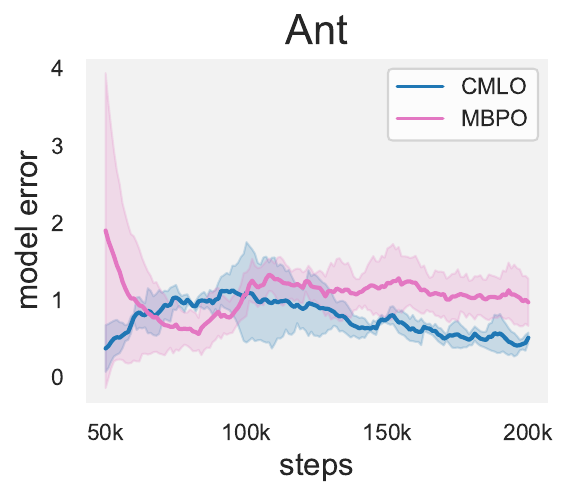}%
        \includegraphics[width = 0.25\textwidth]{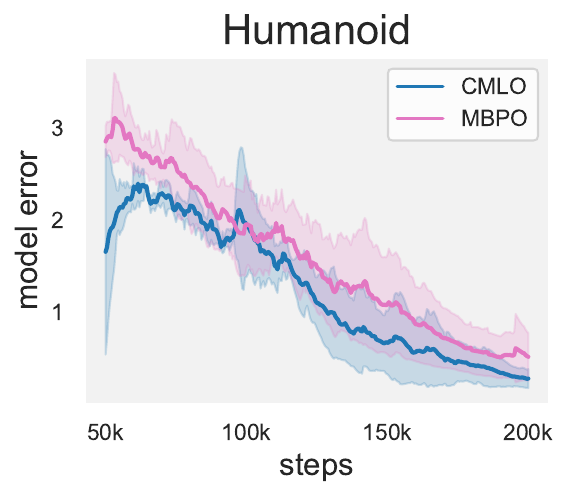}
     \hfill
    \caption{One-step model error in four benchmark tasks.}
    \label{fig:model-error}
\end{figure}

\paragraph{Model accuracy.}
Figure~\ref{fig:model-error} shows the one-step model error during the training under four benchmark tasks. 
We find that CMLO achieves a more accurate model than the state-of-the-art baseline MBPO. This result agrees with our insight that, a smarter scheme to choose different numbers of explorations at different steps instead of the unchanged setting in current methods, will promote a better model.

\paragraph{Policy Coverage Comparison.}
Policy coverage represents the exploration ability of the policy. The policy coverage increasing with the stages means that the policy has new explorations at every stage and may not fall into a local optimum. Here, we present the numerical comparison to MBPO in Table~\ref{ab-pc}.

\begin{table}[htbp]
    \scriptsize
    \caption{Policy Coverage Comparison to MBPO in different stages.}
    
    \begin{center}
    \resizebox{0.8\textwidth}{!}{
        
        \begin{tabular}{
            c|c|ccccc
        }
            \toprule
            \multicolumn{2}{c}{} & Stage1 & Stage2 & Stage3 & Stage4 & Stage5\\
            \midrule
            
            \multirow{2}{*}{\makecell[c]{HalfCheetah}}
                & CMLO
                & 138.57
                & 182.28
                & 243.47
                & 302.82
                & 344.36

            \\
            
            {}
                & MBPO
                & 129.25
                & 173.09
                & 242.49
                & 264.85
                & 338.55
            \\

            \midrule
             \multicolumn{2}{c}{} & Stage1 & Stage2 & Stage3 & Stage4 & Stage5 \\
            \midrule
            
            \multirow{2}{*}{\makecell[c]{Ant}}
                & CMLO
                & 354.15
                & 744.92
                & 849.47
                & 876.12
                & 909.80
            \\
            
            {}
                & MBPO
                & 342.13
                & 729.30
                & 821.66
                & 864.93
                & 880.25
            \\

            \bottomrule
        \end{tabular}}
    \end{center}
    \label{ab-pc}
\end{table}

Here, each stage $i$ contains $(60\times(i-1), 60\times i ]k$ steps. In HalfCheetah, we find that our policy achieves higher coverage especially in first 4 stages than MBPO. Consistently, we find that our policy enjoys higher performance, with an average return lead of about 1855.29 over MBPO in the first 300k steps. Likewise, the growth of policy coverage in Ant is also consistent with the rise in average return. The increase in policy coverage helps the policy to refrain from falling into a local optimum, thus improving performance.

\begin{figure}
    \centering
    \begin{minipage}[t]{1.0\textwidth}
    \centering
        \begin{subfigure}[t]{1.0\textwidth}
        \centering
            \includegraphics[width = 0.24\textwidth]{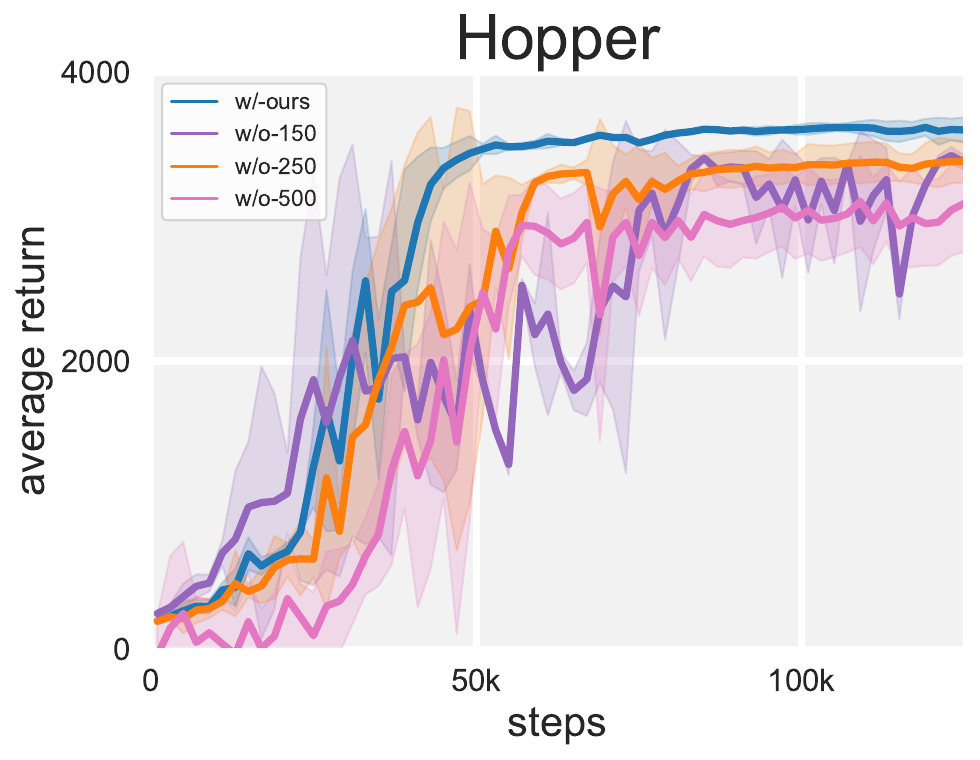}
            \includegraphics[width = 0.24\textwidth]{./Figures/ab-halfcheetah-value.pdf}
            \includegraphics[width = 0.24\textwidth]{./Figures/ab-ant-value.pdf}
            \includegraphics[width = 0.24\textwidth]{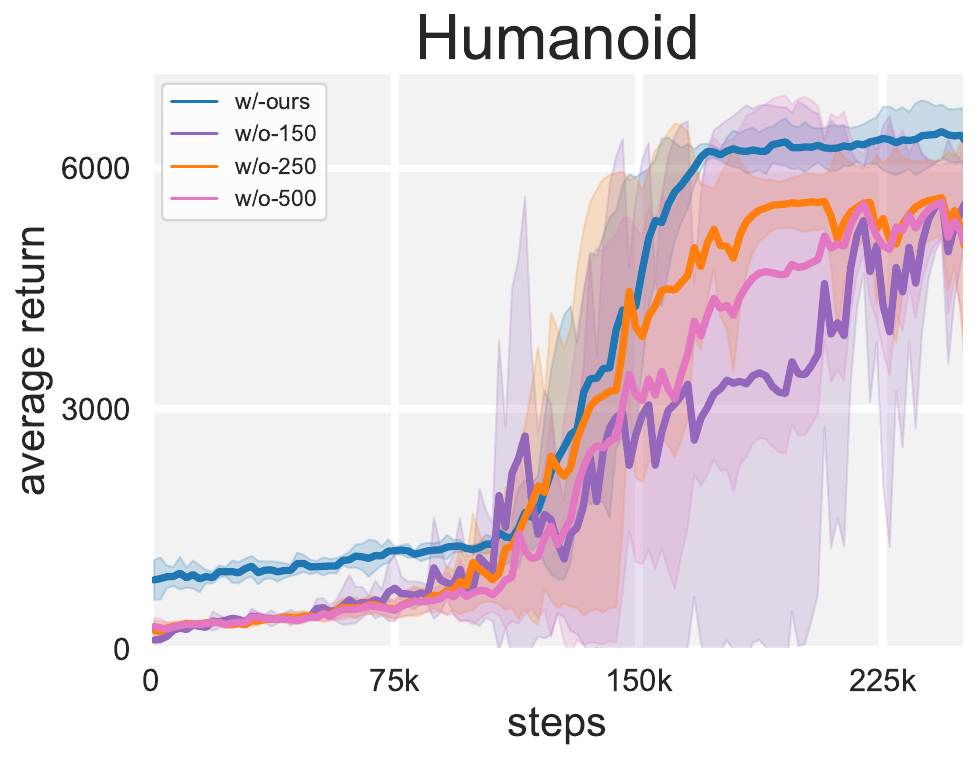}
            \caption{average return}
        \end{subfigure}
    \end{minipage}
    \hspace{10pt}
    \begin{minipage}[t]{1.0\textwidth}
    \centering
        \begin{subfigure}[t]{1.0\textwidth}
        \centering
            \includegraphics[width = 0.24\textwidth]{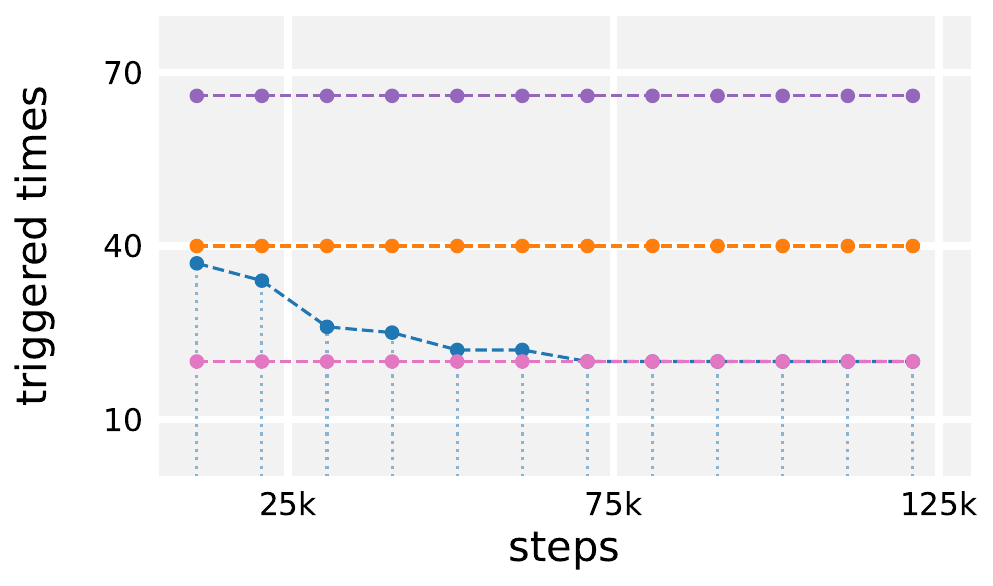}
            \includegraphics[width = 0.24\textwidth]{./Figures/ab-halfcheetah-freq.pdf}
            \includegraphics[width = 0.24\textwidth]{./Figures/ab-ant-freq.pdf}
            \includegraphics[width = 0.24\textwidth]{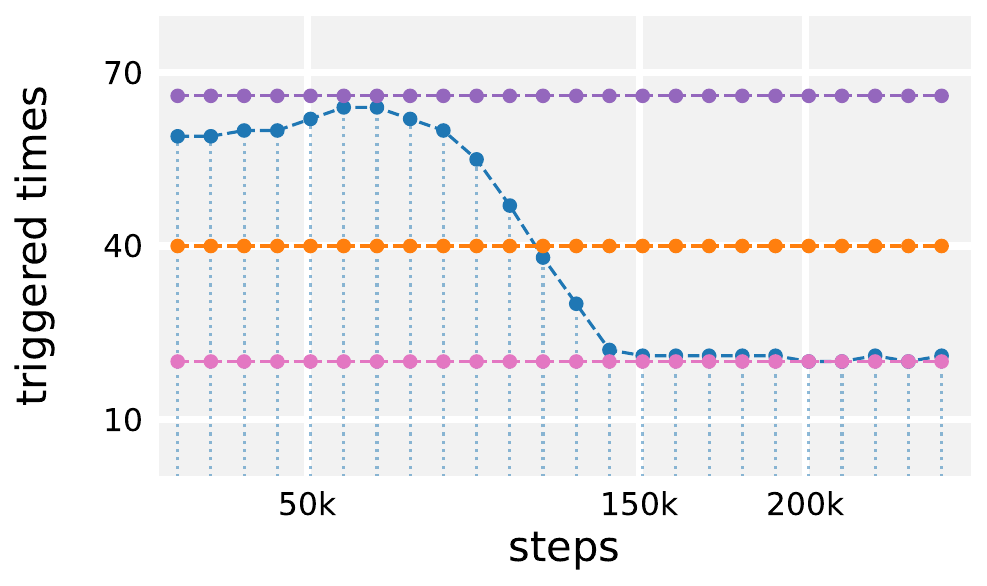}
            \caption{triggered times}
        \end{subfigure}
    \end{minipage}
    \caption{Ablation on event-triggered mechanism.  (a,c) shows the average return with or with-out event-triggered mechanism in HalfCheetah and Ant benchmarks. (b,d) shows the average number of triggered times per 10k step. All the experiments are average over 4 random seeds.}
    \label{fig:returns}
\end{figure}

\paragraph{Effectiveness of event-triggered mechanism.}
We compare applying model shift constraints to the unconstrained cases and the results are shown on Figure \ref{fig:returns}.  To verify the effectiveness of adding suitable constraints, we invalidate the event-triggered mechanism and keep the other part unchanged in our method.  
As observed, our model  is accurate enough when fixing the model training interval at $250$,  but it still performs worse than applying the model shift constraints. We attribute our model's out-performance to our rational model shift design. It improves the performance while minimizing the training cost of the model. Adding such a model shift can protect the model from overfitting on under-explored data, and can also save the model from  fitting large data shifts.

To determine whether the event-triggered mechanism has an effect on the training process, we conduct a t-test to compare the average returns of CMLO with or without the mechanism. We compare the original CMLO to its variant with a fixed setting (w/o-250) and list the p-values in Table~\ref{tb:main-ablation}. 
Our p-values are much smaller than 0.05, so we say with a high degree of confidence that the smartly choosing dynamically varying number of explorations does make a difference in the overall performance.
\begin{table}[h]
    \caption{t-test to the average returns of CMLO with or without event-triggered mechanism.}
    \begin{center}
        \begin{tabular}{
            >{\centering}m{0.15\textwidth}
            | c
            | c
            | c
            | c
            | c
            | c
        }
            \toprule
            & Hopper & Walker &  Swimmer & HalfCheetah & Ant &  Humanoid \\
            \midrule
            p-value 
            & 0.0141
            & 4.74e-10
            & 2.47e-5
            & 7.48e-33
            & 2.78e-26
            & 1.983e-15
            \\
            \bottomrule
        \end{tabular}
    \label{tb:main-ablation}
    \end{center}
\end{table}

\begin{figure}[h]
    \centering

    \begin{subfigure}[t]{0.9\textwidth}
        \includegraphics[width = 1.0\textwidth]{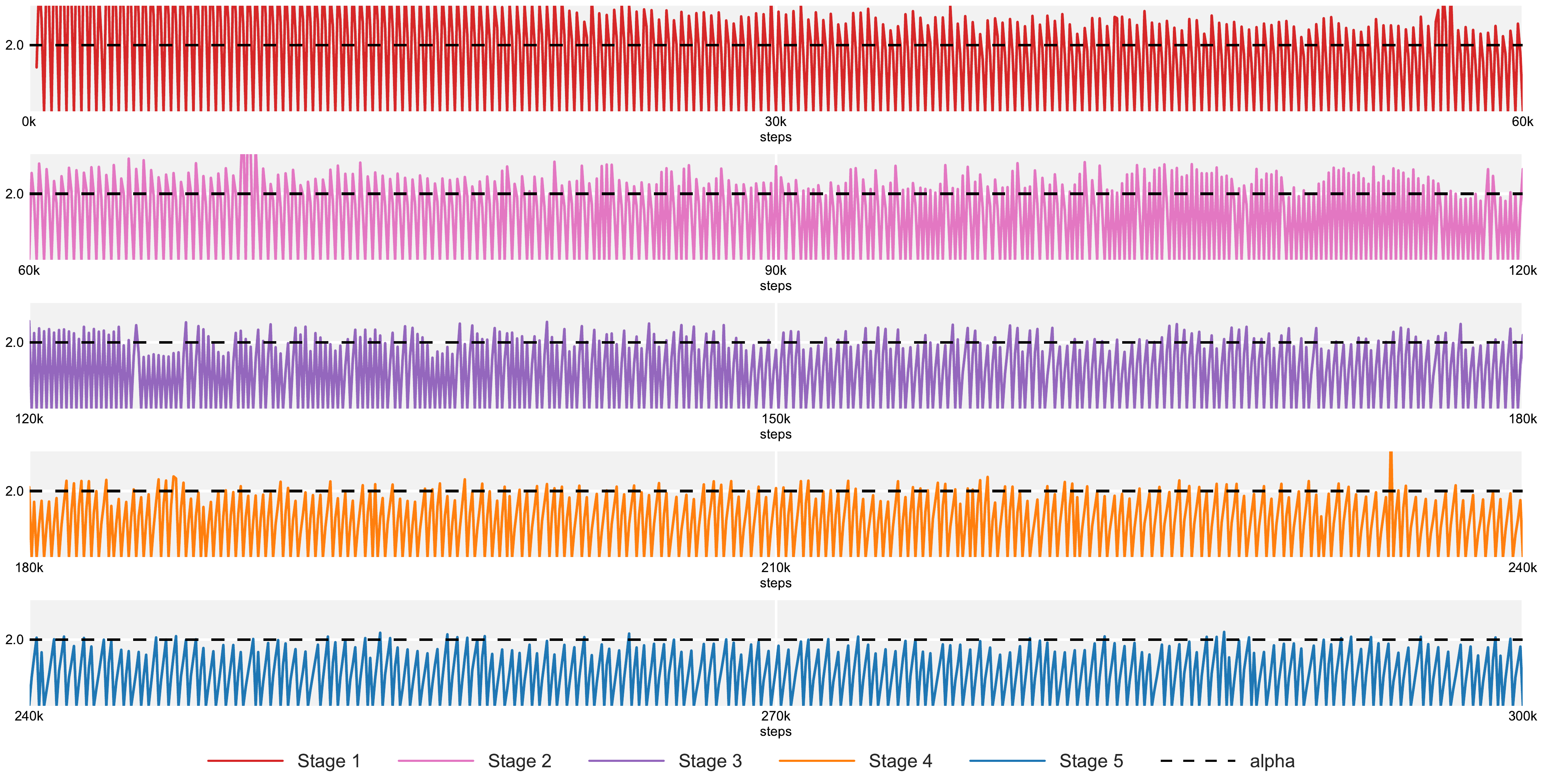}
     \caption{HalfCheetah.}
    \end{subfigure}
    \hfill

    \begin{subfigure}[t]{0.9\textwidth}
        \includegraphics[width = 1.0\textwidth]{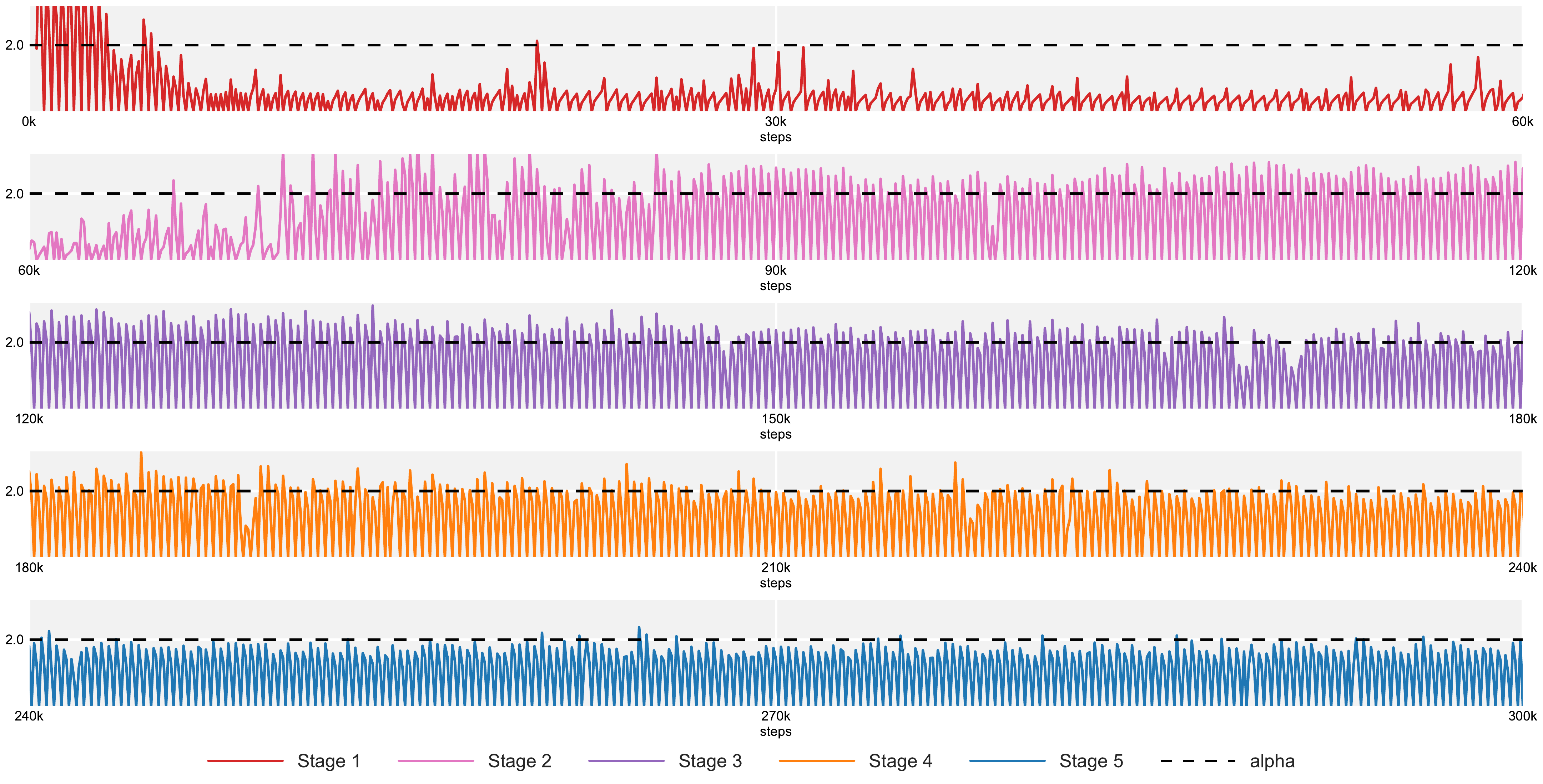}%
        \caption{Ant.}
    \end{subfigure}
    \caption{Visualization of event-triggered mechanism on HalfCheetah and Ant. Solid lines show the model shift estimation and dotted lines are the triggered threshold. Note that here we apply log value.}
    \hfill
    \label{fig:vis-ms}
\end{figure}

Besides, we provide visualization of event-triggered mechanism on HalfCheetah and Ant in Figure~\ref{fig:vis-ms}. The y-axis is our estimation of the triggered condition.  This figure shows the constraint estimation and whether it reaches the triggered threshold (when the peak is above the threshold (dashed line), the primary trigger condition is satisfied) within different stages. Note that in the paper we have shown 4k steps for each stage, and here we present for the  whole 60k.

\subsection{Computing Infrastructure}
Table~\ref{computation} lists our computing infrastructure and the corresponding computational time used for training CMLO on the six benchmark tasks. 
\begin{table}[h]
    \caption{Computing infrastructure and the computational time for each benchmark task.}
    \begin{center}
        \begin{tabular}{
            >{\centering}m{0.15\textwidth}
            | c
            | c
            | c
            | c
            | c
            | c
        }
            \toprule
            &  Hopper & Walker &  Swimmer & HalfCheetah & Ant  & Humanoid \\
            \midrule
            CPU & \multicolumn{6}{c}{
                Intel Core i7-6900K (16 threads)
            } \\
            \midrule
            GPU & \multicolumn{6}{c}{
                NVIDIA TITAN X (Pascal) x 3
            } \\
            \midrule
            computation time in hours
            & 20.15
            & 19.21
            & 31.58
            & 35.97
            & 29.35
            & 33.31
            \\
            \bottomrule
        \end{tabular}
    \label{computation}
    \end{center}
\end{table}
\end{document}